%% file: neurips_main.tex
\documentclass{article}

\usepackage[eandd, final]{neurips_2026}

\usepackage[utf8]{inputenc} 
\usepackage[T1]{fontenc}    
\usepackage{hyperref}       
\usepackage{url}            
\usepackage{booktabs}       
\usepackage{wrapfig}        
\usepackage{amsfonts}       
\usepackage{nicefrac}       
\usepackage{microtype}      
\usepackage{xcolor}         

\usepackage{graphicx}
\usepackage{subcaption}

\usepackage{amsmath}
\usepackage{amssymb}
\usepackage{mathtools}
\usepackage{amsthm}

\usepackage[capitalize,noabbrev]{cleveref}
\usepackage{tabularx}
\usepackage{makecell}
\usepackage{multirow}
\usepackage[many,most]{tcolorbox}
\usepackage{siunitx}
\usepackage[table]{xcolor}
\usepackage{svg}
\usepackage{array}
\usepackage{arydshln}
\usepackage{xurl}
\usepackage{float}          
\usepackage{algorithm}
\usepackage{setspace}

\theoremstyle{plain}

\theoremstyle{definition}

\theoremstyle{remark}

\usepackage[textsize=tiny]{todonotes}

\definecolor{midnightgreen}{rgb}{0.0, 0.29, 0.33}

\title{Evaluating Test-Time Scaling of General LLM Agents}

\author{%
  Xiaochuan Li$^{1}$ \quad
  Ryan Ming$^{1}$ \quad
  Pranav Setlur$^{1}$ \quad
  Abhijay Paladugu$^{1}$ \\
  \textbf{Andy Tang}$^{1}$ \quad
  \textbf{Hao Kang}$^{1}$ \quad
  \textbf{Shuai Shao}$^{2}$ \quad
  \textbf{Rong Jin}$^{2}$ \quad
  \textbf{Chenyan Xiong}$^{1}$ \\[0.2em]
  $^{1}$Carnegie Mellon University
  $^{2}$Meta \thanks{All experiments, data collection, and processing activities were conducted by Carnegie Mellon University. Meta was involved solely in an advisory role. Correspondence to: \texttt{xiaochu4@andrew.cmu.edu}}
}

\begin{document}

\maketitle

\begin{abstract}
LLM agents are increasingly expected to operate as general-purpose systems that resolve real-world user requests, yet their dynamic scaling behavior in realistic environments remains poorly understood. In this paper, we systematically investigate two principal test-time scaling axes of LLM agents: sequential scaling through extended interaction and parallel scaling through trajectory sampling. We first introduce a realistic benchmark that provides one unified framework for evaluating LLM agents across search, coding, reasoning, and tool-use domains, more faithfully reflecting the heterogeneity of real-world deployments. Evaluating ten leading LLM agents reveals substantial performance degradation when transitioning from domain-specific evaluations to this realistic setting. Building on this foundation, we progressively scale test-time compute along fine-grained increments to characterize the performance upper bound. We find that \textbf{neither scaling axis can consistently yield meaningful gains from additional test-time compute in realistic environments}, a phenomenon we attribute to two fundamental limitations: the \textit{scaling plateau} that bottlenecks sequential scaling and the \textit{verification gap} that undermines parallel scaling. Code is publicly available
at \url{https://github.com/cxcscmu/General-AgentBench}.
\end{abstract}

\input{src/1intro}
\input{src/2benchmark}

\input{src/3mainresults}
\input{src/4sequential_tts.tex}

\input{src/4parallel_tts.tex}

\input{src/related_work}

\input{src/6conclusions}

\bibliographystyle{plainnat}
\bibliography{neurips_references}


\newpage
\appendix

\input{src/7appendix}


\newpage
\input{checklist.tex}

\end{document}

%% file: src/1intro.tex
\begin{figure*}[h]
    \vspace{-1.5em}
    \centering
    \begin{subfigure}[t]{0.38\linewidth}
        \centering
        \includegraphics[width=\linewidth]{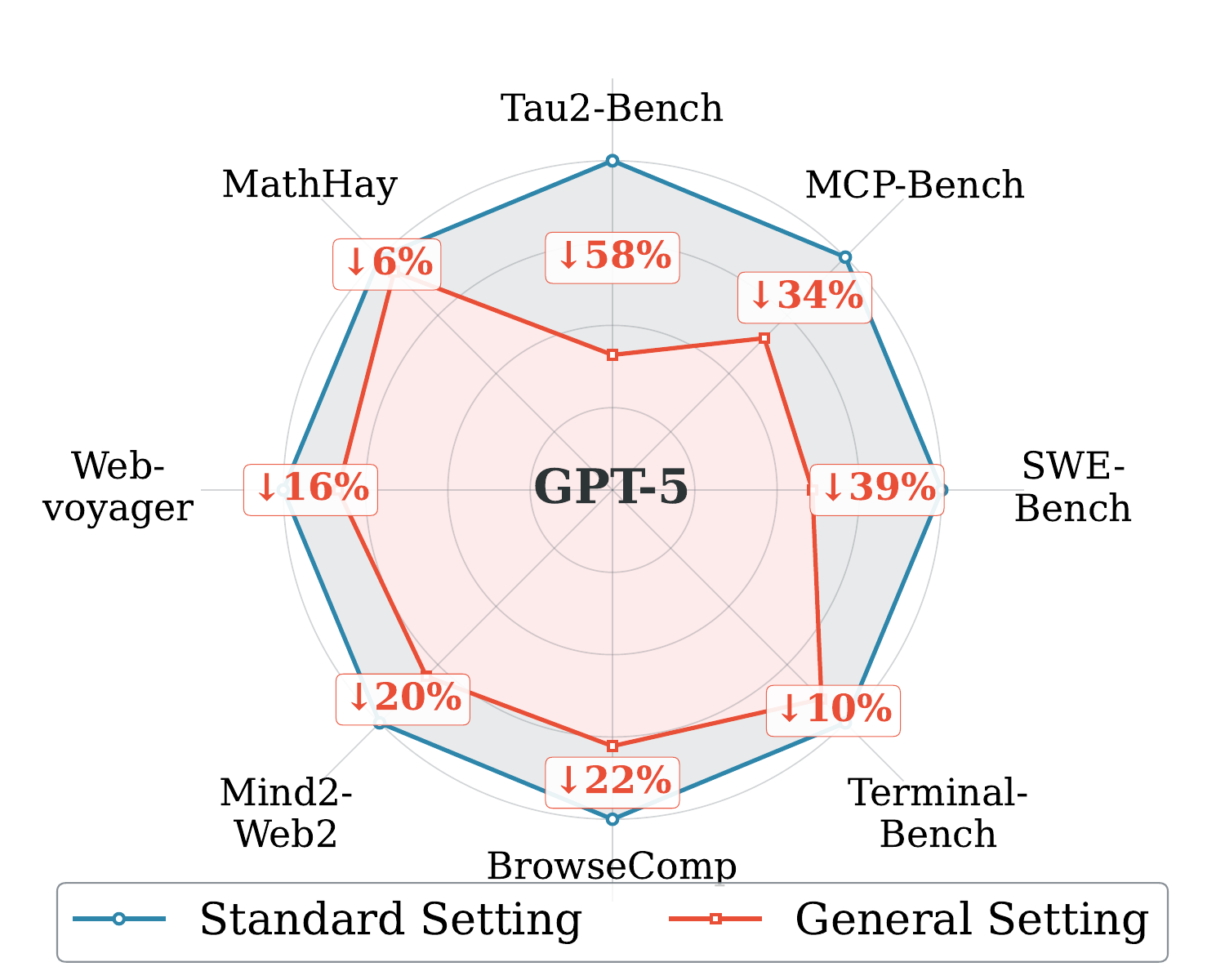}
        \caption{Performance comparison.}
        \label{fig:intro_a}
    \end{subfigure}
    \hfill
    \begin{subfigure}[t]{0.3\linewidth}
        \centering
        \includegraphics[width=\linewidth]{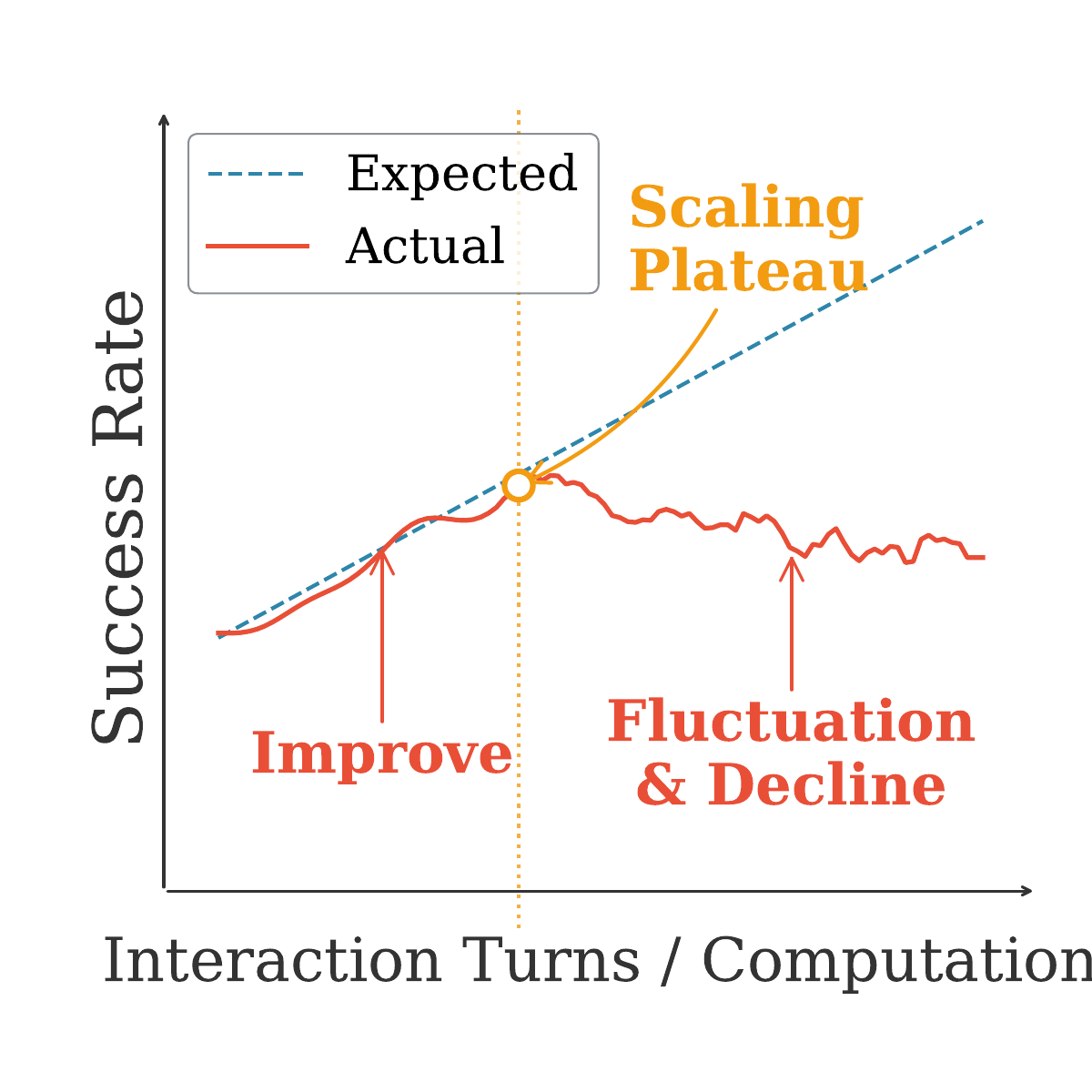}
        \caption{Sequential test-time scaling.}
        \label{fig:intro_b}
    \end{subfigure}
    \hfill
    \begin{subfigure}[t]{0.3\linewidth}
        \centering
        \includegraphics[width=\linewidth]{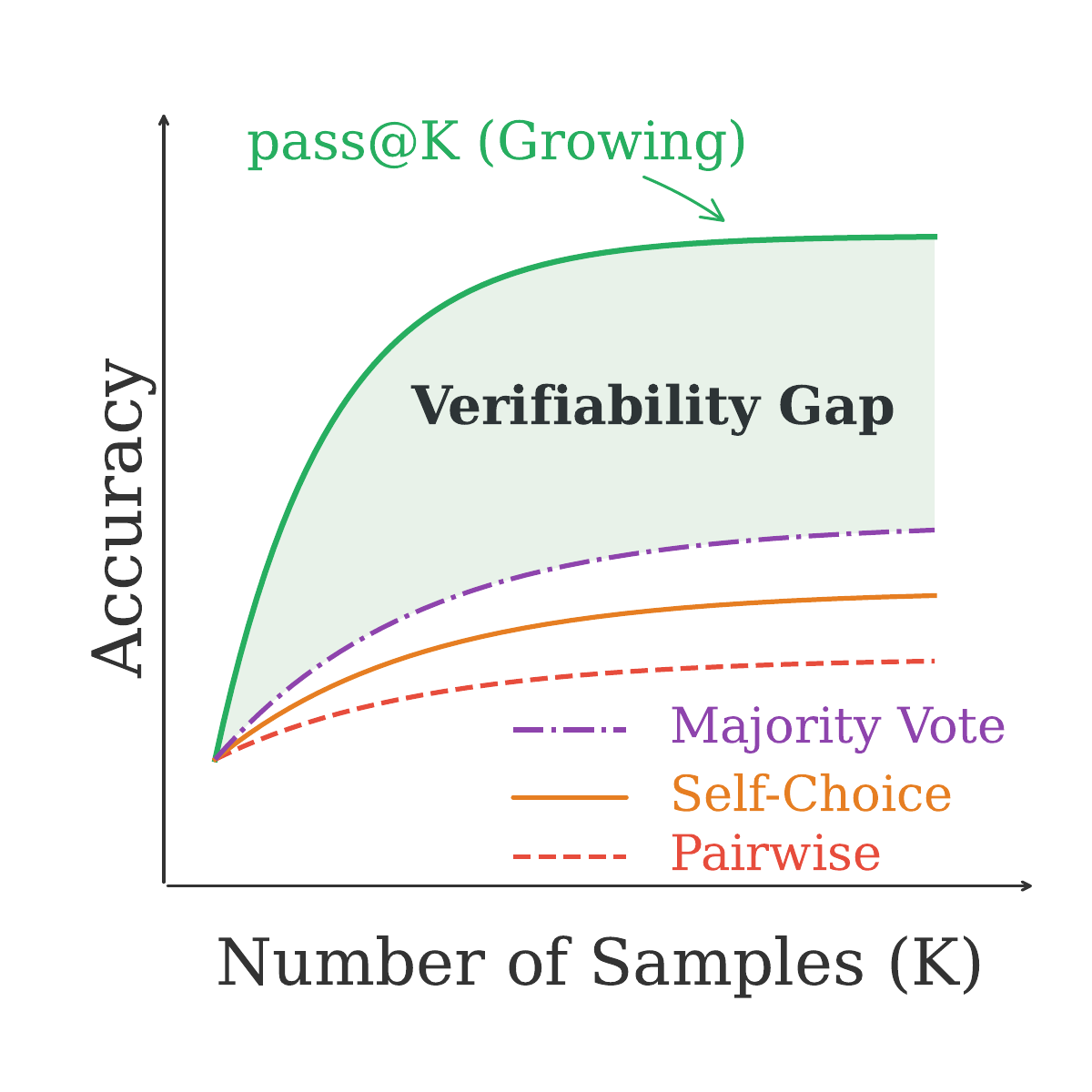}
        \caption{Parallel test-time scaling.}
        \label{fig:intro_c}
    \end{subfigure}

    \caption{\textbf{Main contribution of this paper.}
    \textbf{A}: GPT-5's performance drop under General AgentBench compared to domain-specific evaluation.
    \textbf{B}: Sequential test-time scaling via longer interaction turns out to be unstable or to degrade performance.
    \textbf{C}: Agents fail to utilize the improvement brought by multiple sampling due to verification gap.}
    \label{fig:intro_figure}
\end{figure*}

\section{Introduction}

Agents powered by large language models are at a scaling plateau, transitioning from domain-specific to general-purpose~\citep{liu2023fingpt, yang2024sweagent, luo2025large}. Real-world user requests are often unpredictable and require LLM agents to operate end-to-end through planning~\citep{wang2023plan, erdogan2025planandact}, reasoning~\citep{Wei2022CoT, yao2022react}, and tool use~\citep{schick2023toolformer, patil2024gorilla} under uncertain and evolving conditions that far exceed what the training distribution can cover. Thus, a capable general agent is expected to have strong \textbf{test-time scaling ability}—continuously refining its thoughts and combining multiple skills to solve problems, much like how humans learn by acting. Although the community has developed many effective test-time scaling strategies, an important question has been overlooked: 

\emph{Is there an upper bound of test-time scaling for LLM agents? If so, how can we benchmark and characterize its behavior?}

Existing agentic benchmarks cannot address this. They evaluate LLM agents in domain-specific settings, where the environment and available tools are explicitly designed for a particular task category (e.g., software engineering with terminal tools only~\citep{jimenez2023swebench}, or web navigation with a browsing interface only~\citep{zhou2023webarena, he2024webvoyager, wei2025browsecomp}). In real-world deployment, user queries are rarely so constrained; they do not specify the task category in advance, requiring agents to remain ready across a broad toolset to handle unpredictable requests. As a result, current benchmarks can not fully capture the demands of real-world usage and may overestimate agent robustness under realistic, uncertain conditions. 

In this paper, we first address this gap by introducing General AgentBench, a benchmark designed to evaluate LLM agents under a unified framework that more closely reflects real-world deployment. We consolidate tools from different domains into a shared library that is consistently exposed, while domain-specific environments and implementations remain hidden. We evaluate ten leading LLM agents, each of which must first interpret the user request, then choose suitable tools from a large and diverse tool pool, and interact with the environment until producing a final response.

This benchmark also serves as a well-suited testbed for studying the dynamic behaviors of test-time scaling~\citep{cobbe2021training, zelikman2024quiet} since all selected tasks can benefit from additional compute. We focus on two complementary paradigms: (1) sequential scaling, which extends interaction histories to support continued exploration and exploitation; and (2) parallel scaling, which independently samples $K$ candidate trajectories and selects a single candidate to return. These two correspond to distinct axes of computational allocation—breadth of exploration versus depth of exploitation. We point out that effective parallel scaling requires agents not only to \emph{generate correct solutions}, but also to \emph{reliably identify the correct one}, since in real-world deployment we do not have golden references. The agents must consolidate to a common answer either by majority voting~\citep{Wang2022SelfConsistency} or LLM judging~\citep{zheng2023judging}. Together, these settings enable the first systematic study of dynamic test-time scaling behaviors in general LLM agents.

Our results lead to three key conclusions. (1) Across ten leading LLMs, we observe a substantial performance drop when moving from domain-specific configurations to the general-agent setting (Figure~\ref{fig:intro_a}), with pronounced differences in robustness across model families. (2) Sequential scaling exhibits a \textbf{scaling plateau}: while performance improves within a modest range of additional interaction turns, it often fluctuates or degrades thereafter. On certain tasks, this plateau even happens at the very beginning. This suggests that the accumulated history eventually overwhelms the agent's reasoning capacity, leading to instability in long-horizon tasks (Figure~\ref{fig:intro_b}). (3) Although parallel scaling increases the theoretical upper bound of performance (pass@$K$), we consistently observe a gap between this upper bound and the agent's commitment accuracy, revealing a substantial \textbf{verification gap} that ultimately limits achievable performance in realistic settings (Figure~\ref{fig:intro_c}). In summary, our contributions are:


\begin{itemize}
\item \textbf{General AgentBench for Realistic Evaluation.}
We introduce a benchmark for evaluating general LLM agents' ability under a unified framework across diverse domains, more closely reflecting real-world deployment.

\item \textbf{Sequential Test-time Scaling in General Agents.}
We study the sequential test-time scaling behavior of general agents, showing and defining a scaling plateau that bounds performance improvements, beyond which additional computation often leads to instability and performance degradation. This plateau varies across models and domains.

\item \textbf{Parallel Test-time Scaling.}
We analyze parallel test-time scaling and show that, despite increasing the theoretical performance upper bound pass@$K$, its practical gains are limited by a verification gap between generation and commitment accuracy.
\end{itemize}

%% file: src/2benchmark.tex
\section{General AgentBench}

In this section, we introduce the construction of General AgentBench (Section~\ref{2.1}) and the unified evaluation framework (Section~\ref{2.2}). Detailed benchmark information are deferred to Appendix~\ref{appendix:benchmarks_details}.

\subsection{Domains and Sources}\label{2.1}

\begin{wraptable}{r}{0.55\linewidth}
    \centering
    \small
    \vspace{-1em}
    \caption{Composition of General \mbox{AgentBench}.}
    \label{tab:composition}
    \resizebox{\linewidth}{!}{%
    \begin{tabular}{llrr}
    \toprule
    \textbf{Domain} & \textbf{Dataset} & \textbf{Original} & \textbf{Sampled} \\
    \midrule
    \multirow{2}{*}{Search}
      & BrowseComp   & 1266 & 124\\
      & WebVoyager   & 643  & 65 \\
    \midrule
    \multirow{2}{*}{Coding}
      & SWE-Bench Verified & 500 & 50 \\
      & Terminal-Bench     & 230 & 80 \\
    \midrule
    Reason
      & MathHay            & 602 & 75 \\
    \midrule
    \multirow{2}{*}{Tool-Calling}
      & Tau2-Bench         & 278 & 50 \\
      & MCP-Bench          & 104 & 52 \\
    \bottomrule
    \end{tabular}%
    }
    \vspace{-1em}
\end{wraptable}

Our benchmark spans four task domains: \textbf{Coding}~\citep{openai_swebench_verified_2024}, \textbf{Search}~\citep{wei2025browsecomp,he2024webvoyager}, \textbf{Tool-use}~\citep{barres2025tau2, wang2025mcpbench}, and \textbf{Reason}~\citep{wang2024mathhay}. These domains reflect common real-world applications such as software engineering, information seeking, service workflows, and analytical reasoning, positioning General AgentBench as an initial step toward evaluating general-purpose agents under unified settings. Table~\ref{tab:composition} summarizes the benchmark composition. We do not include short-step problems but instead focus entirely on long-horizon problems, which leaves sufficient space for studying the dynamic behaviors of test-time scaling.





\subsection{Unified Realistic Evaluation Framework}\label{2.2}

\begin{wrapfigure}{r}{0.6\linewidth}
    \centering
    \vspace{-1em}
    \includegraphics[width=\linewidth]{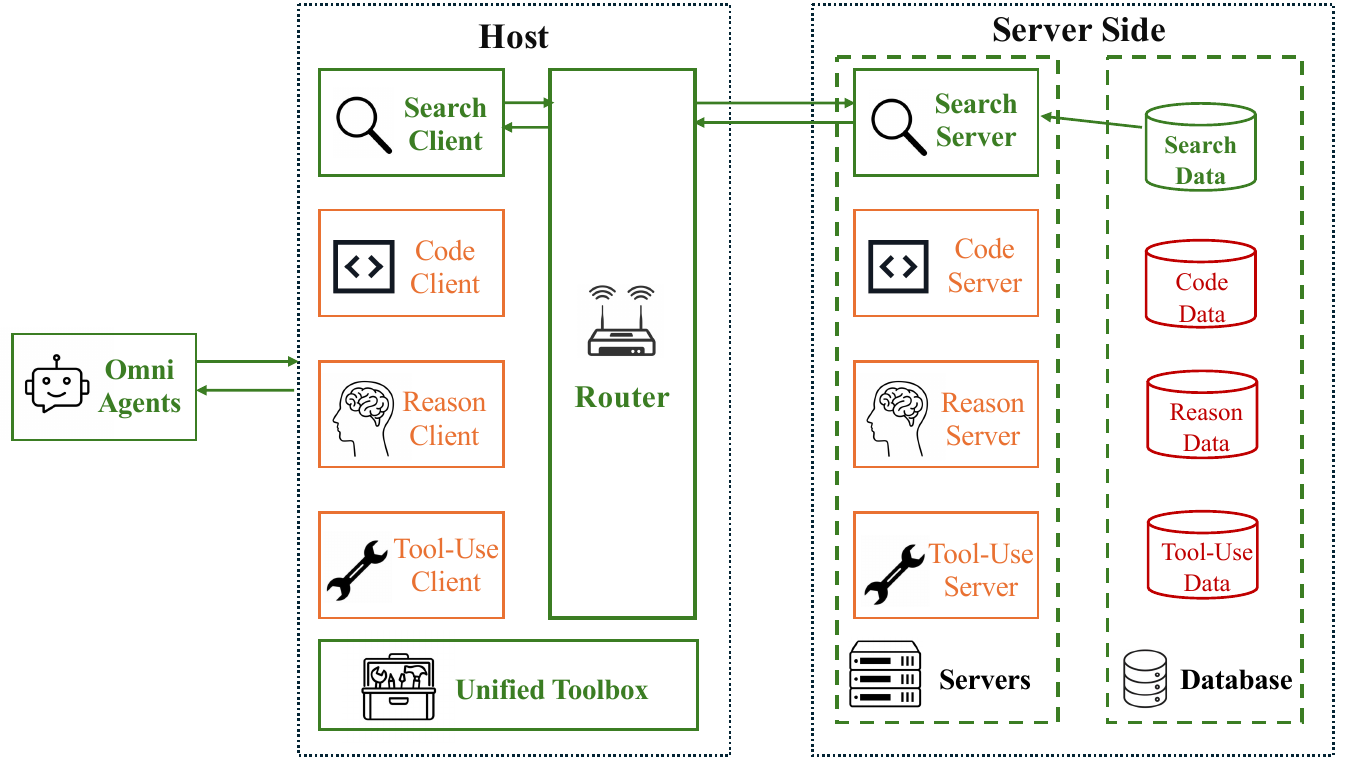}
    \caption{\textbf{Overview of the General AgentBench framework.} Green indicates the active task; orange denotes idle but available servers; red indicates excluded data.}
    \label{fig:omni-setting}
    \vspace{-1em}
\end{wrapfigure}

To support realistic evaluation of general agents, we design a unified framework that exposes all tasks and tools through a shared interaction interface. These choices reflect three fundamental properties of real-world agent usage: cross-domain task diversity, comprehensive skill requirements, and dynamically evolving interactions. An overview of the framework is illustrated in Figure~\ref{fig:omni-setting}.

\paragraph{Unified tool interface.}
In practical deployments, agents must select appropriate tools from a large pool without prior knowledge of task domains. We adopt the Model Context Protocol (MCP)~\citep{mcp} as the backbone of our framework. Each benchmark environment is instantiated as an MCP server, while all servers are centrally managed by a unified \emph{Host}. The \emph{Host} maintains a global tool registry that records all available tools and their corresponding server mappings, presenting the agent with a single, unified tool space across all domains. It serves as the sole interaction interface for the general agent, abstracting away individual benchmark implementations. When the agent invokes a tool, the \emph{Host} resolves the call via the tool registry and routes the request to the appropriate server for execution.

\paragraph{Execution process.}
For each evaluation problem, the framework provides it to the agent together with the unified policy and toolset as context. All benchmark servers (e.g., Docker-based environments in the coding domain) are instantiated simultaneously and remain idle while awaiting requests from the agent. When the agent issues a tool call, the Host routes the request to the corresponding server, executes the tool, and returns the result in a unified response format. Tool execution is decoupled from task type: even for a search-oriented task, code-related tool calls can still be executed, returning valid outputs despite having no direct relevance to the final solution. This design intentionally exposes the agent to a realistic setting where incorrect or irrelevant tool usage remains possible.


\input{tables/main_results.tex}

%% file: tables/main_results.tex
\providecolor{neg}{HTML}{C00000}   
\providecolor{pos}{HTML}{2E8B57}   
\providecolor{baserow}{HTML}{FDF6EC}   
\providecolor{genrow}{HTML}{EBF3FA}    

\providecommand{\dndelta}[1]{\textcolor{neg}{$\blacktriangledown$\,#1\%}}
\providecommand{\updelta}[1]{\textcolor{pos}{$\blacktriangle$\,#1\%}}
\providecommand{\nodelta}[1]{\textcolor{gray}{#1\%}}
\providecommand{\pmv}[1]{{\tiny\,$\pm$\textit{#1}}}
\providecommand{\bc}{\cellcolor{baserow}}
\providecommand{\gc}{\cellcolor{genrow}}

\begin{table*}[t]
\centering
\scriptsize
\caption{Per-benchmark decomposition of \textbf{General AgentBench}. For each model we report performance (mean $\pm$ \textit{std} over $n{=}3$ runs) on every constituent benchmark, grouped by domain; the \colorbox{baserow}{first row} is the Baseline ($B$) setting, the \colorbox{genrow}{second row} is the General ($G$) setting, and the third row shows the relative change ($\Delta\%$) from $B$ to $G$: \textcolor{neg}{red $\blacktriangledown$} denotes degradation, \textcolor{pos}{green $\blacktriangle$} denotes improvement. \textbf{Bold} marks the best and \underline{underline} the second-best score.}
\label{tab:baseline_vs_general}
\setlength{\tabcolsep}{3.0pt}
\renewcommand{\arraystretch}{0.95}
\setlength{\aboverulesep}{0.15ex}
\setlength{\belowrulesep}{0.15ex}
\begin{tabular}{@{} c cc cc c cc c @{}}
\toprule
\multirow{2}{*}{\textbf{Models}}
 & \multicolumn{2}{c}{\textbf{Search}}
 & \multicolumn{2}{c}{\textbf{Code}}
 & \textbf{Reasoning}
 & \multicolumn{2}{c}{\textbf{Tool-Call}}
 & \multirow{2}{*}{\textbf{Avg}} \\
\cmidrule(lr){2-3} \cmidrule(lr){4-5} \cmidrule(lr){6-6} \cmidrule(lr){7-8}
 & BrowseComp & WebVoyager & SWE-Bench & Terminal-Bench & MathHay & Tau2-Bench & MCP-Bench & \\
\midrule
\rowcolor{gray!10} \multicolumn{9}{@{}l}{\textit{Open-Source}} \\
\multirow{3}{*}[-0.5ex]{\shortstack{GPT-OSS\\-120B}}
 & \bc 8.0\pmv{0.17} & \bc 35.9\pmv{0.64} & \bc 15.0\pmv{0.16} & \bc 18.1\pmv{1.26} & \bc 46.7\pmv{0.56} & \bc 63.0\pmv{5.12} & \bc 67.3\pmv{1.63} & \bc 36.6 \\
 & \gc 4.0\pmv{0.25} & \gc 27.7\pmv{0.84} & \gc 12.0\pmv{0.62} & \gc 6.3\pmv{0.35} & \gc 38.7\pmv{0.49} & \gc 26.0\pmv{1.32} & \gc 63.3\pmv{2.92} & \gc 26.1 \\
 & \dndelta{50.0} & \dndelta{22.8} & \dndelta{20.0} & \dndelta{65.2} & \dndelta{17.1} & \dndelta{58.7} & \dndelta{5.9} & \dndelta{28.7} \\
\midrule
\multirow{3}{*}[-0.5ex]{\shortstack{Qwen3-235B\\A22B}}
 & \bc 12.0\pmv{0.28} & \bc 40.2\pmv{2.55} & \bc 30.0\pmv{0.68} & \bc 40.1\pmv{0.78} & \bc 45.3\pmv{0.49} & \bc 50.0\pmv{5.05} & \bc 74.7\pmv{2.55} & \bc 41.5 \\
 & \gc 8.9\pmv{0.47} & \gc 30.8\pmv{0.58} & \gc 20.4\pmv{0.31} & \gc 23.8\pmv{0.91} & \gc 32.0\pmv{0.38} & \gc 38.3\pmv{3.20} & \gc 66.1\pmv{0.64} & \gc 30.9 \\
 & \dndelta{25.8} & \dndelta{23.4} & \dndelta{32.0} & \dndelta{40.6} & \dndelta{29.4} & \dndelta{23.4} & \dndelta{11.5} & \dndelta{25.5} \\
\midrule
\multirow{3}{*}[-0.5ex]{\shortstack{Qwen3\\Next}}
 & \bc 11.0\pmv{1.12} & \bc 25.8\pmv{0.84} & \bc 20.0\pmv{0.54} & \bc 23.7\pmv{1.16} & \bc 44.0\pmv{0.68} & \bc 50.0\pmv{2.96} & \bc 76.1\pmv{3.62} & \bc 36.4 \\
 & \gc 10.5\pmv{0.44} & \gc 35.4\pmv{1.02} & \gc 18.0\pmv{0.73} & \gc 8.8\pmv{0.41} & \gc 42.0\pmv{0.46} & \gc 48.9\pmv{2.36} & \gc 64.6\pmv{0.91} & \gc 32.6 \\
 & \dndelta{4.5} & \updelta{37.2} & \dndelta{10.0} & \dndelta{62.9} & \dndelta{4.5} & \dndelta{2.2} & \dndelta{15.1} & \dndelta{10.4} \\
\midrule
\multirow{3}{*}[-0.5ex]{\shortstack{Deepseek\\V3.2}}
 & \bc\underline{22.0}\pmv{0.49} & \bc\underline{58.6}\pmv{0.75} & \bc 35.0\pmv{0.55} & \bc 30.6\pmv{1.66} & \bc 34.7\pmv{0.48} & \bc\underline{55.0}\pmv{1.87} & \bc 68.3\pmv{1.65} & \bc 40.9 \\
 & \gc 19.4\pmv{0.81} & \gc 46.2\pmv{0.75} & \gc 31.8\pmv{1.04} & \gc 22.2\pmv{1.41} & \gc 33.3\pmv{0.20} & \gc\underline{54.0}\pmv{4.54} & \gc 66.0\pmv{1.98} & \gc 37.0 \\
 & \dndelta{11.8} & \dndelta{21.2} & \dndelta{9.1} & \dndelta{27.5} & \dndelta{4.0} & \dndelta{1.8} & \dndelta{3.4} & \dndelta{9.5} \\
\midrule
\multirow{3}{*}[-0.5ex]{\shortstack{Deepseek\\R1}}
 & \bc 10.0\pmv{0.49} & \bc 29.2\pmv{0.22} & \bc 20.0\pmv{0.51} & \bc 25.0\pmv{0.25} & \bc 47.3\pmv{0.47} & \bc 20.0\pmv{1.13} & \bc 68.4\pmv{0.58} & \bc 32.9 \\
 & \gc 9.7\pmv{0.55} & \gc 43.1\pmv{0.72} & \gc 14.0\pmv{0.50} & \gc 8.8\pmv{0.38} & \gc\underline{46.7}\pmv{0.08} & \gc 17.1\pmv{0.95} & \gc 62.2\pmv{0.62} & \gc 29.7 \\
 & \dndelta{3.0} & \updelta{47.6} & \dndelta{30.0} & \dndelta{64.8} & \dndelta{1.3} & \dndelta{14.5} & \dndelta{9.1} & \dndelta{9.8} \\
\midrule
\rowcolor{gray!10} \multicolumn{9}{@{}l}{\textit{Proprietary}} \\
\multirow{3}{*}[-0.5ex]{\shortstack{Gemini2.5\\Flash}}
 & \bc 10.0\pmv{0.30} & \bc 42.3\pmv{0.29} & \bc 18.0\pmv{0.87} & \bc 37.5\pmv{1.48} & \bc 60.0\pmv{0.06} & \bc 50.0\pmv{2.02} & \bc 81.4\pmv{2.06} & \bc 44.3 \\
 & \gc 6.5\pmv{0.56} & \gc 32.3\pmv{1.14} & \gc 14.0\pmv{0.85} & \gc 20.0\pmv{0.15} & \gc 36.0\pmv{0.47} & \gc 38.3\pmv{3.44} & \gc 66.6\pmv{3.60} & \gc 30.5 \\
 & \dndelta{35.0} & \dndelta{23.6} & \dndelta{22.2} & \dndelta{46.7} & \dndelta{40.0} & \dndelta{23.4} & \dndelta{18.2} & \dndelta{31.2} \\
\midrule
\multirow{3}{*}[-0.5ex]{\shortstack{Gemini2.5\\Pro}}
 & \bc 10.0\pmv{0.37} & \bc 49.0\pmv{0.82} & \bc 26.0\pmv{0.33} & \bc 27.5\pmv{0.79} & \bc\underline{61.3}\pmv{0.33} & \bc 50.0\pmv{3.10} & \bc\underline{81.6}\pmv{1.39} & \bc 44.4 \\
 & \gc 8.9\pmv{0.33} & \gc 46.2\pmv{1.14} & \gc 26.0\pmv{0.66} & \gc 27.5\pmv{1.02} & \gc 24.0\pmv{0.27} & \gc 46.0\pmv{5.07} & \gc 67.2\pmv{2.11} & \gc 32.4 \\
 & \dndelta{11.0} & \dndelta{5.7} & \nodelta{0.0} & \nodelta{0.0} & \dndelta{60.8} & \dndelta{8.0} & \dndelta{17.6} & \dndelta{27.2} \\
\midrule
\multirow{3}{*}[-0.5ex]{\shortstack{Claude\\Haiku 4.5}}
 & \bc 15.0\pmv{0.35} & \bc 40.0\pmv{0.99} & \bc\underline{58.0}\pmv{0.98} & \bc 31.2\pmv{0.10} & \bc 54.7\pmv{0.37} & \bc 46.0\pmv{2.07} & \bc\textbf{82.3}\pmv{2.03} & \bc 46.7 \\
 & \gc 17.7\pmv{0.60} & \gc 47.7\pmv{0.81} & \gc\underline{56.0}\pmv{0.95} & \gc 25.0\pmv{1.15} & \gc 34.7\pmv{0.36} & \gc 44.0\pmv{3.60} & \gc 69.0\pmv{1.59} & \gc 39.1 \\
 & \updelta{18.0} & \updelta{19.3} & \dndelta{3.4} & \dndelta{19.9} & \dndelta{36.6} & \dndelta{4.3} & \dndelta{16.2} & \dndelta{16.3} \\
\midrule
\multirow{3}{*}[-0.5ex]{\shortstack{Claude\\Sonnet 4.5}}
 & \bc 16.2\pmv{1.04} & \bc 45.0\pmv{1.22} & \bc\textbf{55.0}\pmv{0.97} & \bc\underline{45.6}\pmv{1.69} & \bc 32.0\pmv{0.32} & \bc 49.0\pmv{4.36} & \bc\underline{96.1}\pmv{3.83} & \bc\underline{48.4} \\
 & \gc\underline{23.1}\pmv{0.24} & \gc\underline{56.9}\pmv{1.00} & \gc\textbf{54.0}\pmv{0.67} & \gc\textbf{45.0}\pmv{2.02} & \gc 36.0\pmv{0.35} & \gc\textbf{48.0}\pmv{1.76} & \gc\textbf{72.9}\pmv{1.45} & \gc\underline{45.0} \\
 & \updelta{42.6} & \updelta{26.4} & \dndelta{1.8} & \dndelta{1.3} & \updelta{12.5} & \dndelta{2.0} & \dndelta{24.1} & \dndelta{7.0} \\
\midrule
\multirow{3}{*}[-0.5ex]{GPT5}
 & \bc\textbf{45.0}\pmv{1.94} & \bc\textbf{76.4}\pmv{0.63} & \bc 40.0\pmv{1.02} & \bc\textbf{48.8}\pmv{0.44} & \bc\textbf{64.0}\pmv{1.02} & \bc\textbf{60.0}\pmv{5.18} & \bc 95.9\pmv{1.57} & \bc\textbf{60.9} \\
 & \gc\textbf{27.4}\pmv{0.98} & \gc\textbf{61.5}\pmv{1.53} & \gc 36.0\pmv{0.73} & \gc\underline{41.3}\pmv{0.98} & \gc\textbf{64.0}\pmv{0.20} & \gc 32.0\pmv{2.38} & \gc 59.1\pmv{3.42} & \gc\textbf{47.1} \\
 & \dndelta{39.1} & \dndelta{19.5} & \dndelta{10.0} & \dndelta{15.4} & \nodelta{0.0} & \dndelta{46.7} & \dndelta{38.4} & \dndelta{22.7} \\
\bottomrule
\end{tabular}
\end{table*}

%% file: src/3mainresults.tex
\section{Main Results}\label{sec:main_results}

In this section, we report results on General AgentBench ($G$) and compare it against domain-specific evaluations ($B$) to explicitly show the gap between specialized and general LLM agents.

\subsection{Experimental Details}
We evaluate ten frontier language models spanning open-source and proprietary families: Qwen3-235B~\citep{yang2025qwen3}, DeepSeek-R1~\citep{guo2025deepseek}, Qwen3-Next~\citep{qwen3next}, DeepSeek-v3.2~\citep{liu2025deepseekv3.2}, Gemini 2.5 Flash~\citep{comanici2025gemini}, GPT-5~\citep{openai_gpt5}, and Claude Sonnet 4.5~\citep{anthropic2025claude45}, among others. Models are accessed via Amazon Bedrock and the Hugging Face Inference API (URLs in Appendix~\ref{appendix:prompt}). We ensure that each model's native context window exceeds the benchmark's maximum input length.

\subsection{Result analysis}

Table~\ref{tab:baseline_vs_general} summarizes the results across models and domains on General AgentBench. Claude Sonnet 4.5 achieves the strongest overall performance, driven primarily by its tool-use and coding capabilities, while GPT-5 attains the highest scores in the search and reason domains. Among open-source models, DeepSeek-V3.2 outperforms both Gemini variants, demonstrating the significant scaling potential of efficient, sparse-attention architectures~\citep{liu2025deepseekv3.2}. Compared with domain-specific setting, most LLM agents experience substantial degradation in the general-agent setting, with average relative drops ranging from 10\% to 30\%. The magnitude of this degradation varies widely: for example, Gemini 2.5-Pro suffers a drop exceeding 60\% in the reason domain, falling from top-tier performance in the baseline setting to near-average performance as a general agent. In contrast, Claude Sonnet 4.5 remains notably robust, with only a 7.0\% average degradation.

\subsection{Cross-domain tool usage}

Interestingly, several models, including Qwen3-Next, Deepseek-R1, and Claude, exhibit \textbf{performance gains} on search domains. Trajectory-level analysis shows that these improvements arise from effective \textbf{cross-domain tool usage}, where agents repurpose tools beyond their originally intended domains to support information retrieval. Most of these tools originate from MCP-related benchmarks and exhibit a high degree of overlap with the topics in search tasks. This reinforces the  conclusion that tool harness is an independent dimension for agent improvement, and poor design of tools can hinder model performance~\citep{chen2025browsecompplus, li2025mm}. We present a case study demonstrating how cross-domain tools outperform plain web search.


\begin{tcolorbox}[
  colback=white,
  colframe=gray!75,
  title={\small\textbf{Case Study: Retrieving the Latest HF Model}},
  boxrule=0.5pt,
  left=2pt, right=2pt, top=3pt, bottom=3pt,
  sharp corners
]
\fontsize{7pt}{8.5pt}\selectfont
\textbf{Task.} Identify the most recent pre-trained language model on HuggingFace suitable for text classification.

\vspace{3pt}
\noindent
\begin{minipage}[t]{0.47\linewidth}
\begin{tcolorbox}[colback=red!3, colframe=red!40, left=2pt, right=2pt, top=2pt, bottom=2pt, boxrule=0.4pt, title={\scriptsize\textbf{Baseline: Web Search Only (6 turns)}}, fonttitle=\bfseries, colbacktitle=red!10, coltitle=black, sharp corners]
\fontsize{6pt}{7.5pt}\selectfont\ttfamily
T1: <search> latest HF pre-trained\\
\phantom{T1: }language model text-cls 2024</search>\\
T2: <search> HF text-cls models\\
\phantom{T2: }sorted by trending 2024</search>\\
T3: <search> HF newest text-cls\\
\phantom{T3: }models Jan 2025</search>\\
\textrm{...}\\
T6: <answer> tabularisai/multilingual\\
\phantom{T6: }-sentiment-analysis</answer>\\[1pt]
{\sffamily\fontseries{b}\selectfont\color{red!70!black} \texttimes\; Incorrect --- fine-tuned, not pre-trained}
\end{tcolorbox}
\end{minipage}%
\hfill
\begin{minipage}[t]{0.52\linewidth}
\begin{tcolorbox}[colback=blue!3, colframe=blue!40, left=2pt, right=2pt, top=2pt, bottom=2pt, boxrule=0.4pt, title={\scriptsize\textbf{General Agent (3 turns)}}, fonttitle=\bfseries, colbacktitle=blue!10, coltitle=black, sharp corners]
\fontsize{6pt}{7.5pt}\selectfont\ttfamily
T1: HF\_\_search-models(\{"tags":\\
\phantom{T1: }"text-classification","limit":15\})\\
\phantom{T1: }$\rightarrow$ [\{"id":"FutureMa/Qwen3-4B\\
\phantom{T1: }-Evasion"\}, \{"id":"BAAI/bge-\\
\phantom{T1: }reranker-v2-m3"\}, ...]\\
T2: web\_search("ModernBERT HF 2025")\\
\phantom{T2: }$\rightarrow$ "8192-tok, 139M/395M params"\\
T3: HF\_\_get-model-info("answerdotai/\\
\phantom{T3: }ModernBERT-base") $\rightarrow$ Full card\\[1pt]
{\sffamily\fontseries{b}\selectfont\color{blue!70!black} \checkmark\; Correct --- API + web search + verify}
\end{tcolorbox}
\end{minipage}

\end{tcolorbox}

We observe that the plain search baseline iteratively refines web queries across 6 turns, ultimately finding \texttt{\small tabularisai/multilingual-sentiment-analysis} with only surface-level information (``\texttt{\small fine-tuned DistilBERT}''). In contrast, the general agent discovers the \texttt{Hugging\_Face\_\_search-models} tool and directly queries the model hub with structured filters. The subsequent \texttt{Hugging\_Face\_\_get-model-info} call retrieves comprehensive metadata based on the previous turn's information, which is difficult to locate through web search snippets. 

%% file: src/4sequential_tts.tex
\section{The Plateau of Sequential Scaling}

Sequential scaling allocates more test-time compute by extending the interaction turns. As the agent engages with the environment, the conversation grows. When the agent attempts to terminate an episode, we can replace the termination action with an additional round of instruction to encourage further reasoning. This is our main method used in the following experiments. More complex sequential scaling strategies include summarization~\citep{park2023generative}, verifier-guided revision~\citep{zhu2025scaling}, and recent discarding and trajectory folding~\citep{ye2025agentfold}. We conduct ablations on the choice of strategy in Section~\ref{4.3}.

\subsection{Empirical observations}

\begin{figure}[h]
    \centering
    \includegraphics[width=1.0\linewidth]{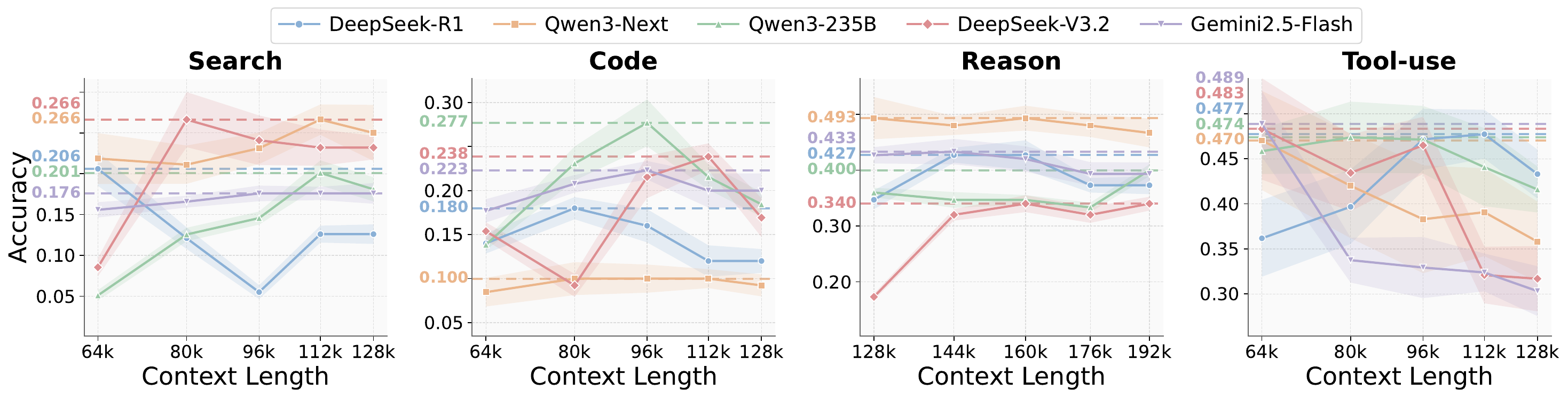}
    \caption{\textbf{Sequential scaling behavior of 5 models over 4 domains on General AgentBench.} The $x$-axis is the context budget and the $y$-axis is accuracy. The shaded area represents the bootstrapped 95\% CI. Performance generally improves at first but saturates or degrades beyond a model- and domain-dependent scaling plateau, sometime even degrades at first, indicating an inherent limitation.}
    \label{fig:sequential_tts}
\end{figure}


We first investigate whether sequential scaling continues to yield meaningful improvement and, if not, when it hits its upper bound. We pre-select several context budgets depending on the task domain. This token budget is defined as the accumulated input context length for the next turn, including all previously generated text and tool results / environmental feedback. For each budget, the agent starts fresh from the question---rather than continuing from a shorter run---and is forced to keep working until reaching the predefined budget. Figure~\ref{fig:sequential_tts} plots each agent's accuracy against the context budget. We observe that (1) \textbf{most agents utilize additional compute effectively at first, but after a certain point performance saturates or even degrades;} (2) \textbf{some agents performance fluctuates or even degrades from the very beginning.}

\begin{figure}[h]
    \centering
    \includegraphics[width=1.0\linewidth]{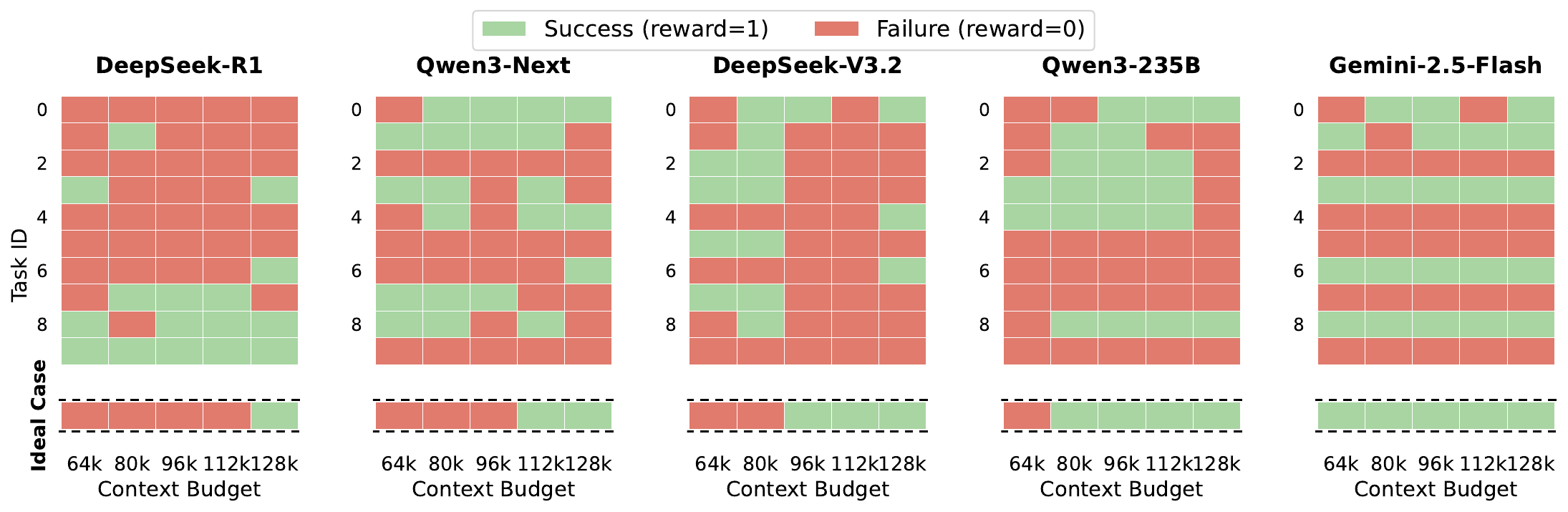}
    \caption{\textbf{Instance-level correctness dynamics.} We randomly sample 10 instances from the General AgentBench and track their correctness across context budgets. Red indicates incorrect cases and green indicates correct cases. The bottom ``ideal case'' illustrates how we expect the model to behave.}
    \label{fig:per_instance}
\end{figure}

To understand this degradation, we examine instance-level trajectories under different context budgets. As shown in Figure~\ref{fig:per_instance}, accuracy on individual instances fluctuates between correct and incorrect across budget levels rather than forming a stable upward trend. We attribute this to two factors. First, agents fail to preserve earlier correct answers as context grows: the accumulated interaction history interferes with the agent's original reasoning, causing instances that were solved at shorter budgets to become incorrect at longer ones. Second, since each evaluation starts from scratch, the inherent non-determinism of agentic execution means the agent may explore a different---and ultimately unsuccessful---reasoning path at a higher budget, even for the same instance~\citep{mehta2026consistency, bjarnason2026randomness}. This non-determinism makes the scaling plateau hard to pinpoint from a single run, motivating a statistically grounded definition in the next section.

\subsection{Formal Definition of Scaling Plateau}\label{sec:plateau_definition}

To formalize the scaling plateau beyond merely empirical observation, we provide a precise definition. We evaluate model $\mathcal{M}$ on $n$ instances sampled uniformly without replacement from a benchmark of $N$ instances, under increasing context budgets $c_1 < c_2 < \cdots < c_K$. We assume that a model's accuracy on instance $b_i$ and context budget $c_k$ follows the Bernoulli distribution $B(p_{i,k})$, supported by the inherent randomness of evaluation~\citep{bjarnason2026randomness}. The population-level accuracy is $\mu_k = \frac{1}{N} \sum_{i=1}^{N} p_{i,k}$. During our evaluation, each instance $b_i$ is evaluated $T$ times at each context budget, yielding an estimated solve probability $\hat{p}_{i,k}$. When $T=1$, this reduces to a single binary outcome $x_{i,k} \in \{0, 1\}$. 

\paragraph{Paired Difference Test.}
Since the same instances are evaluated across context budgets, we define the per-instance gain:
\[
d_i^{(k)} = \hat{p}_{i,k+1} - \hat{p}_{i,k}
\]
and the average gain $\bar{d}^{(k)} = \frac{1}{n} \sum_{i=1}^{n} d_i^{(k)}$, which is an unbiased estimator of the population-level marginal gain $\Delta_k = \mu_{k+1} - \mu_k$.
We construct a one-sided $(1-\alpha)$ confidence lower bound for $\Delta_k$ via the paired bootstrap (resample $\{d_1^{(k)}, \dots, d_n^{(k)}\}$ with replacement $B$ times) with a finite population correction:
\[
C_k = \bar{d}^{(k)} - z_{1-\alpha} \cdot \hat{\sigma}_{\text{boot}} \cdot \sqrt{\frac{N - n}{N - 1}}
\]
where $z_{1-\alpha}$ is the standard normal quantile and the last factor corrects for without-replacement sampling (see Appendix~\ref{appendix:derivation} for detailed derivation).

\paragraph{Scaling Plateau Definition.}

We define the \textit{scaling plateau} at significance level $\alpha$ as:
\[
k^* = \max\{k : C_k > 0\} + 1
\]
That is, $k^*$ is the index such that $c_{k^*}$ is the first context budget at which scaling no longer yields a statistically significant improvement. If no transition is significant ($C_k \leq 0$ for all $k$), we set $k^* = 1$, indicating the model does not benefit from any scaling. If $C_k > 0$ for all $k \in \{1, \dots, K-1\}$, the scaling plateau lies beyond the measured range. Note that $k^*$ depends on the granularity of the chosen context budgets $\{c_k\}$: finer spacing yields a more precise estimate of where the scaling plateau occurs.

\begin{figure*}[t]
    \centering
    \begin{subfigure}[t]{0.48\linewidth}
        \centering
        \includegraphics[width=\linewidth]{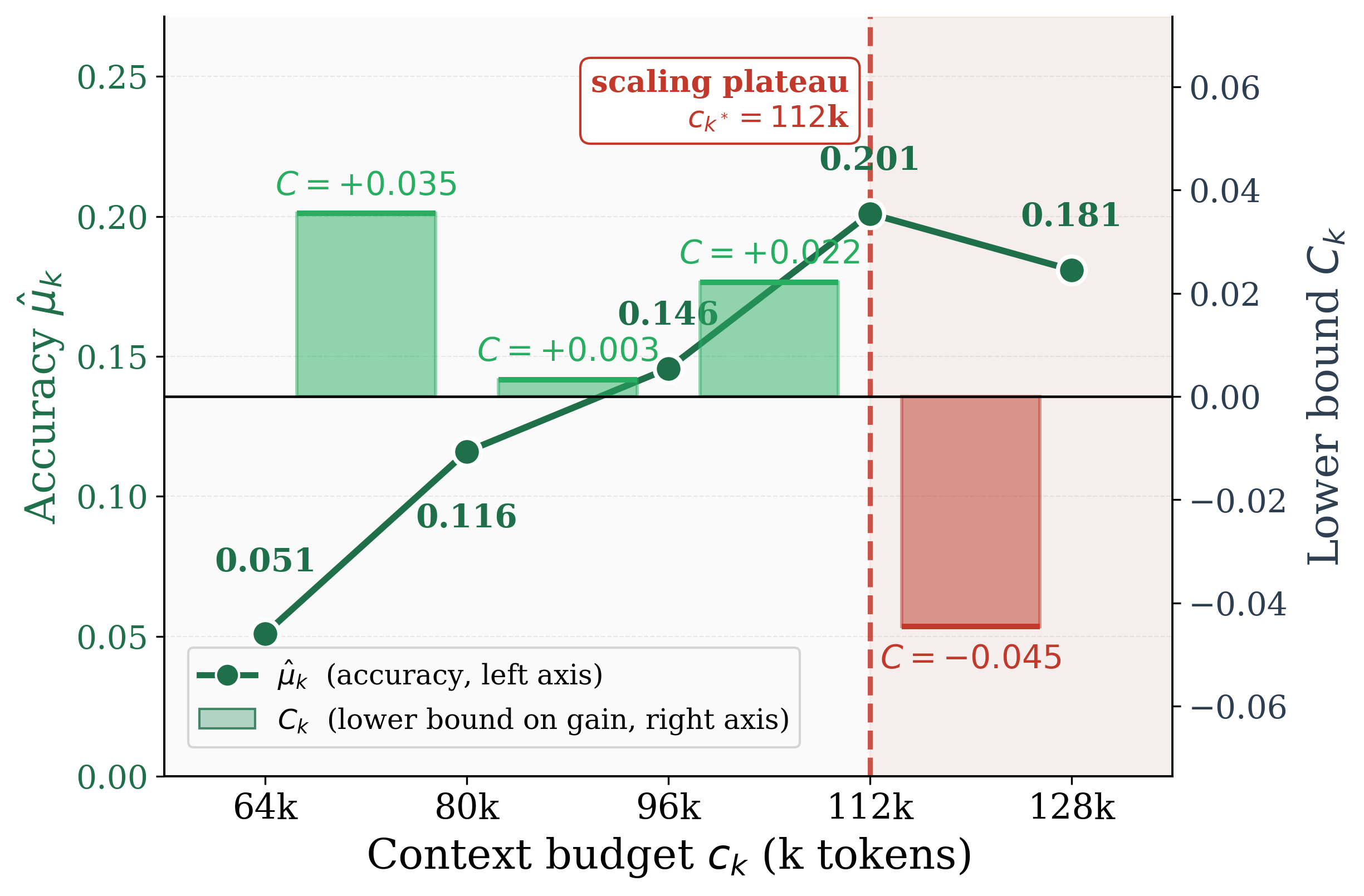}
        \caption{Accuracy curve and Lower Bound Changes by the context budget.}
        \label{fig:walkthrough_a}
    \end{subfigure}
    \hfill
    \begin{subfigure}[t]{0.48\linewidth}
        \centering
        \includegraphics[width=\linewidth]{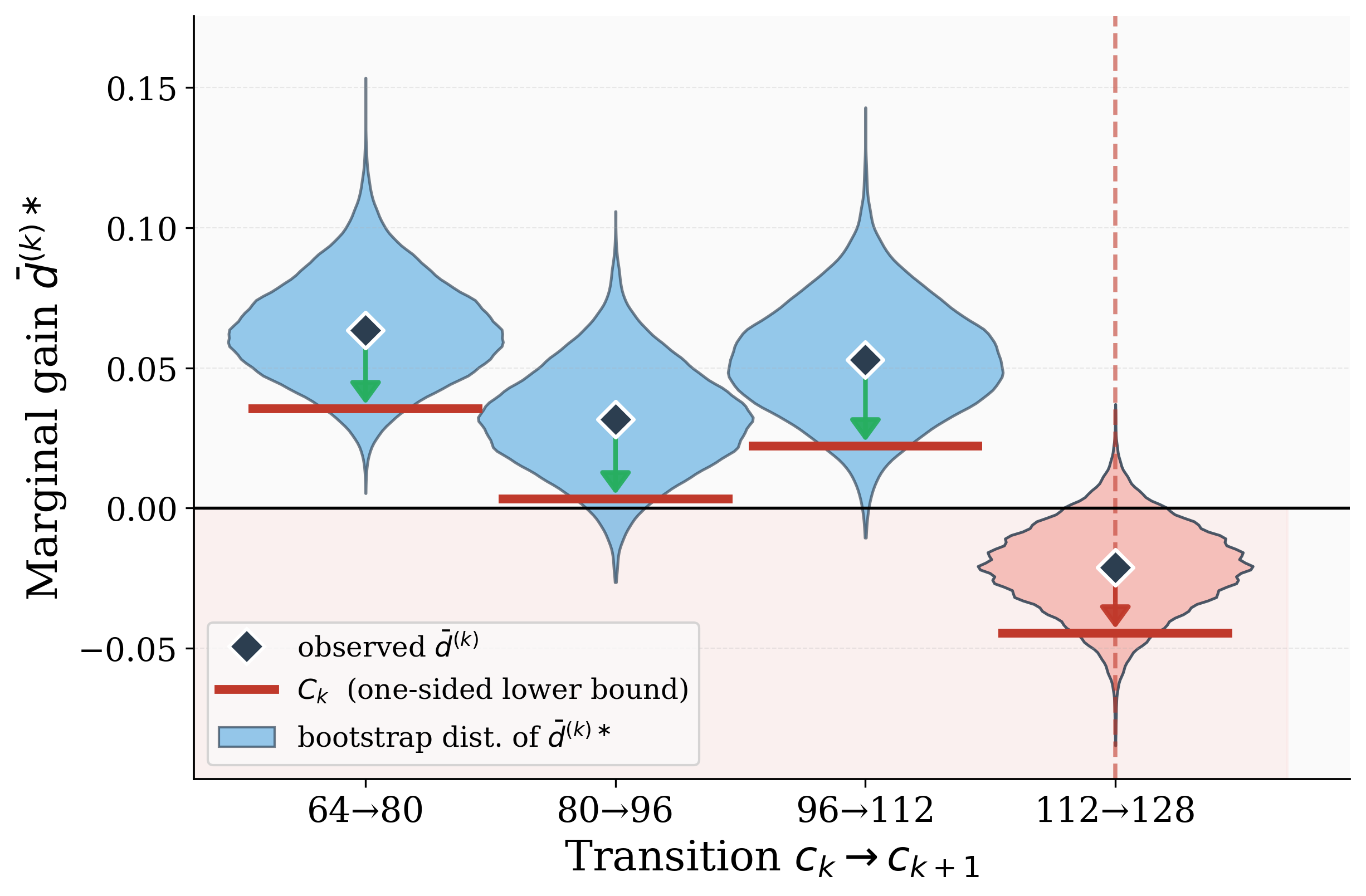}
        \caption{Demonstrate when the confidence lower bound is strictly less than 0.}
        \label{fig:walkthrough_b}
    \end{subfigure}
    \caption{\textbf{Empirical proof of the scaling plateau definition} on Qwen3-235B in the Search domain. }
    \label{fig:walkthrough}
\end{figure*}

We run the above procedure on Qwen3-235B in the Search domain ($N=1909$, $n=189$, $B=10000$, $\alpha=0.05$, $T=1$), over context budgets  $c_k \in \{64, 80, 96, 112, 128\}k$ where the model exhibits non-trivial performance. Figure~\ref{fig:walkthrough} traces the computation end-to-end. The first three transitions all produce positive lower bounds $C_k > 0$, confirming statistically significant gains---though the margins shrink progressively. At the fourth transition $112$k$\to128$k, the lower bound turns negative ($C_4 = -0.045$), triggering the $\min$ rule at $k^*=4$. This gives $c_{k^*}=112$k as the scaling plateau---the largest budget at which sequential scaling on Search still delivers a certified return for Qwen3-235B. More scaling plateau analyses are listed in Appendix~\ref{appendix:scaling_plateau}.    

\subsection{Ablation on Scaling Strategy}\label{4.3}

\begin{figure}[h]
    \centering
    \includegraphics[width=\linewidth]{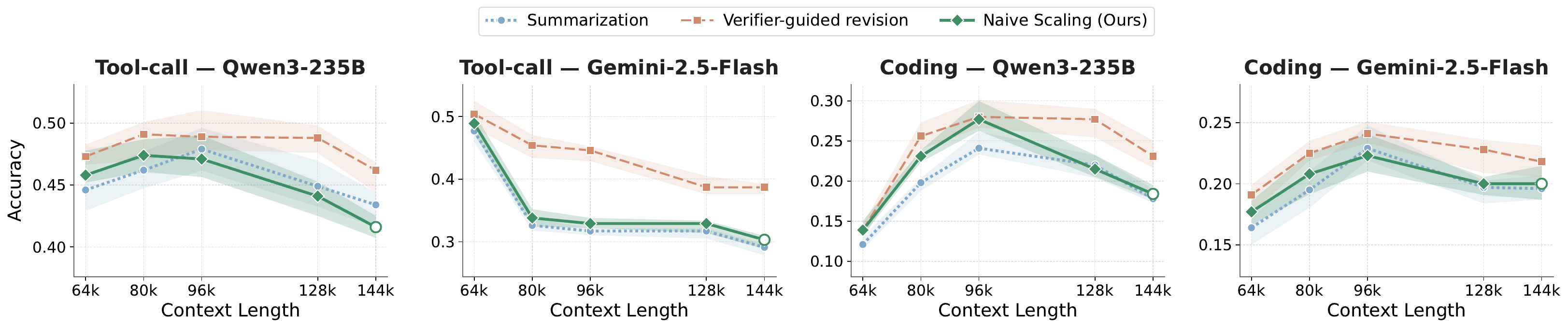}
    \caption{\textbf{Ablation studies on sequential scaling strategies.}}
    \label{fig:ablation_strategies}
\end{figure}

A natural question is whether the scaling plateau is an artifact of our default strategy (continue-thinking prompting) or a more general phenomenon. We compare two stronger alternatives on Qwen3-235B and Gemini-2.5-Flash across the coding and tool-call domains:

\textbf{(1) Summarization.} Before each new turn, the agent summarizes the preceding context and explicitly states its current progress, reducing redundancy in the interaction history.

\textbf{(2) Verifier-guided revision.} The agent receives verification from a stronger teacher model (GPT-5) and revises its approach accordingly.


As shown in Figure~\ref{fig:ablation_strategies}, all three strategies exhibit a similar pattern: performance gradually saturates or even degrades beyond a certain budget, regardless of feedback sophistication. Verifier-guided revision, despite leveraging the strongest available signal, achieves higher absolute scores---most notably on coding---yet does not escape the degradation regime. These results confirm that stronger strategies uniformly raise the scaling curve without altering its shape, supporting the conclusion that the scaling plateau reflects an inherent limitation of sequential scaling rather than a weakness of any particular strategy.

%% file: src/4parallel_tts.tex
\begin{figure*}[t]
    \centering
    \includegraphics[width=0.98\linewidth]{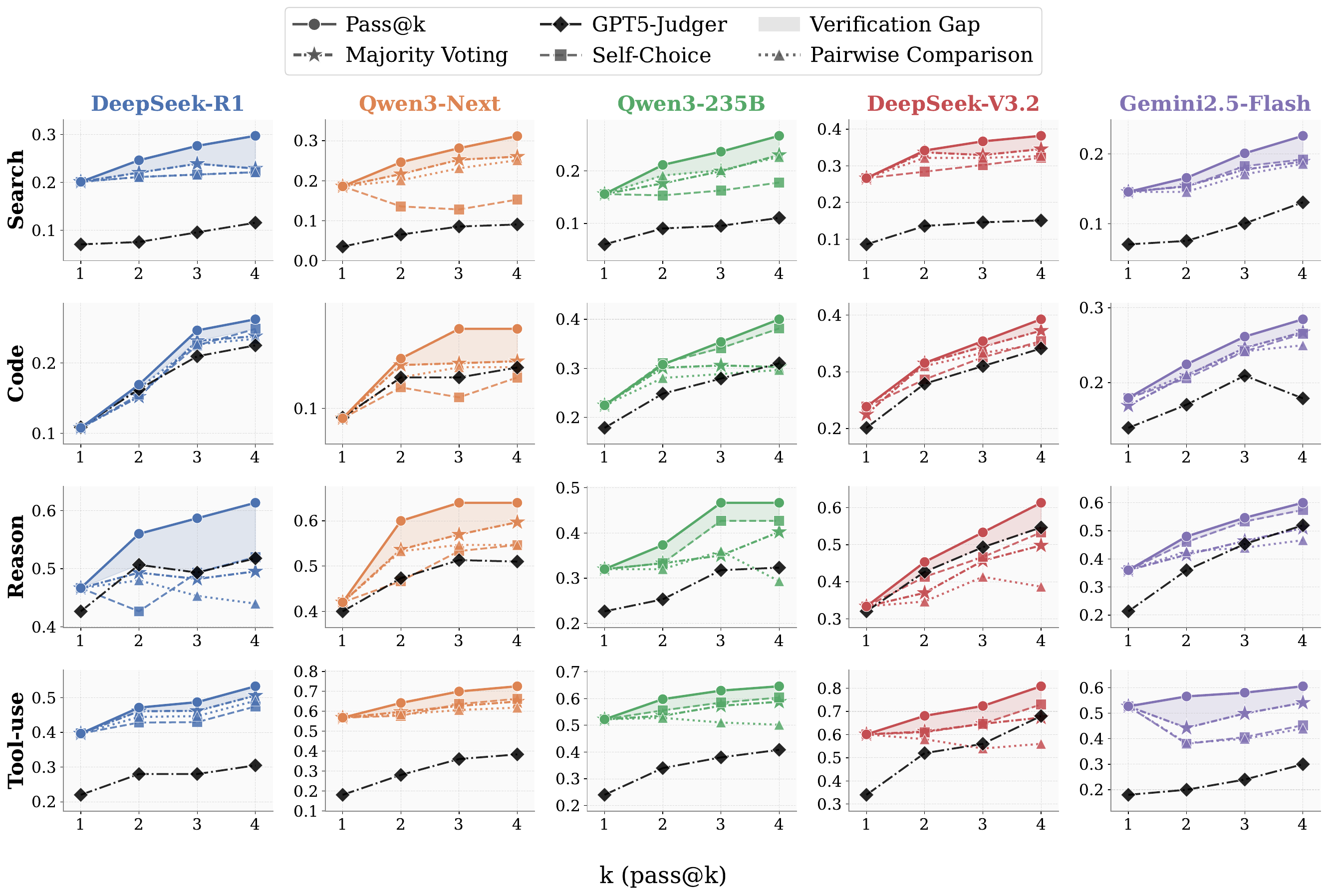}
    \caption{\textbf{Verification gap between generation and self-choice.}
Across four domains, we observe a consistent gap between solution generation and verification: as the number of samples increases, correct solutions appear more frequently in the sampled set, yet models often fail to identify and select them. The dashed and dotted curves represent two self-choice strategies, while the diamond denotes a stronger evaluator, GPT-5.}
    \label{fig:parallel_scaling}
\end{figure*}

\section{Parallel Scaling}

Parallel scaling is one of the most common test-time scaling strategies due to its simplicity---we only need to independently sample $K$ trajectories for each query. Increasing $K$ expands the reachable action space, thereby increasing the likelihood that the agent explores at least one trajectory containing a correct solution. 

However, in the absence of external oracles or human feedback—particularly in real-world deployments—parallel scaling alone is insufficient: agents must also be capable of evaluating and selecting the best outcome from their own generated trajectories, i.e., commit to one answer. We assess this ability by comparing pass@$K$ with three different commitment strategies:

\textbf{(1) Self-choice.}  The agent independently evaluates each sampled trajectory and assigns a binary judgment. Performance is measured by the alignment between the model’s judgments and oracle labels, averaged over trajectories that are correct under oracle evaluation.

\textbf{(2) Pairwise comparison.}  
The agent compares two sampled trajectories at a time and iteratively promotes the superior one through a bubble-sort-style selection process. After $K-1$ pairwise comparisons, a single trajectory is selected as the final output, and performance is evaluated based on the correctness of this selected trajectory.

\textbf{(3) Majority voting.} Choose the answer that receives the most agreement over all sampled answers.

These three strategies reflect the practical effectiveness of parallel scaling, while pass@$K$ serves as an upper bound that reveals the solution potential.

The pass@K growing trend alongside the three commitment strategy are shown in Figure~\ref{fig:parallel_scaling}. We observe a monotonic increase in \textbf{pass@K} as the number of sampled trajectories grows. Increasing $K$ from 1 to 4 yields an average improvement of roughly 50\%. However, pass@K only reflects an idealized upper bound. As shown by the gap between the solid lines (pass@K) and the dashed/dotted lines, all commitement strategies' performance consistently lags behind the pass@K. In some cases, their performance even degrades as $K$ increases.

To examine whether this gap stems from limited verifier capability, we also let a strong teacher model (GPT-5) severing as the external judger.  As shown by the black diamond line, GPT-5 generally underperforms models’ own self-judgment. Notably, even at $K=1$, GPT-5 occasionally misclassifies correct trajectories as incorrect, introducing a non-trivial gap between pass@1 and GPT-5–based verification. We hypothesize that this effect arises from \textbf{solution familiarity}: models are better at evaluating their own generations, which align closely with their internal reasoning patterns, whereas external verifiers may struggle to accurately assess unfamiliar execution traces. Overall, these results reveal a fundamental gap between models’ generation capacity and their ability to reliably select correct solutions, which ultimately limits the practical utility of parallel scaling. We include the results on higher $K$ along with additional prompt sensitivity experiments in Appendix~\ref{appendix:k816} and~\ref{appendix:prompt_sensitivity}.

%% file: src/related_work.tex
\section{Related Work}

Prior work on LLM agents spans benchmarking, agent development, and inference-time scaling. Most agentic benchmarks target domain-specific settings: software engineering (SWE-bench~\citep{jimenez2023swebench} and variants~\citep{Zhang2025SWEBenchGoesLive,yang2024sweagent}), web navigation (Mind2Web~\citep{Deng2023Mind2Web}, WebArena~\citep{zhou2023webarena}, WebVoyager~\citep{he2024webvoyager}), and tool use with curated APIs (ToolLLM~\citep{Qin2023ToolLLM}, BFCL~\citep{patil2025bfcl}). Broader suites such as $\tau$-bench~\citep{Yao2024TauBench}, AgentBench~\citep{barres2025tau2}, and GAIA~\citep{Mialon2023GAIA} incorporate multi-turn interaction across task categories, while OSWorld~\citep{xie2024osworld} and AndroidWorld~\citep{rawles2024androidworld} evaluate agents in real OS environments. On the agent-building side, early methods such as ReAct~\citep{yao2022react}, Reflexion~\citep{shinn2023reflexion}, and Toolformer~\citep{schick2023toolformer} introduce structured reasoning--action loops, reflection, and learned tool invocation, while recent frameworks support multi-agent coordination and rich tool ecosystems~\citep{Xie2023OpenAgents,gao2024agentscope,wang2025opencua}. A complementary line of research studies test-time scaling (TTS), allocating additional inference-time computation to improve performance. Chain-of-thought prompting~\citep{Wei2022CoT} and self-consistency~\citep{Wang2022SelfConsistency} show that deeper reasoning and multi-sample decoding boost accuracy. More explicit approaches search or refine over solution paths, including Tree-of-Thoughts~\citep{Yao2023ToT}, MCTS-based planning~\citep{Zhou2023LATS}, and iterative refinement~\citep{shinn2023reflexion,Madaan2023SelfRefine}. More recently, ``internal scaling’’ trains models to autonomously allocate inference computation~\citep{guo2025deepseek, openai2024learning-to-reason}, while verifier-based methods augment search with learned ranking or process-level supervision~\citep{Lightman2023VerifyStepByStep, hosseini2024v, wang2024math}.


%% file: src/6conclusions.tex
\section{Conclusions}\label{sec:conclusions}
We systematically investigate the dynamic behaviors of test-time scaling in real-world deployment through two complementary axes to answer the question: can a model continuously benefit from additional compute? We first present General AgentBench, a realistic benchmark for evaluating LLM agents under unified framework. Across ten leading models, we find a substantial robustness gap when moving from domain-specific to general-agent evaluation. Our test-time scaling analysis reveals two fundamental limits: sequential scaling is bounded by a scaling plateau, and parallel scaling is limited due to a persistent verification gap between generation and commitment. We provide a precise statistical definition of the scaling plateau and verify different scaling strategies to demonstrate the pervasiveness of this problem. We do not include recent compression-based scaling strategies~\citep{ye2025agentfold, lu2025scaling}, as they shrink the context size rather than expand inference compute, which falls outside our setting. We leave this direction for future work.

%% file: src/7appendix.tex
\input{src/appendix_derivation.tex}

\newpage

\input{src/appendix_details_benchmark}

\newpage

\input{src/appendix_cost}

\newpage

\input{src/appendix_detailed_numerical_results}

\newpage

\input{src/appendix_attention_analysis_only}

\input{src/appendix_details_our_benchmark}


\input{src/appendix_prompts}

%% file: src/appendix_derivation.tex
\section{Theoretical Details for the Scaling Plateau}\label{appendix:derivation}

For convenience, we briefly recap the notation and definitions introduced in the main text. We evaluate model $\mathcal{M}$ on $n$ instances sampled uniformly without replacement from a benchmark of $N$ instances, under increasing context budgets $c_1 < c_2 < \cdots < c_K$. Each instance $b_i$ is evaluated $T$ times at each budget, yielding an estimated solve probability $\hat{p}_{i,k}$ (reducing to a single binary outcome $x_{i,k} \in \{0,1\}$ when $T{=}1$). The per-instance gain is $d_i^{(k)} = \hat{p}_{i,k+1} - \hat{p}_{i,k}$, with average $\bar{d}^{(k)} = \frac{1}{n}\sum_{i=1}^{n} d_i^{(k)}$ serving as an unbiased estimator of the population-level marginal gain $\Delta_k$. The one-sided $(1{-}\alpha)$ confidence lower bound is:
\[
C_k = \bar{d}^{(k)} - z_{1-\alpha} \cdot \hat{\sigma}_{\mathrm{boot}} \cdot \sqrt{\frac{N - n}{N - 1}}
\]
and the \emph{scaling plateau} is defined as $k^* = \max\{k : C_k > 0\} + 1$ (or $k^*{=}1$ if no $C_k > 0$). We now provide the theoretical justifications underlying this procedure.

\subsection{Statistical Model}

Let $p_{j,k} \in [0,1]$ denote the true solve probability of instance $b_j$ under context budget $c_k$. We assume (i) that the $T$ trials for each instance--budget pair are i.i.d.\ $\text{Bernoulli}(p_{j,k})$, and (ii) that outcomes are independent across instances. The quantity of interest is the population marginal gain

\[
\Delta_k = \frac{1}{N}\sum_{j=1}^{N}\bigl(p_{j,k+1} - p_{j,k}\bigr).
\]

\subsection{Paired Estimation}

A subset of $n$ instances is drawn uniformly without replacement from the $N$ available. For each sampled instance $b_i$, we observe the estimated difference $d_i^{(k)} = \hat{p}_{i,k+1} - \hat{p}_{i,k}$. In the single-trial setting ($T=1$), this takes values in $\{-1, 0, +1\}$. The sample mean $\bar{d}^{(k)} = \frac{1}{n}\sum_{i=1}^{n} d_i^{(k)}$ is unbiased for $\Delta_k$.

\subsection{Asymptotic Normality}

The differences $\{d_i^{(k)}\}_{i=1}^{n}$ are independent (by assumption (ii)) and bounded in $[-1, 1]$, so the Lindeberg condition holds. By the finite population central limit theorem,

\[
\frac{\bar{d}^{(k)} - \Delta_k}{\sqrt{\text{Var}(\bar{d}^{(k)})}} \xrightarrow{d} \mathcal{N}(0, 1)
\]

as $n \to \infty$ with $n/N \to \lambda \in (0,1)$.

\subsection{Finite Population Correction}

Under with-replacement sampling, $\text{Var}(\bar{d}^{(k)})$ would equal $\sigma_d^2/n$, where $\sigma_d^2$ is the population variance of the per-instance differences. Without-replacement sampling reduces this by a factor $(N-n)/(N-1)$:

\[
\text{Var}_{\text{wor}}(\bar{d}^{(k)}) = \frac{\sigma_d^2}{n} \cdot \frac{N - n}{N - 1}.
\]

For $n/N = 0.1$ this amounts to a 5\% reduction in the standard error; for $n/N = 0.5$ the reduction reaches 29\%.

\subsection{Bootstrap Variance Estimation}

The population variance $\sigma_d^2$ is unknown. We estimate $\text{Var}(\bar{d}^{(k)})$ via the nonparametric bootstrap: for $b = 1, \dots, B$, draw $n$ values with replacement from $\{d_1^{(k)}, \dots, d_n^{(k)}\}$ and compute the resample mean $\bar{d}_b^*$. The bootstrap variance

\[
\hat{\sigma}_{\text{boot}}^2 = \frac{1}{B-1}\sum_{b=1}^{B}\bigl(\bar{d}_b^* - \overline{\bar{d}^*}\bigr)^2
\]

consistently estimates $\sigma_d^2 / n$, the with-replacement variance. Since our actual sampling is without replacement, we multiply by the finite population correction to obtain the corrected standard error:

\[
\widehat{\text{SE}} = \hat{\sigma}_{\text{boot}} \cdot \sqrt{\frac{N - n}{N - 1}}.
\]

\subsection{Construction of $C_k$}

Combining the asymptotic normality of $\bar{d}^{(k)}$ with the variance estimate, a one-sided $(1-\alpha)$ confidence lower bound for $\Delta_k$ is given by

\[
C_k = \bar{d}^{(k)} - z_{1-\alpha} \cdot \hat{\sigma}_{\text{boot}} \cdot \sqrt{\frac{N - n}{N - 1}}
\]

where $z_{1-\alpha}$ denotes the $(1-\alpha)$-quantile of the standard normal distribution. By construction, $P(\Delta_k \geq C_k) \to 1 - \alpha$ as $n \to \infty$.

\subsection{Variance Decomposition}

The total variance $\sigma_d^2$ conflates two distinct sources of variability. The first is \emph{sampling variance}, arising from the random selection of instances: different instances contribute different true gains $\delta_j = p_{j,k+1} - p_{j,k}$, whose heterogeneity across the population inflates the variance of $\bar{d}^{(k)}$. The second is \emph{inference variance}, arising from stochastic decoding: repeated evaluation of the same instance under the same context budget may yield different binary outcomes, contributing $\frac{1}{T}[p_{i,k+1}(1-p_{i,k+1}) + p_{i,k}(1-p_{i,k})]$ to the variance of each $d_i^{(k)}$.

When $T=1$, these two components are not separately identifiable from the observed data---each $d_i \in \{-1,0,+1\}$ reflects both. The bootstrap directly targets the total variance, which is the relevant quantity for constructing $C_k$. As $T \to \infty$, inference variance vanishes and the bootstrap estimates pure sampling variance.

\subsection{Multiple Comparisons}

The definition $k^* = \max\{k: C_k > 0\} + 1$ entails testing all $K-1$ transitions, introducing a multiple comparisons concern. To control the family-wise error rate, we apply a Bonferroni correction, replacing $\alpha$ with $\alpha' = \alpha/(K-1)$ in each individual test. Since $K$ is typically small (5--10 context budgets), the resulting adjustment is modest: for $K-1=5$ and $\alpha=0.05$, the per-test significance level becomes $\alpha'=0.01$, corresponding to $z_{0.99} = 2.326$ in place of $z_{0.95} = 1.645$.

%% file: src/appendix_details_benchmark.tex
\section{Additional Experimental Results}

\subsection{Full Results on the Scaling Plateau Analysis}\label{appendix:scaling_plateau}

Following the formal definition of the scaling plateau introduced in Section~4.2 and the worked example on Qwen3-235B in the Search domain (Figure~\ref{fig:walkthrough}), we apply the same paired bootstrap procedure to all five evaluated models---DeepSeek-R1, DeepSeek-V3.2, Gemini-2.5-Flash, Qwen3-235B, and Qwen3-Next---across all four domains (Code, Reason, Search, and Tool-use). For each model--domain pair, we compute the one-sided $(1{-}\alpha)$ confidence lower bound $C_k$ at every consecutive context-budget transition and identify the scaling plateau $c_{k^*}$ as defined by $k^* = \max\{k : C_k > 0\} + 1$.

Figure~\ref{fig:plateau_accuracy} presents the accuracy curves overlaid with $C_k$ bar charts for each model--domain combination. Green bars indicate transitions where $C_k > 0$ (statistically significant improvement), while red bars indicate $C_k \leq 0$ (no significant gain or degradation). The dashed vertical line marks the identified scaling plateau $c_{k^*}$. Figure~\ref{fig:plateau_bootstrap} shows the corresponding paired-bootstrap distributions as violin plots, where each violin depicts the resampled distribution of the marginal gain $\bar{d}^{(k)*}$, the diamond marks the observed $\bar{d}^{(k)}$, and the red tick indicates the FPC-corrected lower bound $C_k$. Transitions whose violins lie entirely above zero correspond to confident gains; those overlapping or falling below zero indicate stagnation or decline.

\newcommand{\plateaucolhead}[1]{\begin{minipage}{0.24\linewidth}\centering\textbf{\small #1}\end{minipage}}

\begin{figure}[t]
    \centering
    \plateaucolhead{Code}\hfill\plateaucolhead{Reason}\hfill\plateaucolhead{Search}\hfill\plateaucolhead{Tool-use}\par\vspace{4pt}
    \noindent\textbf{\small DeepSeek-R1}\par\vspace{1pt}%
    \begin{minipage}{0.24\linewidth}\centering\includegraphics[width=\linewidth]{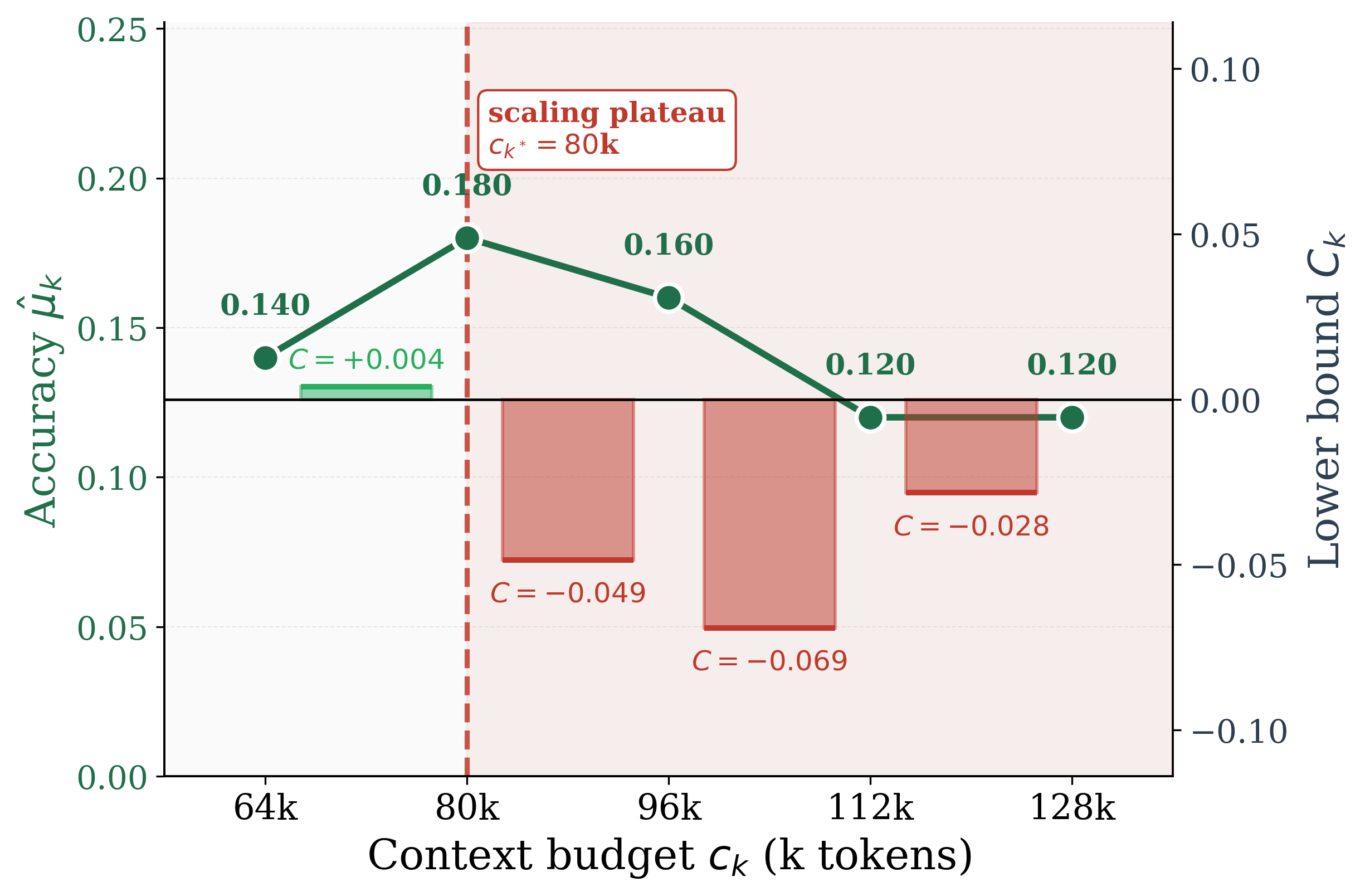}\end{minipage}\hfill
    \begin{minipage}{0.24\linewidth}\centering\includegraphics[width=\linewidth]{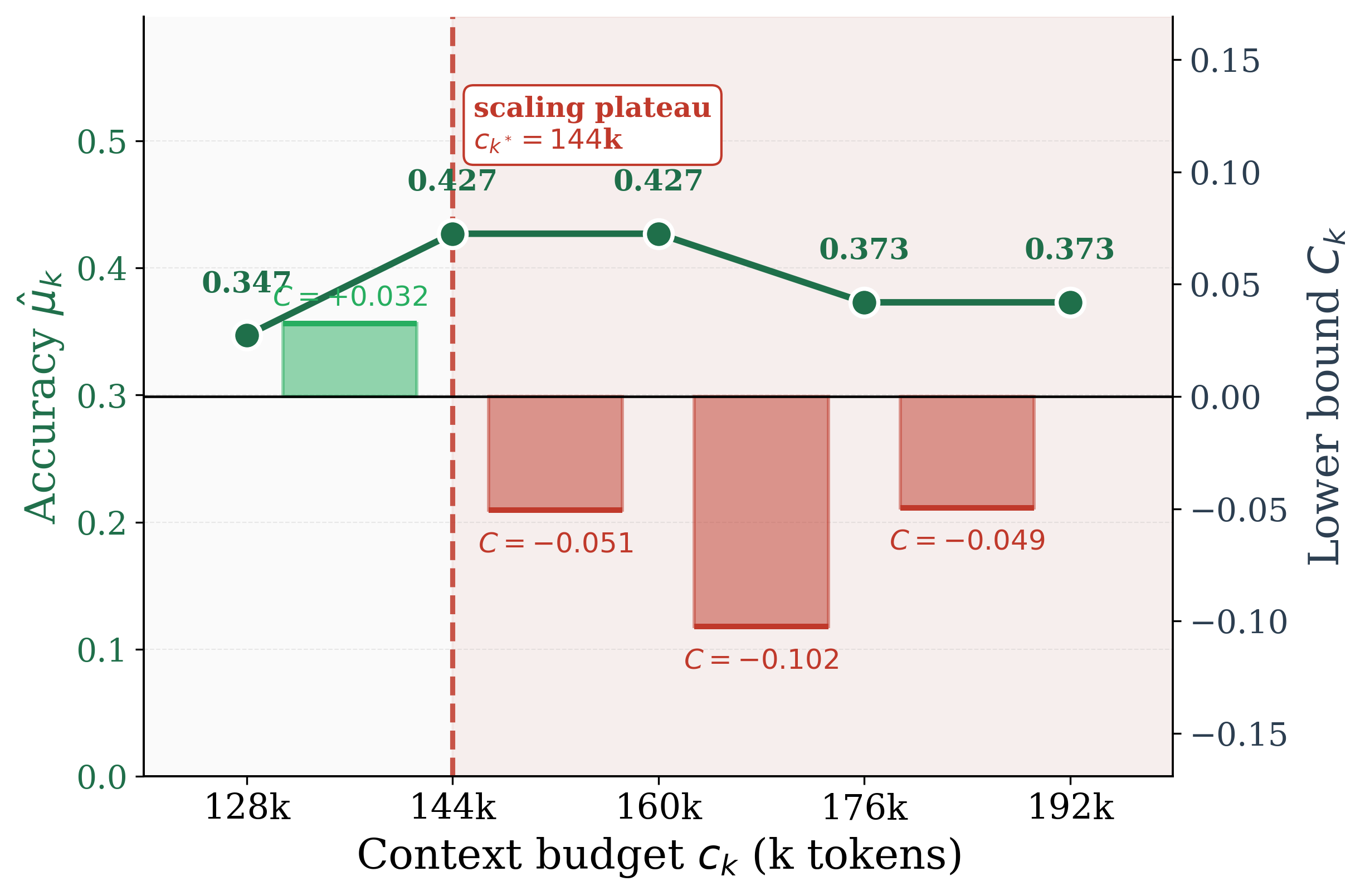}\end{minipage}\hfill
    \begin{minipage}{0.24\linewidth}\centering\includegraphics[width=\linewidth]{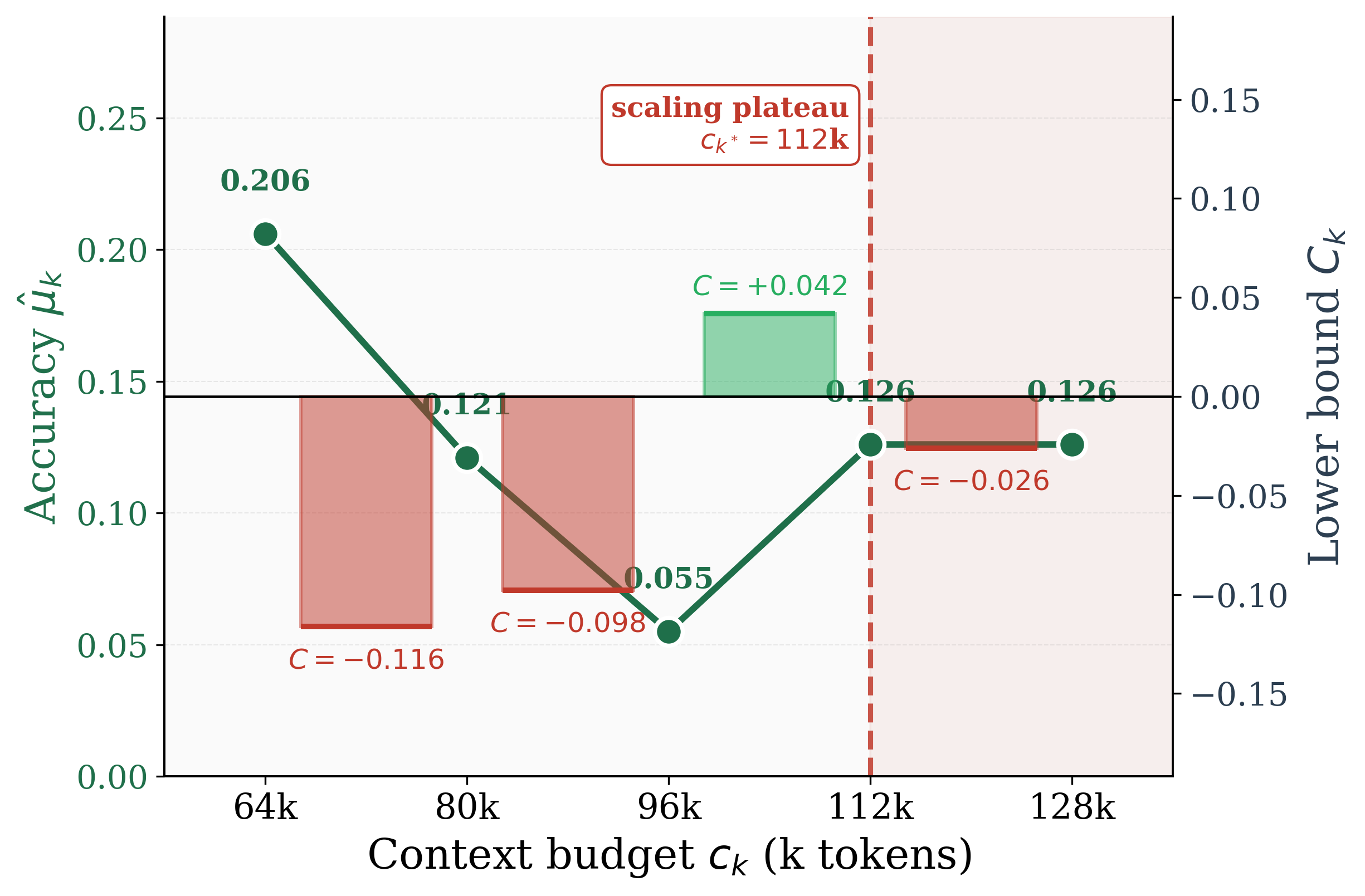}\end{minipage}\hfill
    \begin{minipage}{0.24\linewidth}\centering\includegraphics[width=\linewidth]{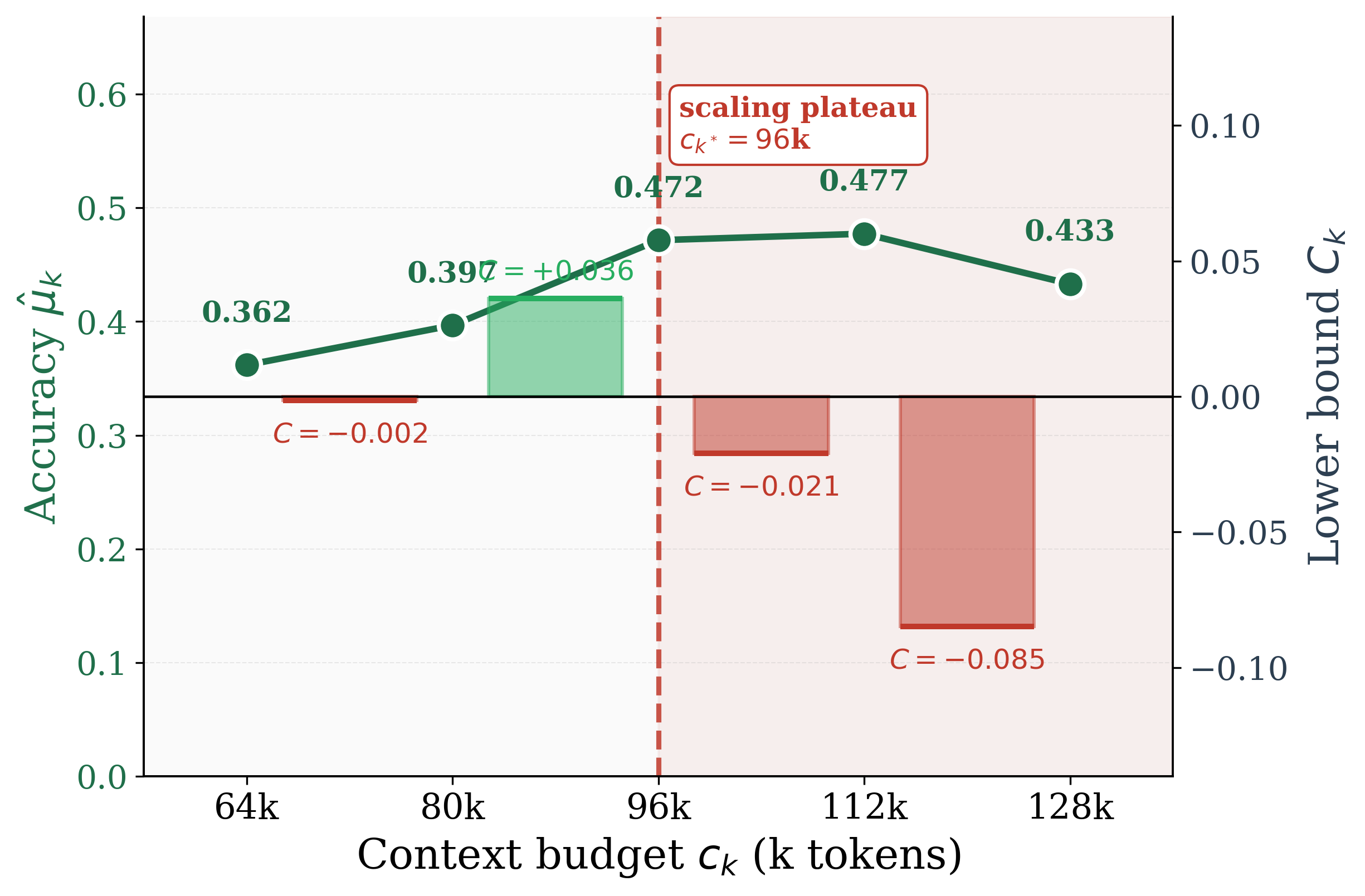}\end{minipage}
    \vspace{4pt}
    \noindent\textbf{\small DeepSeek-V3.2}\par\vspace{1pt}%
    \begin{minipage}{0.24\linewidth}\centering\includegraphics[width=\linewidth]{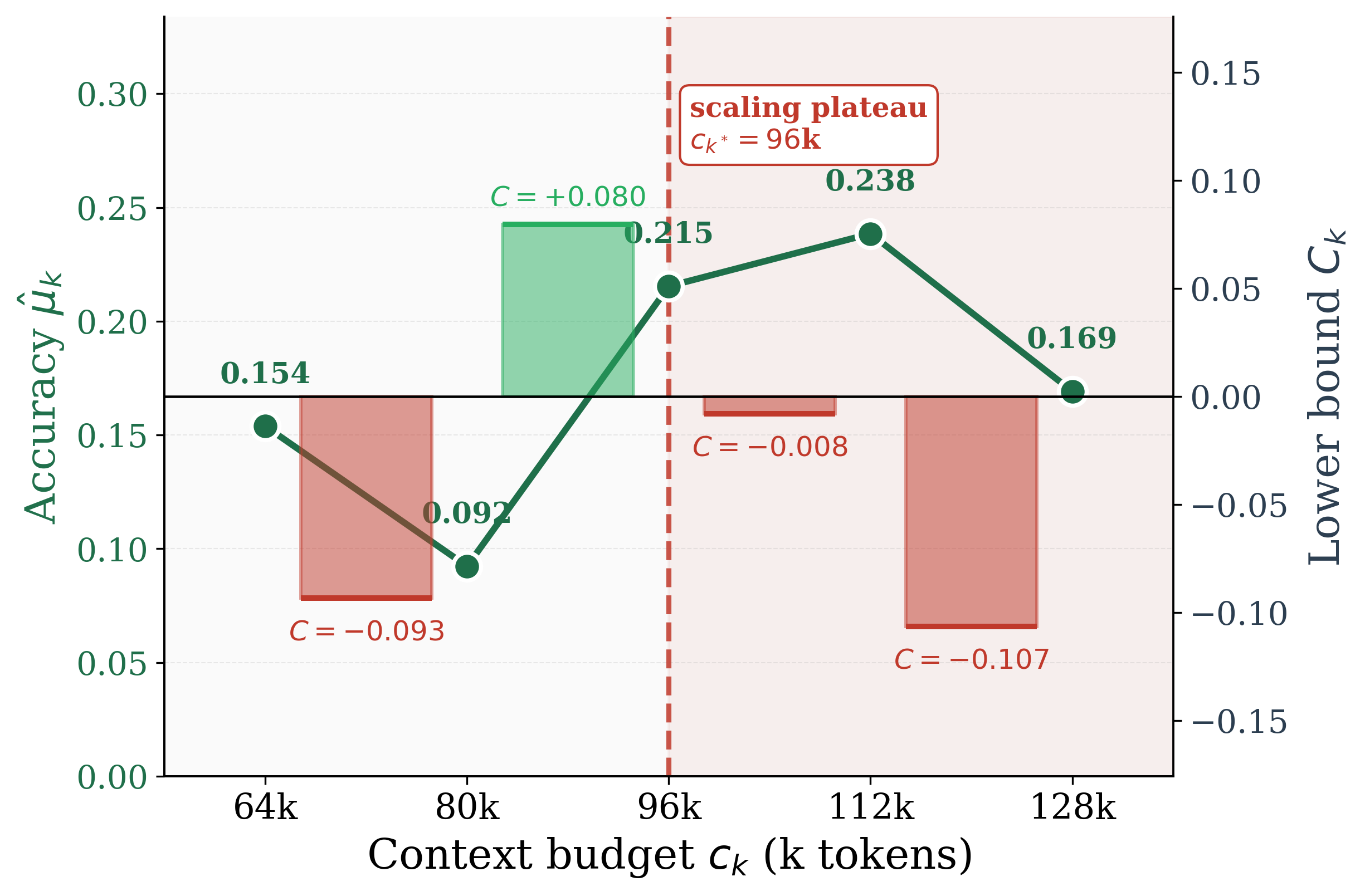}\end{minipage}\hfill
    \begin{minipage}{0.24\linewidth}\centering\includegraphics[width=\linewidth]{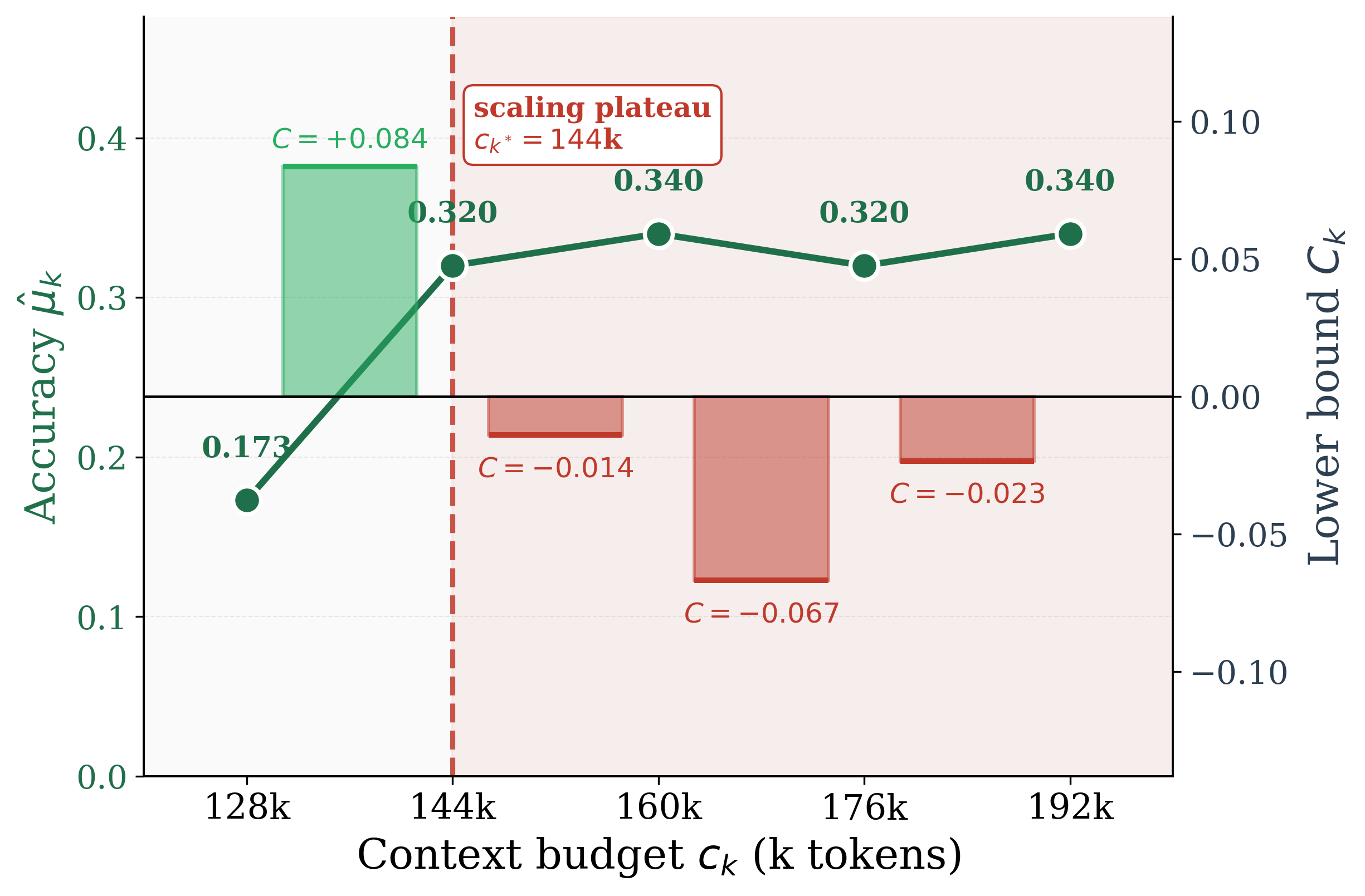}\end{minipage}\hfill
    \begin{minipage}{0.24\linewidth}\centering\includegraphics[width=\linewidth]{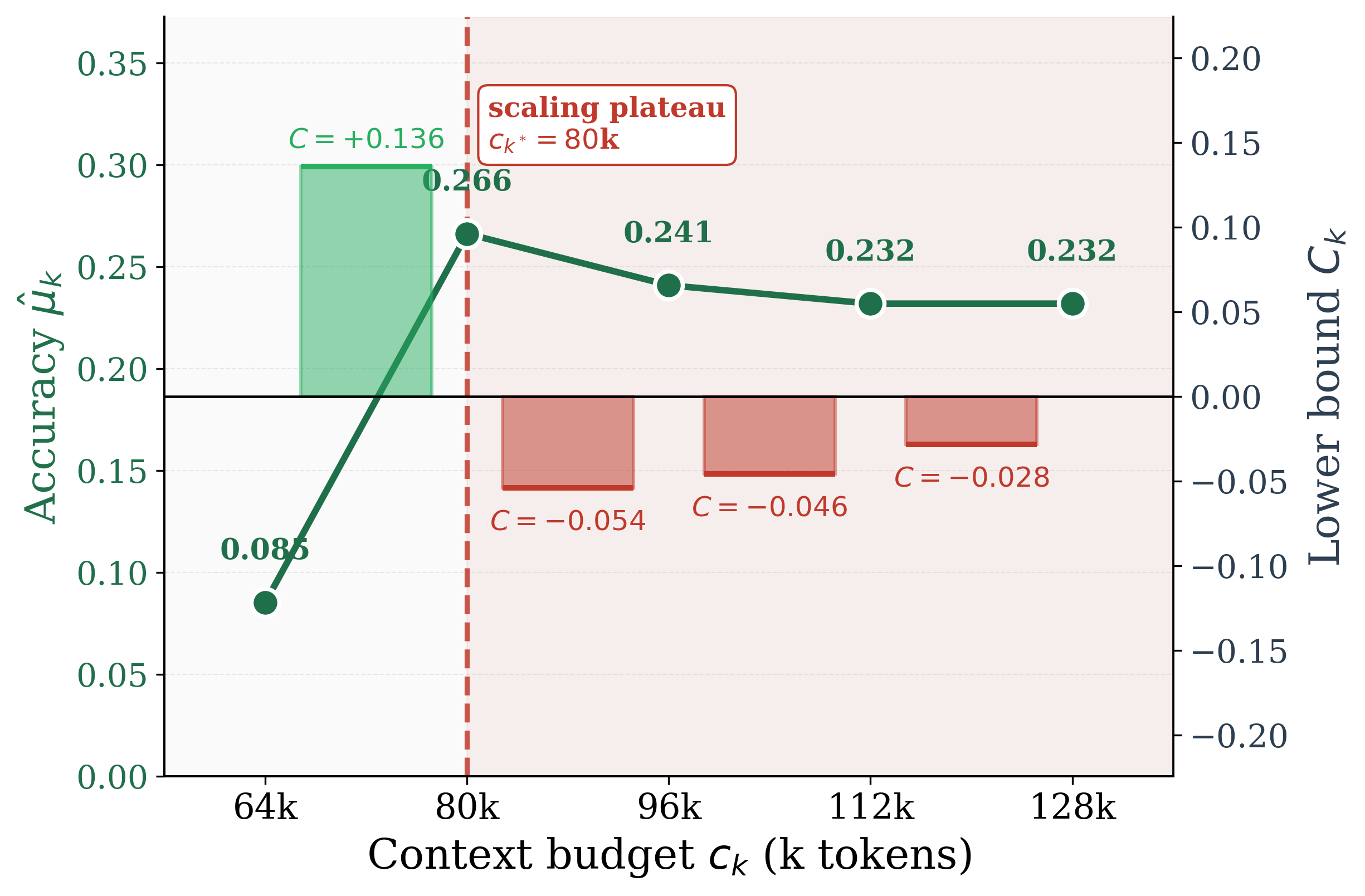}\end{minipage}\hfill
    \begin{minipage}{0.24\linewidth}\centering\includegraphics[width=\linewidth]{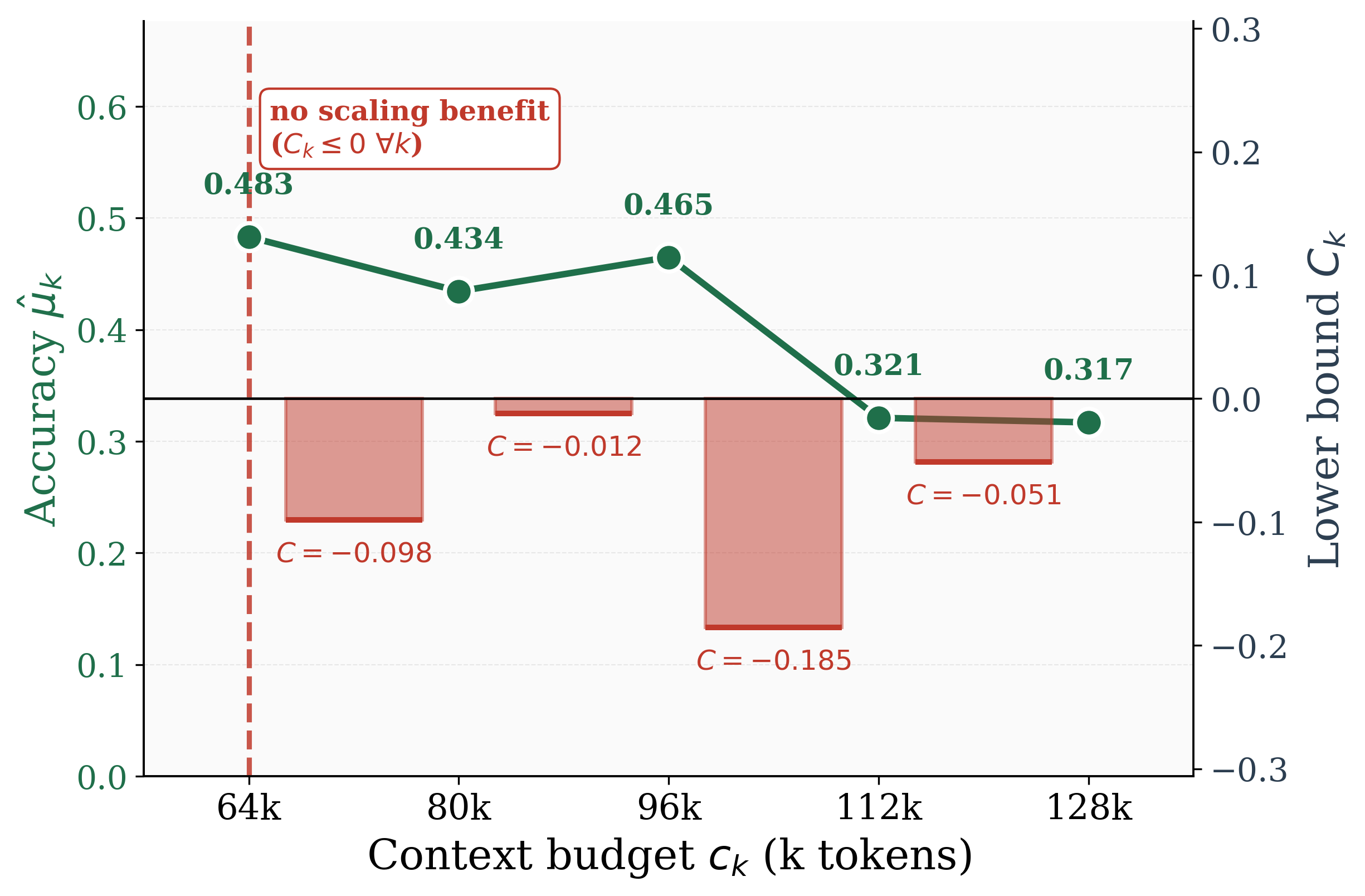}\end{minipage}
    \vspace{4pt}
    \noindent\textbf{\small Gemini-2.5-Flash}\par\vspace{1pt}%
    \begin{minipage}{0.24\linewidth}\centering\includegraphics[width=\linewidth]{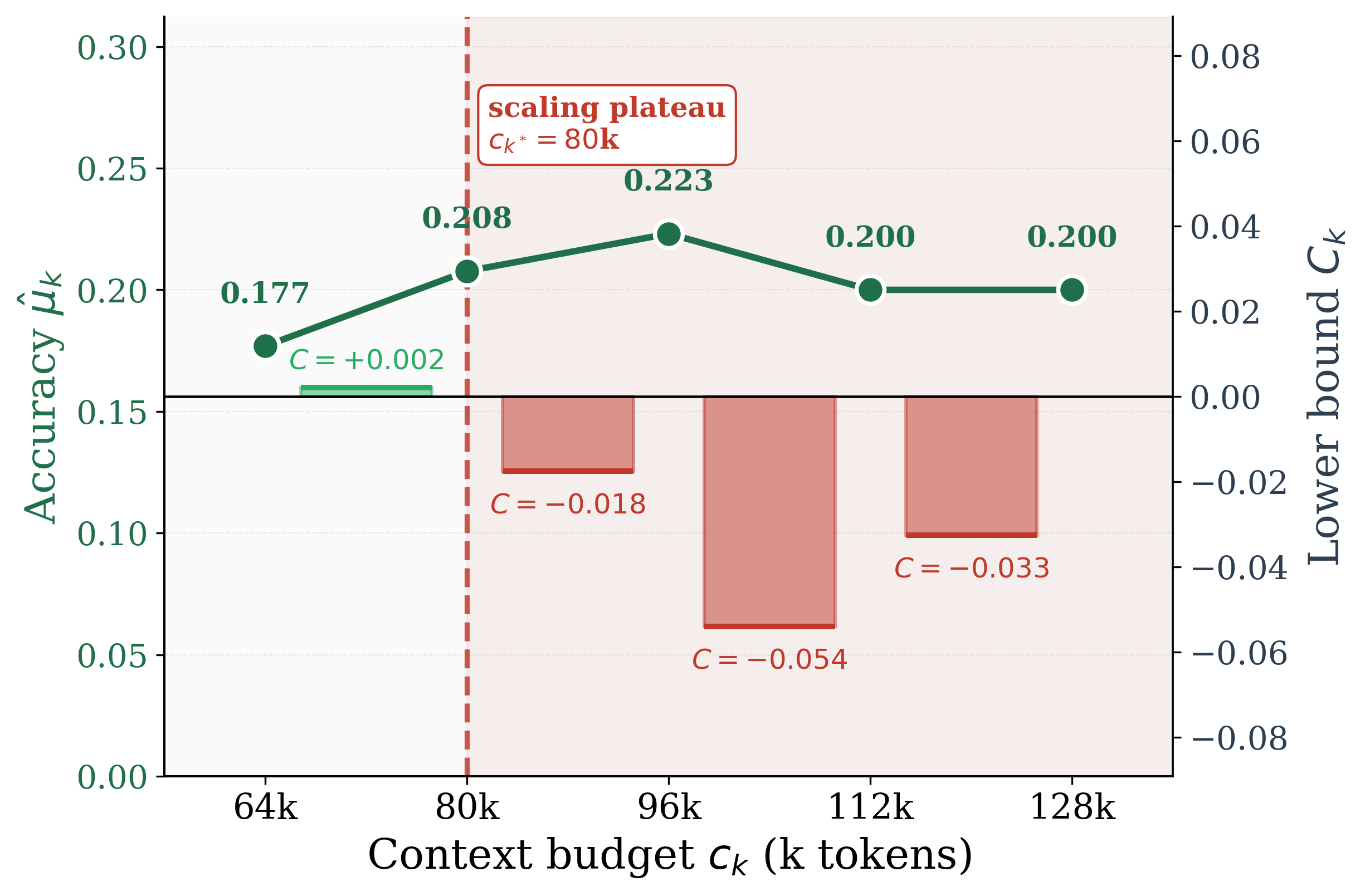}\end{minipage}\hfill
    \begin{minipage}{0.24\linewidth}\centering\includegraphics[width=\linewidth]{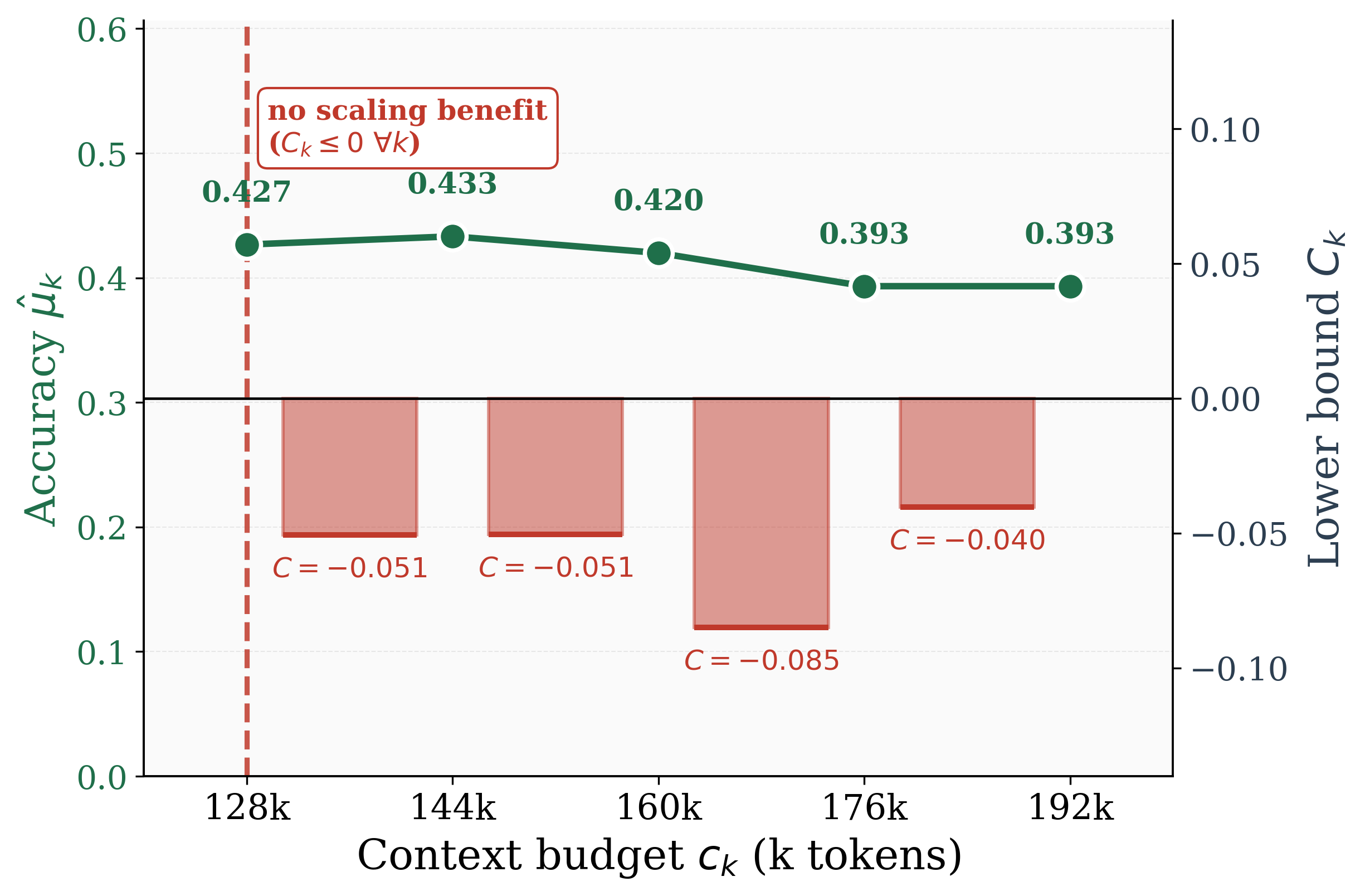}\end{minipage}\hfill
    \begin{minipage}{0.24\linewidth}\centering\includegraphics[width=\linewidth]{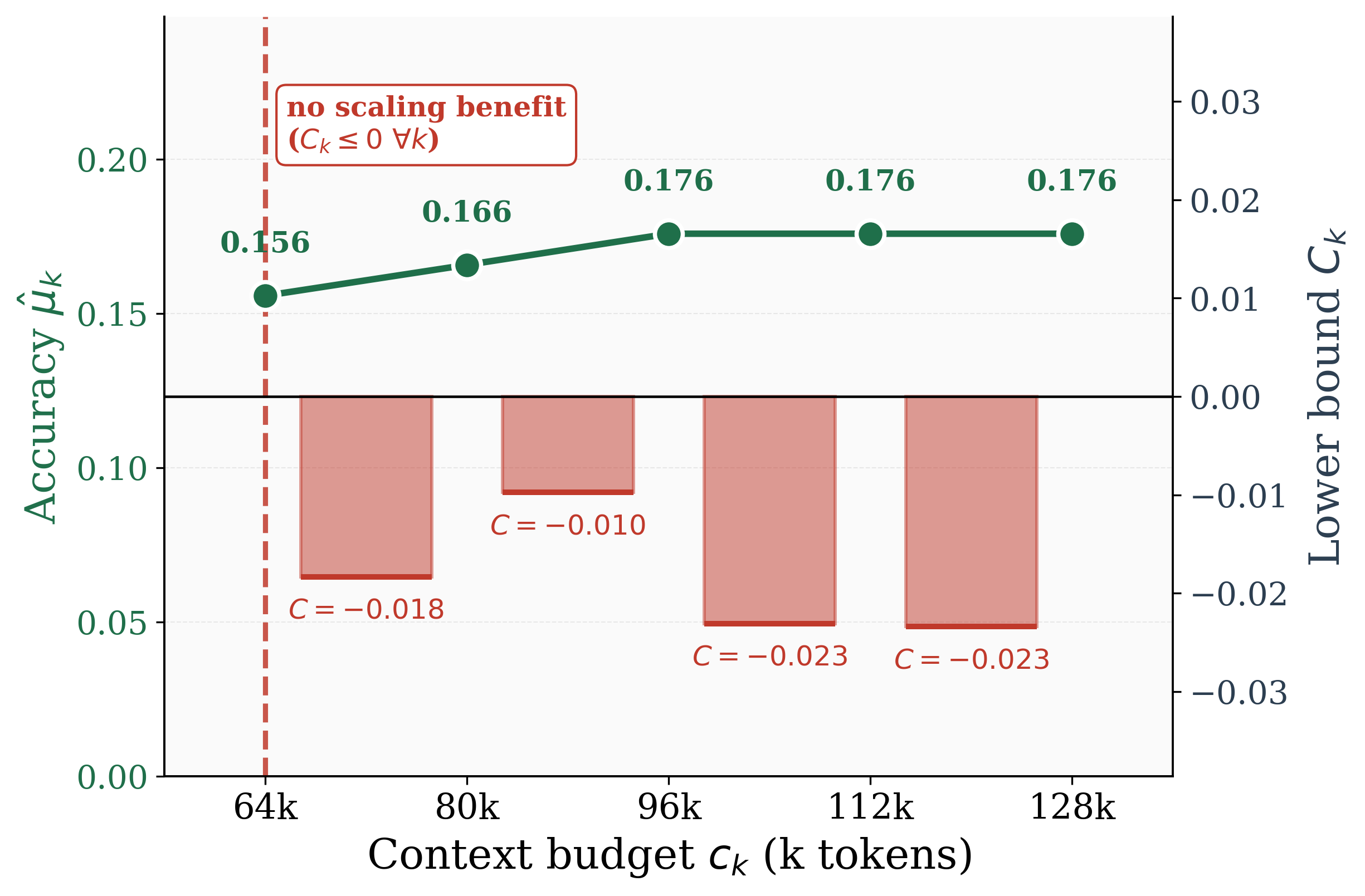}\end{minipage}\hfill
    \begin{minipage}{0.24\linewidth}\centering\includegraphics[width=\linewidth]{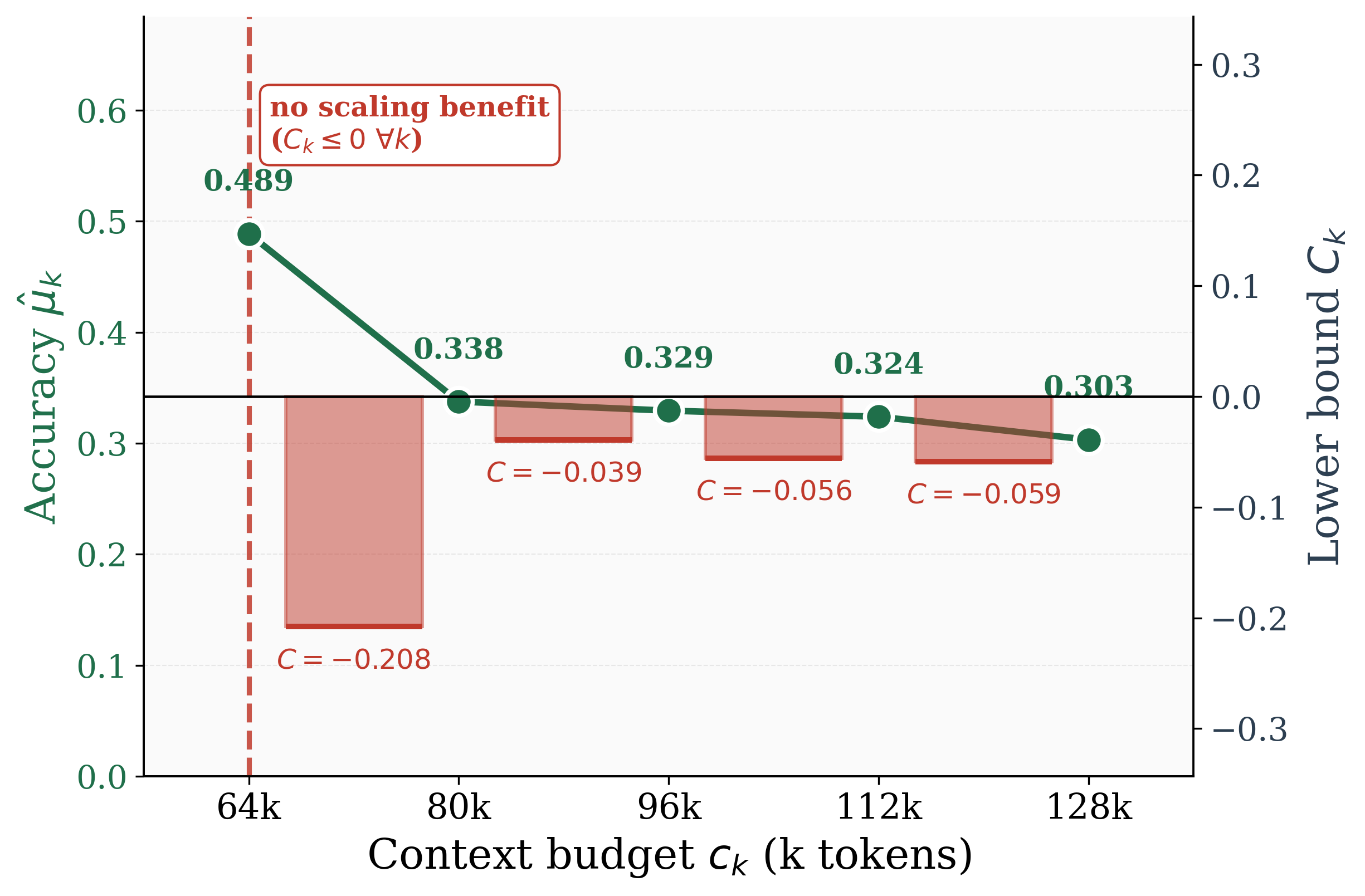}\end{minipage}
    \vspace{4pt}
    \noindent\textbf{\small Qwen3-235B}\par\vspace{1pt}%
    \begin{minipage}{0.24\linewidth}\centering\includegraphics[width=\linewidth]{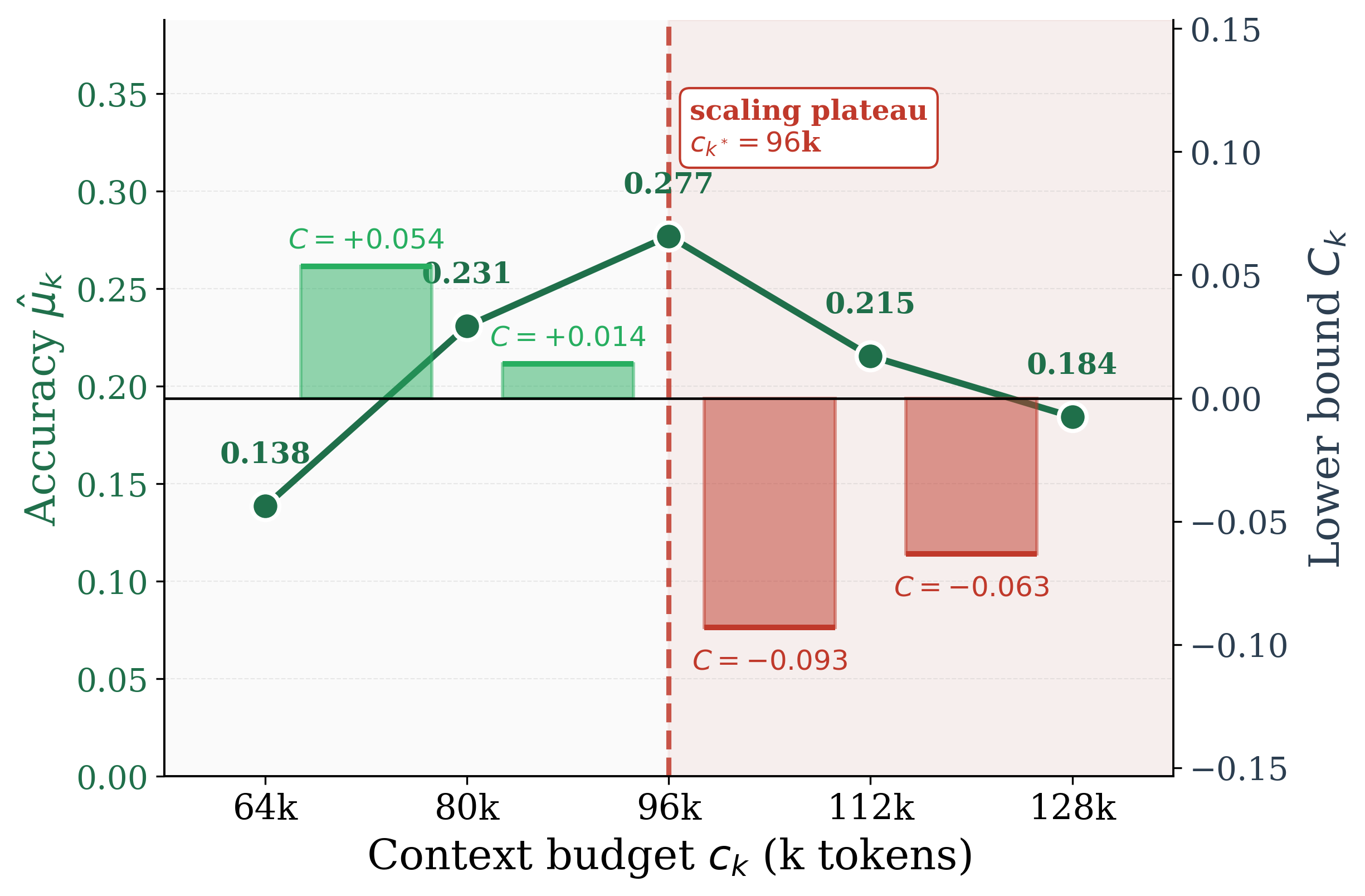}\end{minipage}\hfill
    \begin{minipage}{0.24\linewidth}\centering\includegraphics[width=\linewidth]{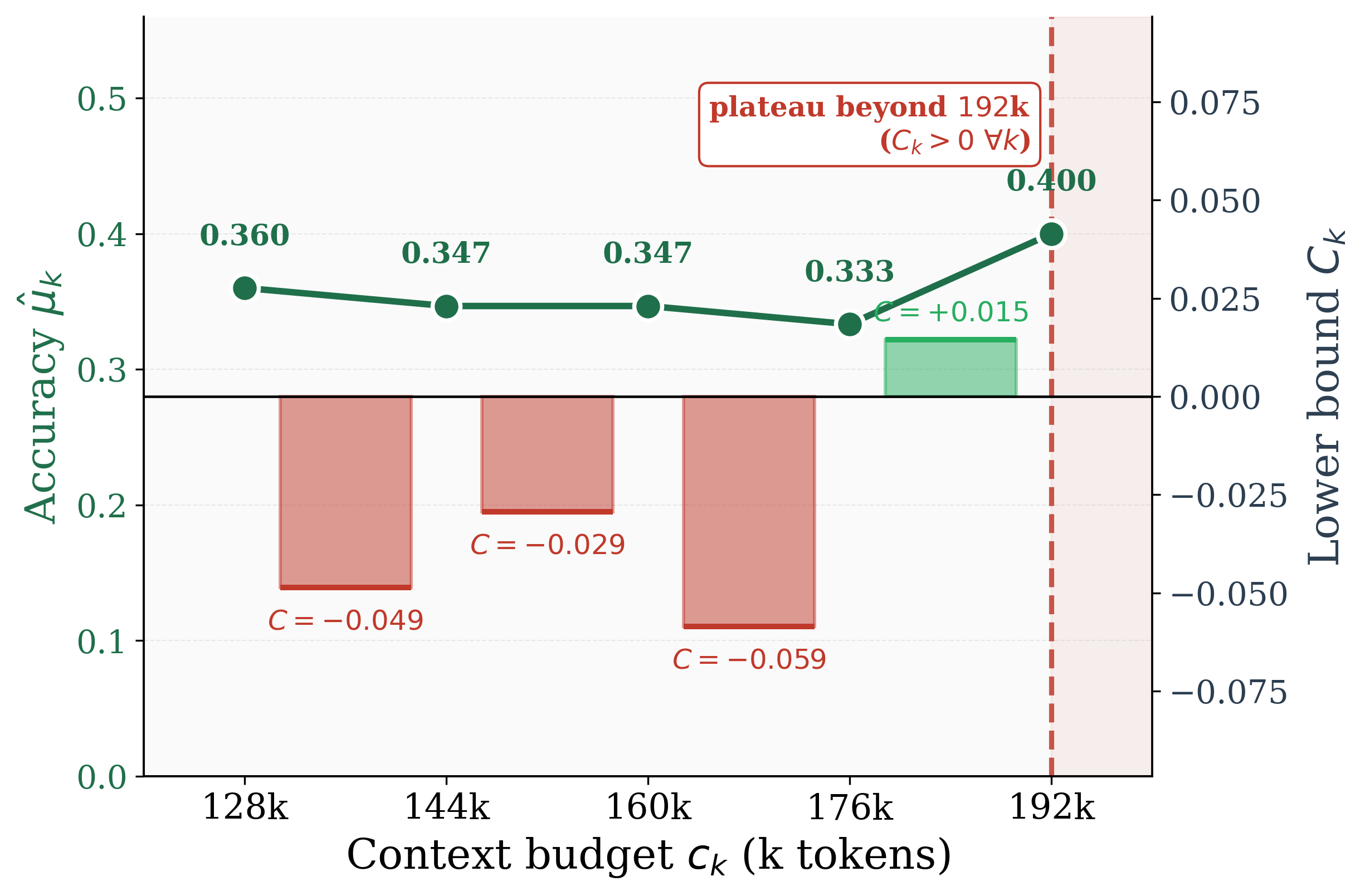}\end{minipage}\hfill
    \begin{minipage}{0.24\linewidth}\centering\includegraphics[width=\linewidth]{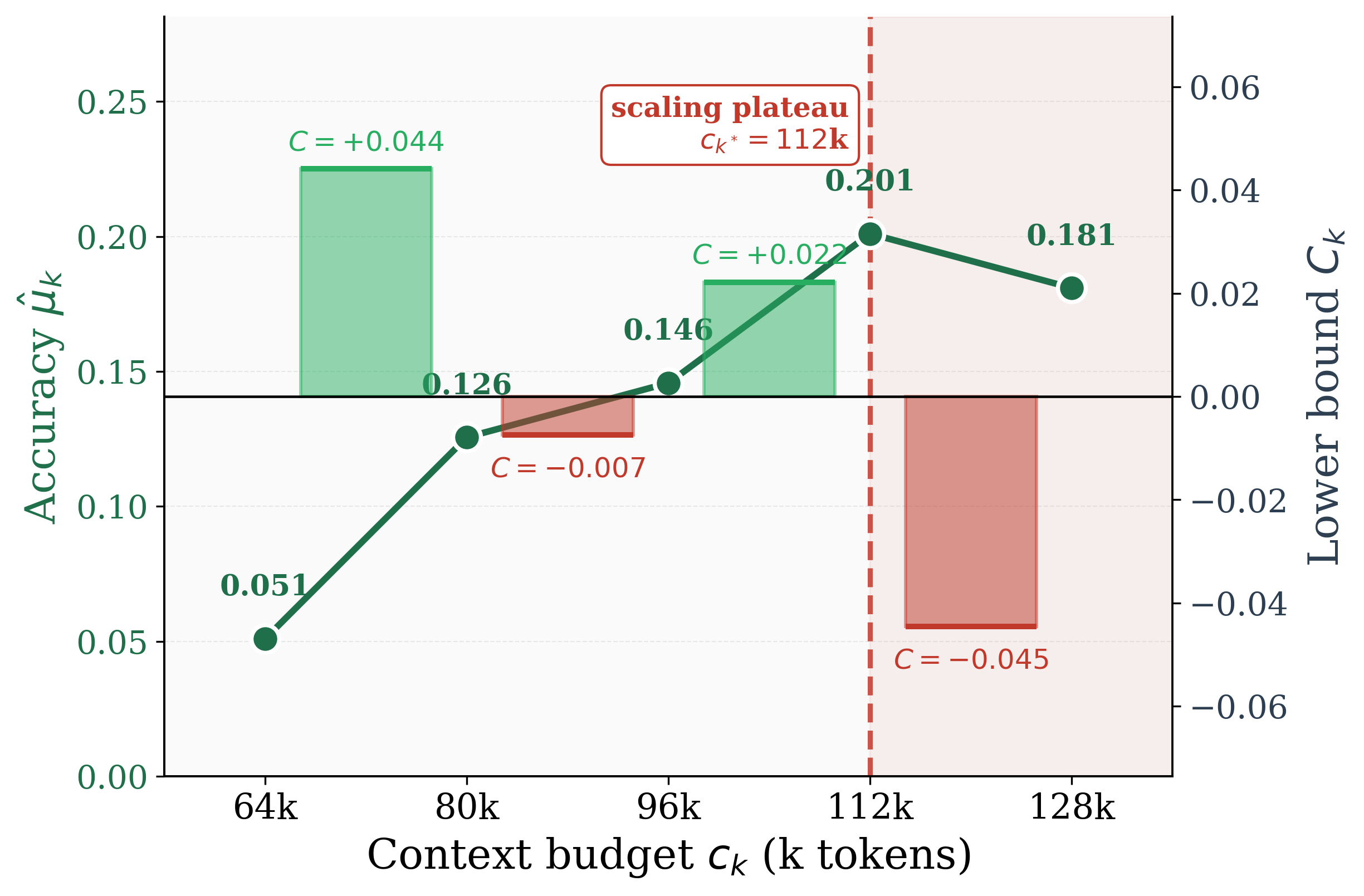}\end{minipage}\hfill
    \begin{minipage}{0.24\linewidth}\centering\includegraphics[width=\linewidth]{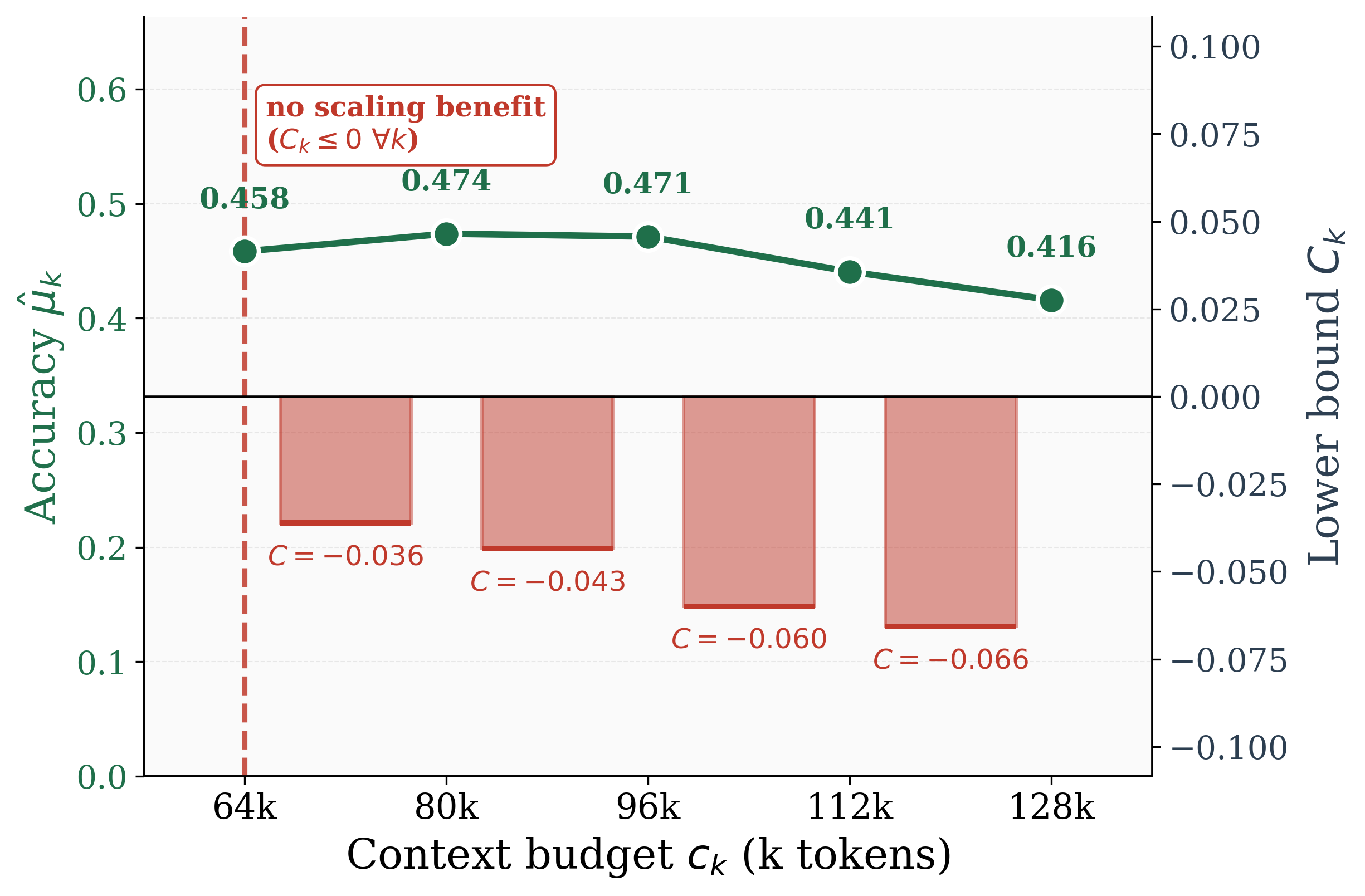}\end{minipage}
    \vspace{4pt}
    \noindent\textbf{\small Qwen3-Next}\par\vspace{1pt}%
    \begin{minipage}{0.24\linewidth}\centering\includegraphics[width=\linewidth]{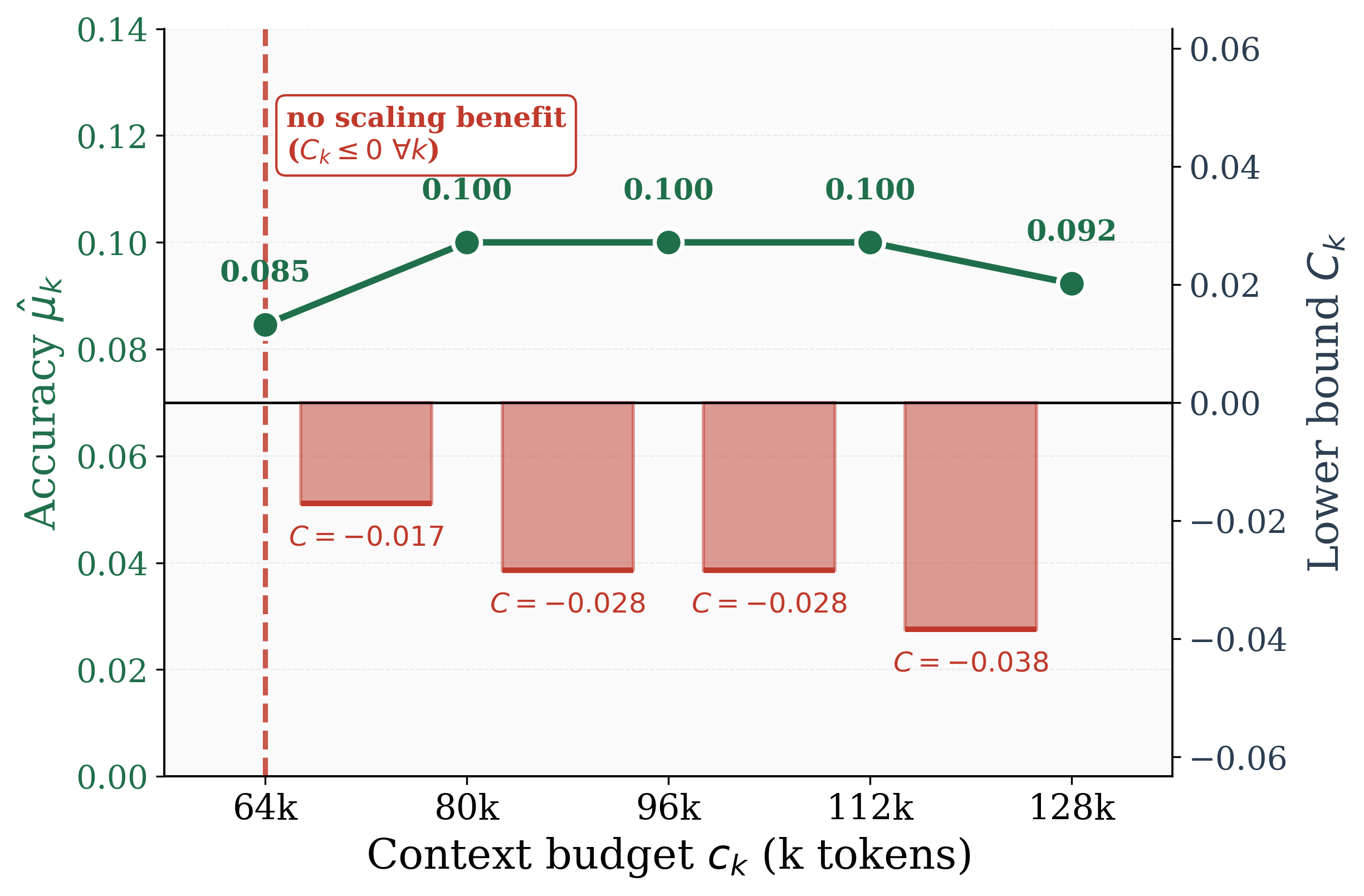}\end{minipage}\hfill
    \begin{minipage}{0.24\linewidth}\centering\includegraphics[width=\linewidth]{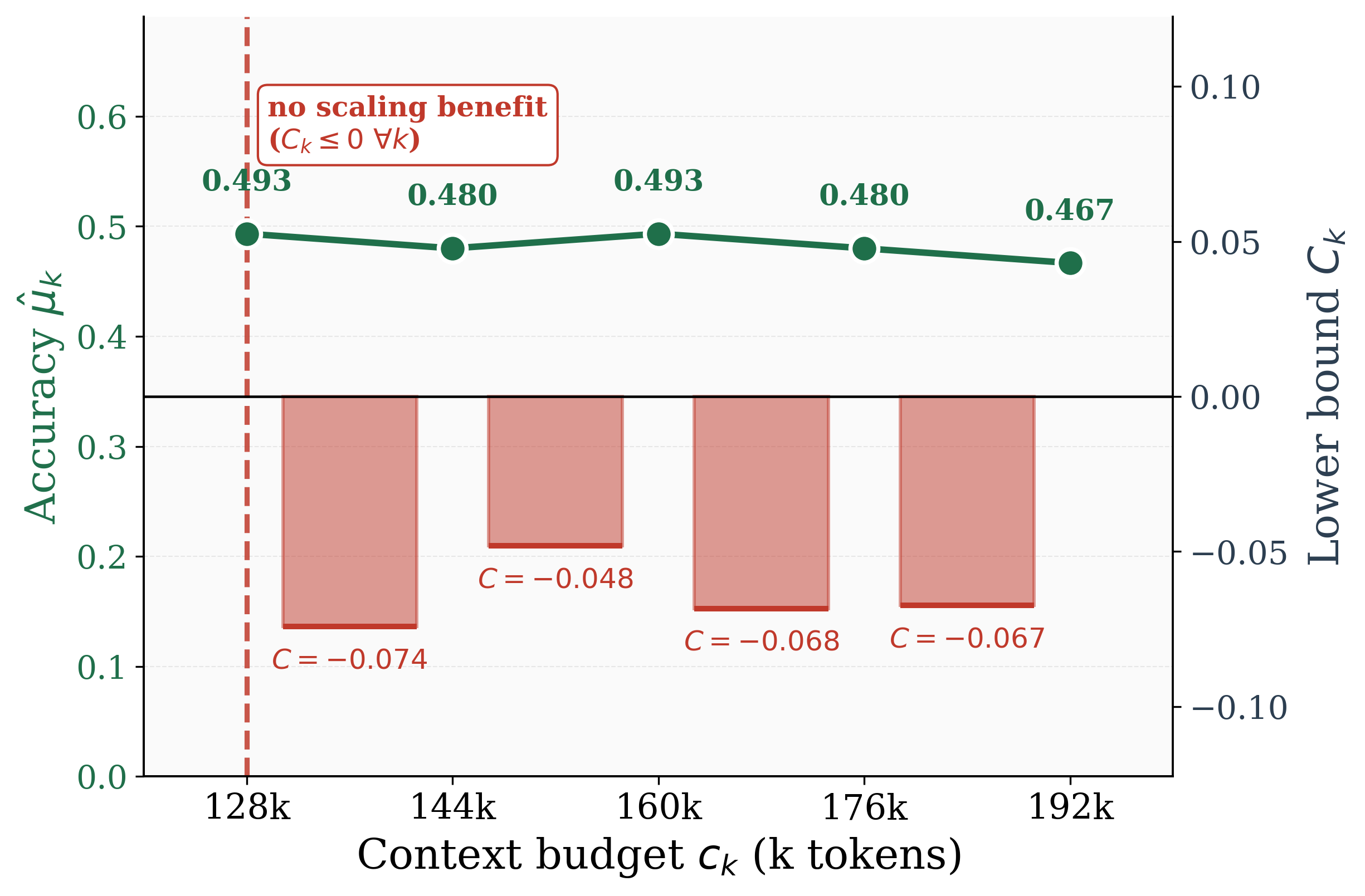}\end{minipage}\hfill
    \begin{minipage}{0.24\linewidth}\centering\includegraphics[width=\linewidth]{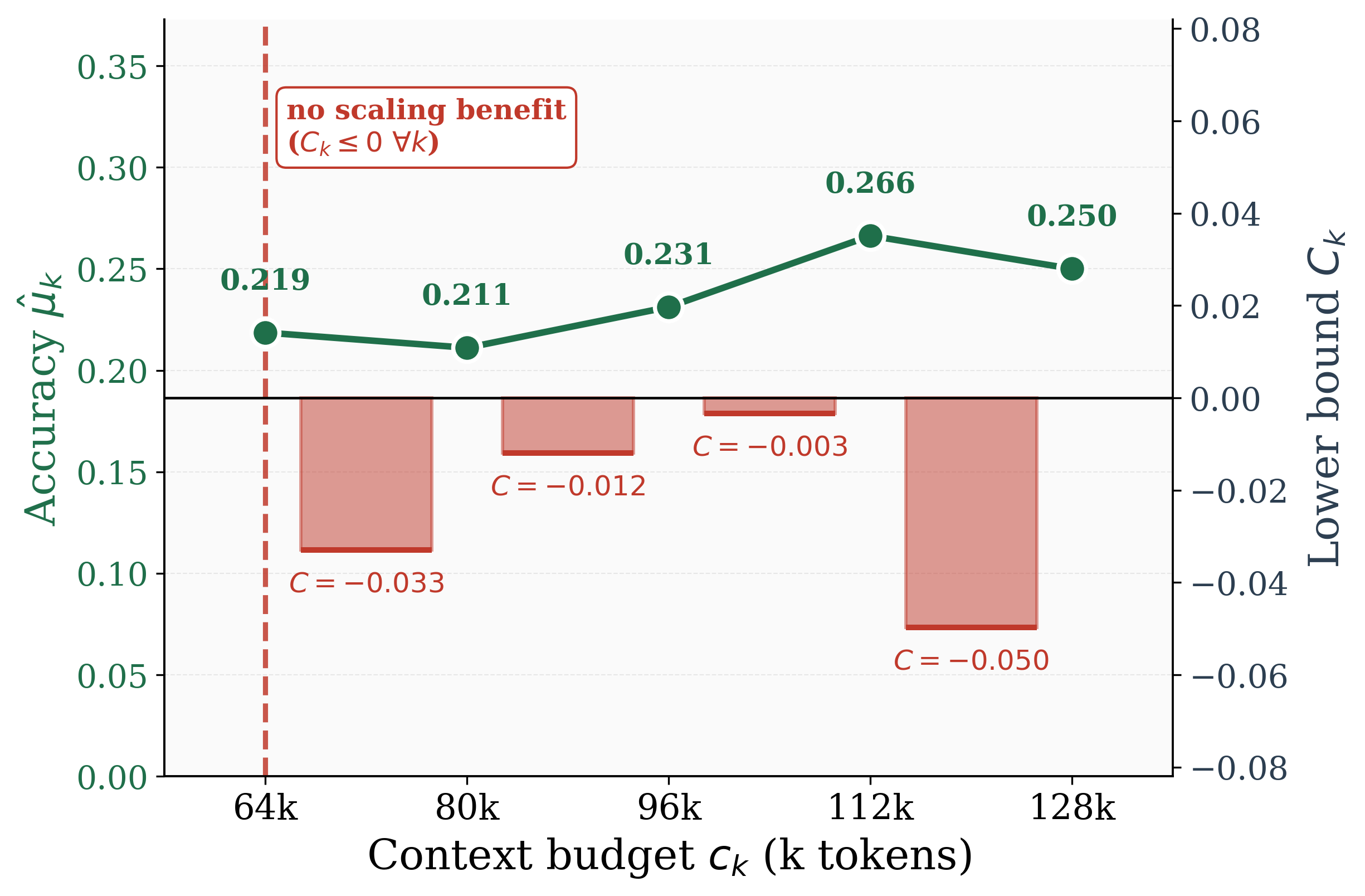}\end{minipage}\hfill
    \begin{minipage}{0.24\linewidth}\centering\includegraphics[width=\linewidth]{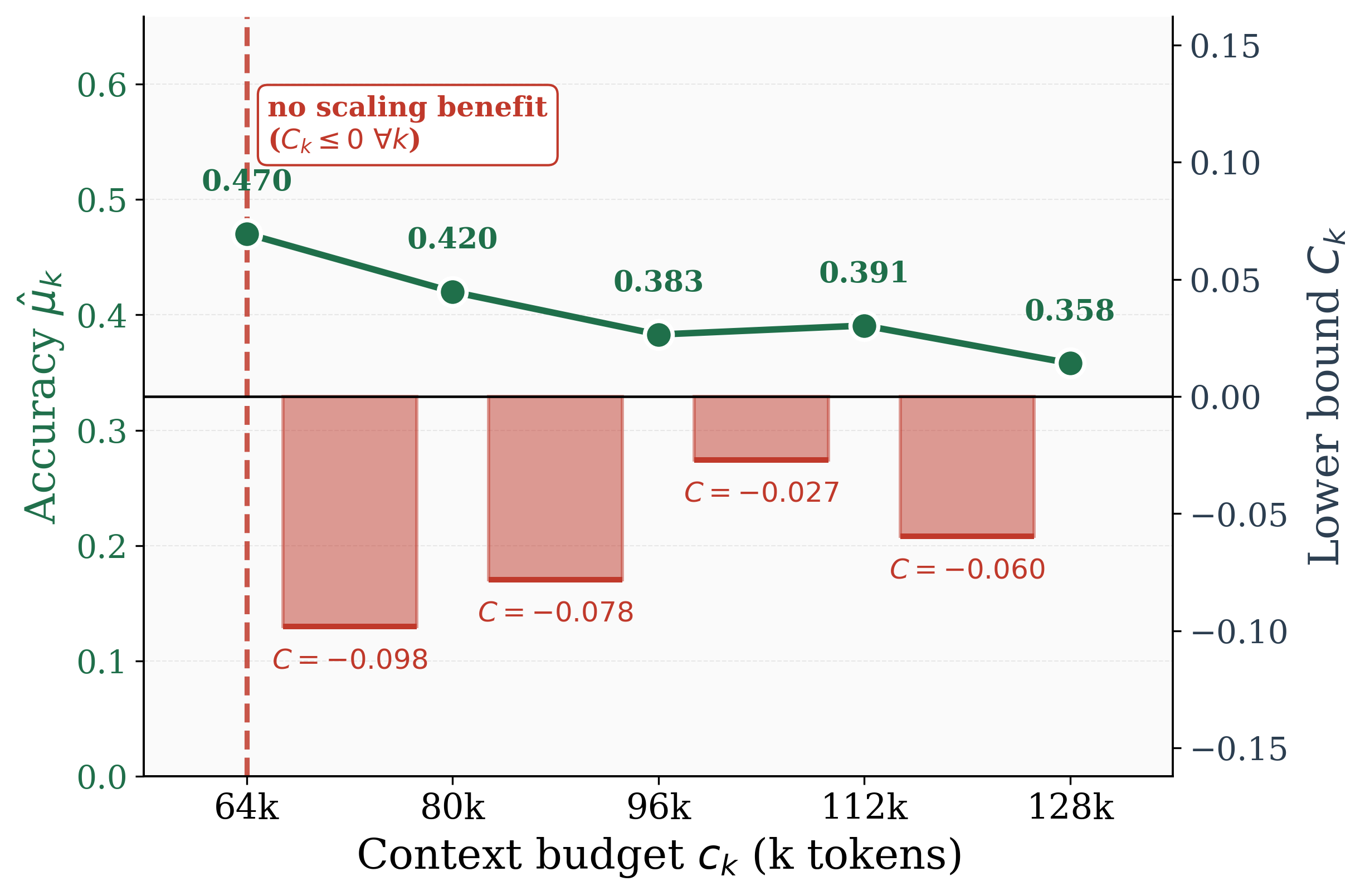}\end{minipage}
    \vspace{4pt}
    \caption{\textbf{Accuracy curves and confidence lower bounds $C_k$ for all model--domain pairs.} Each panel shows the estimated accuracy $\hat{\mu}_k$ (line) and the one-sided lower bound $C_k$ (bars) at each context-budget transition. Green bars indicate $C_k > 0$ (significant gain); red bars indicate $C_k \leq 0$. The dashed vertical line marks the identified scaling plateau $c_{k^*}$.}
    \label{fig:plateau_accuracy}
\end{figure}

\begin{figure}[t]
    \centering
    \plateaucolhead{Code}\hfill\plateaucolhead{Reason}\hfill\plateaucolhead{Search}\hfill\plateaucolhead{Tool-use}\par\vspace{4pt}
    \noindent\textbf{\small DeepSeek-R1}\par\vspace{1pt}%
    \begin{minipage}{0.24\linewidth}\centering\includegraphics[width=\linewidth]{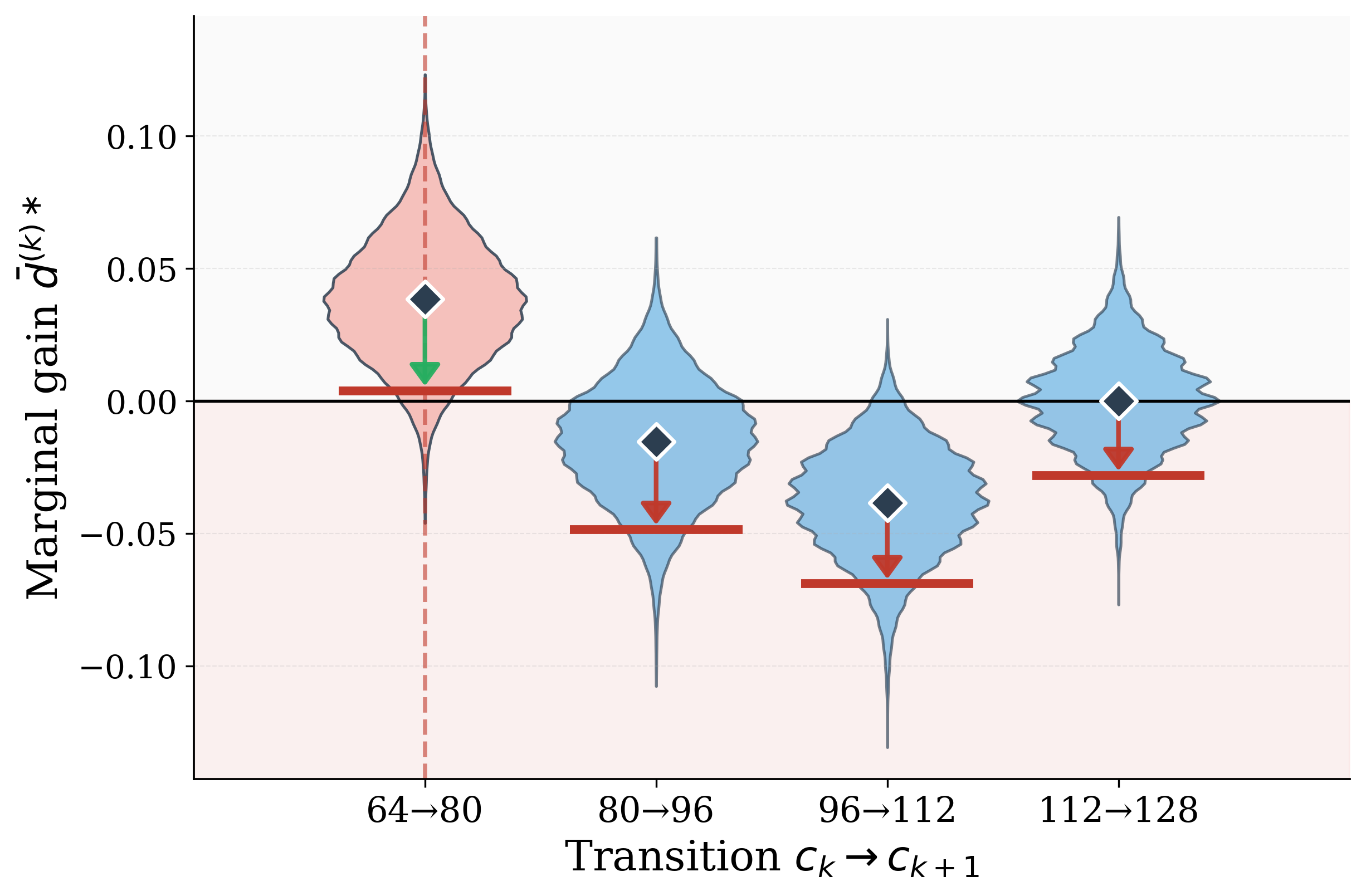}\end{minipage}\hfill
    \begin{minipage}{0.24\linewidth}\centering\includegraphics[width=\linewidth]{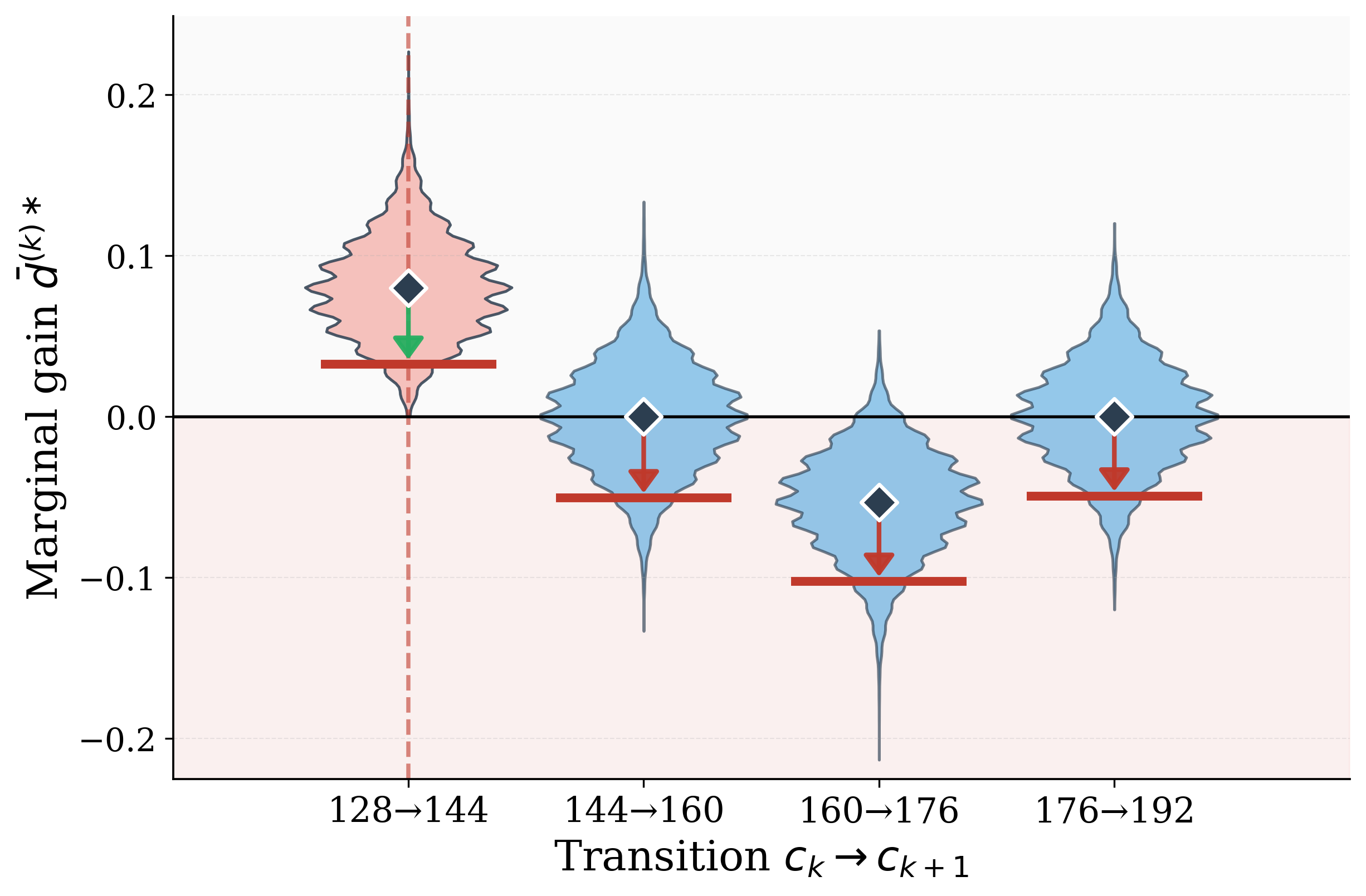}\end{minipage}\hfill
    \begin{minipage}{0.24\linewidth}\centering\includegraphics[width=\linewidth]{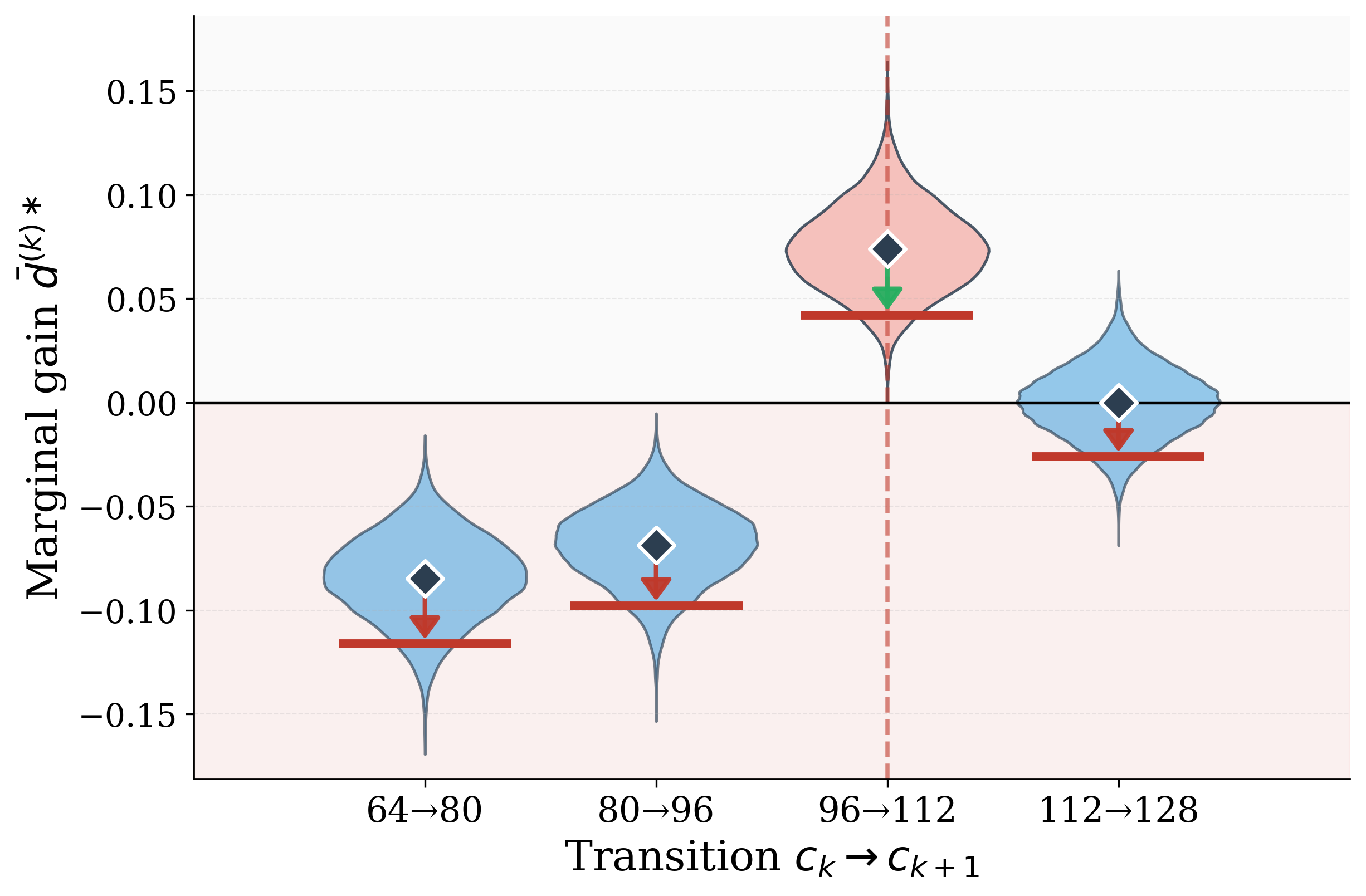}\end{minipage}\hfill
    \begin{minipage}{0.24\linewidth}\centering\includegraphics[width=\linewidth]{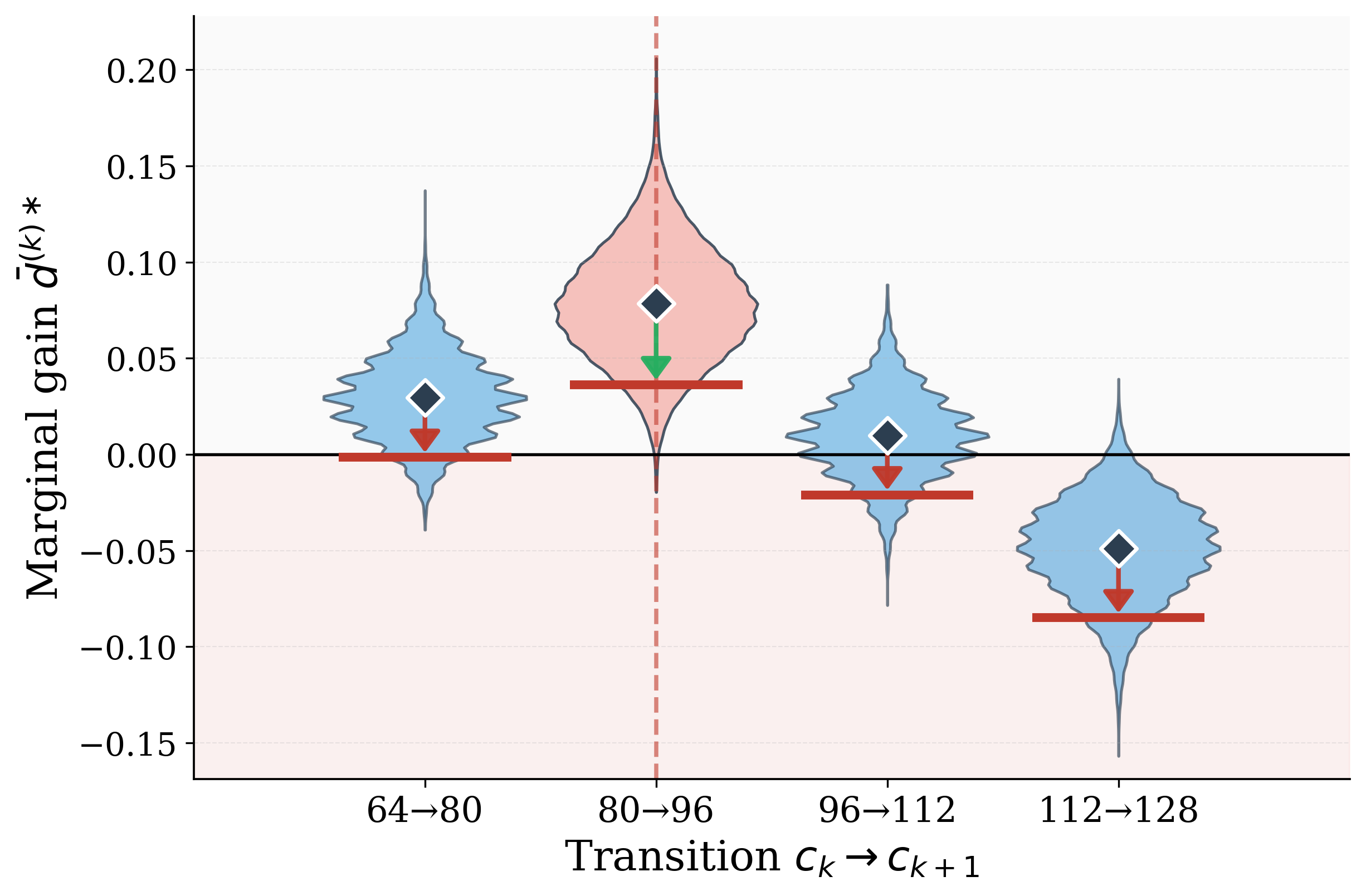}\end{minipage}
    \vspace{4pt}
    \noindent\textbf{\small DeepSeek-V3.2}\par\vspace{1pt}%
    \begin{minipage}{0.24\linewidth}\centering\includegraphics[width=\linewidth]{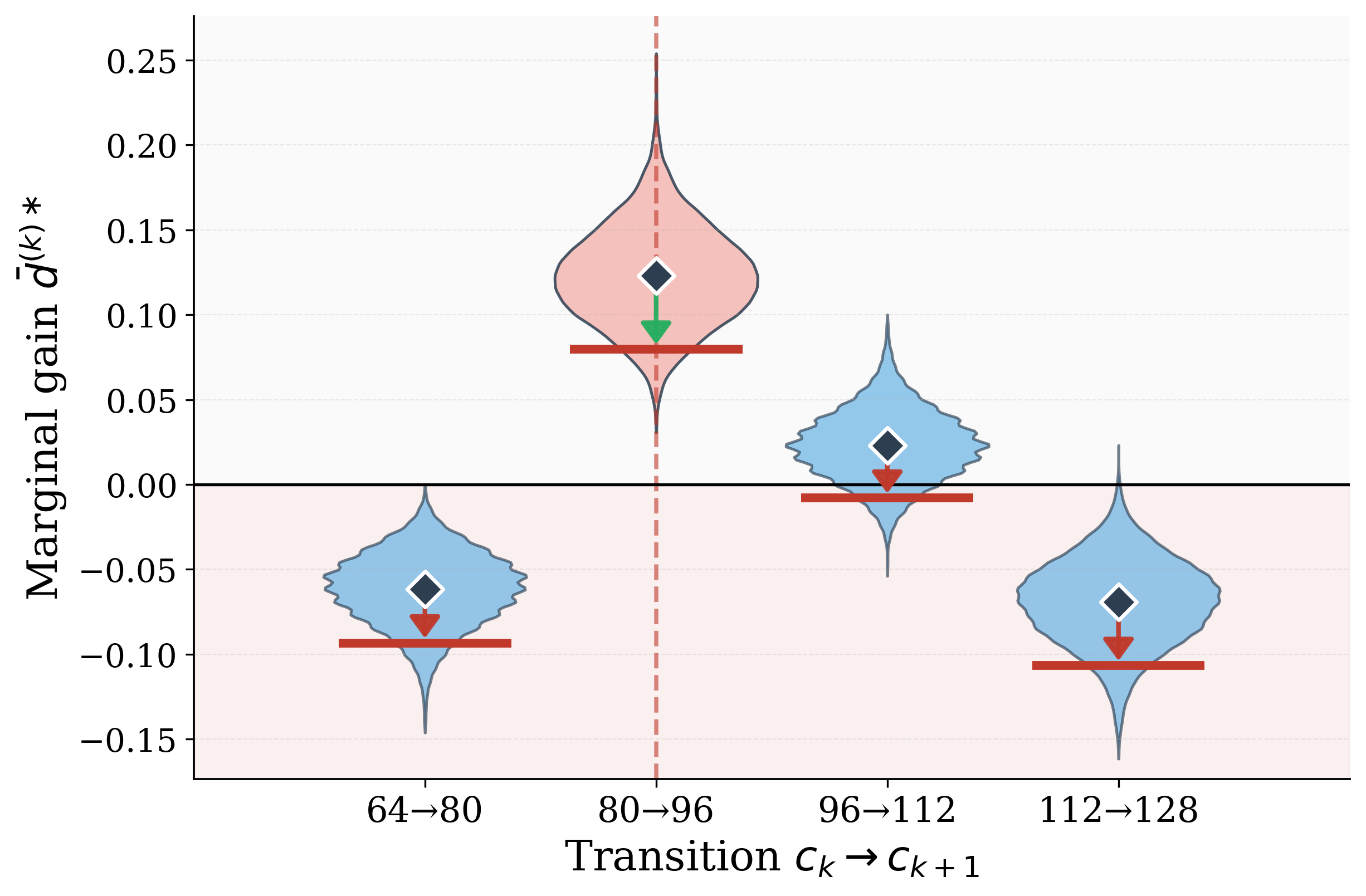}\end{minipage}\hfill
    \begin{minipage}{0.24\linewidth}\centering\includegraphics[width=\linewidth]{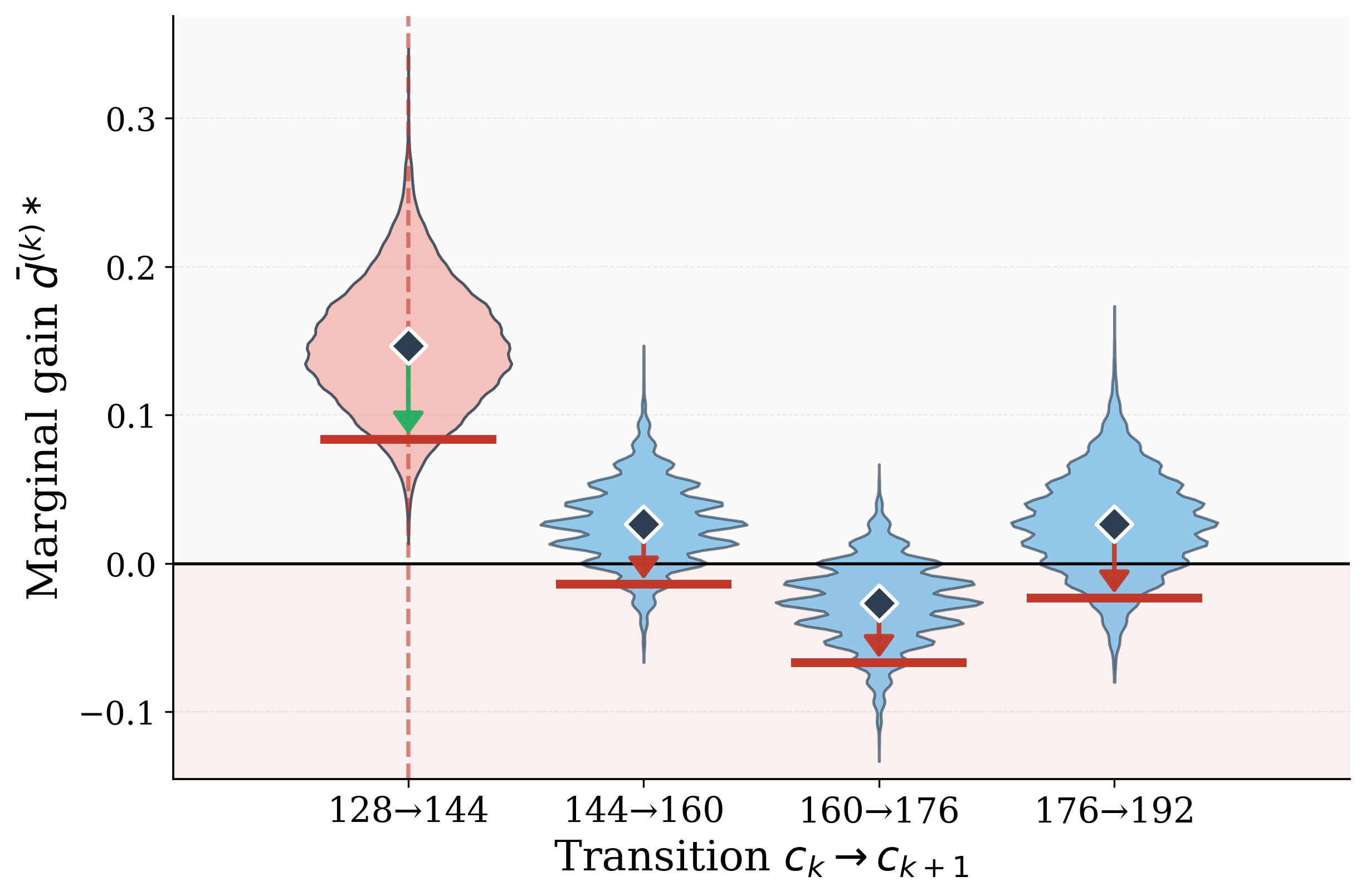}\end{minipage}\hfill
    \begin{minipage}{0.24\linewidth}\centering\includegraphics[width=\linewidth]{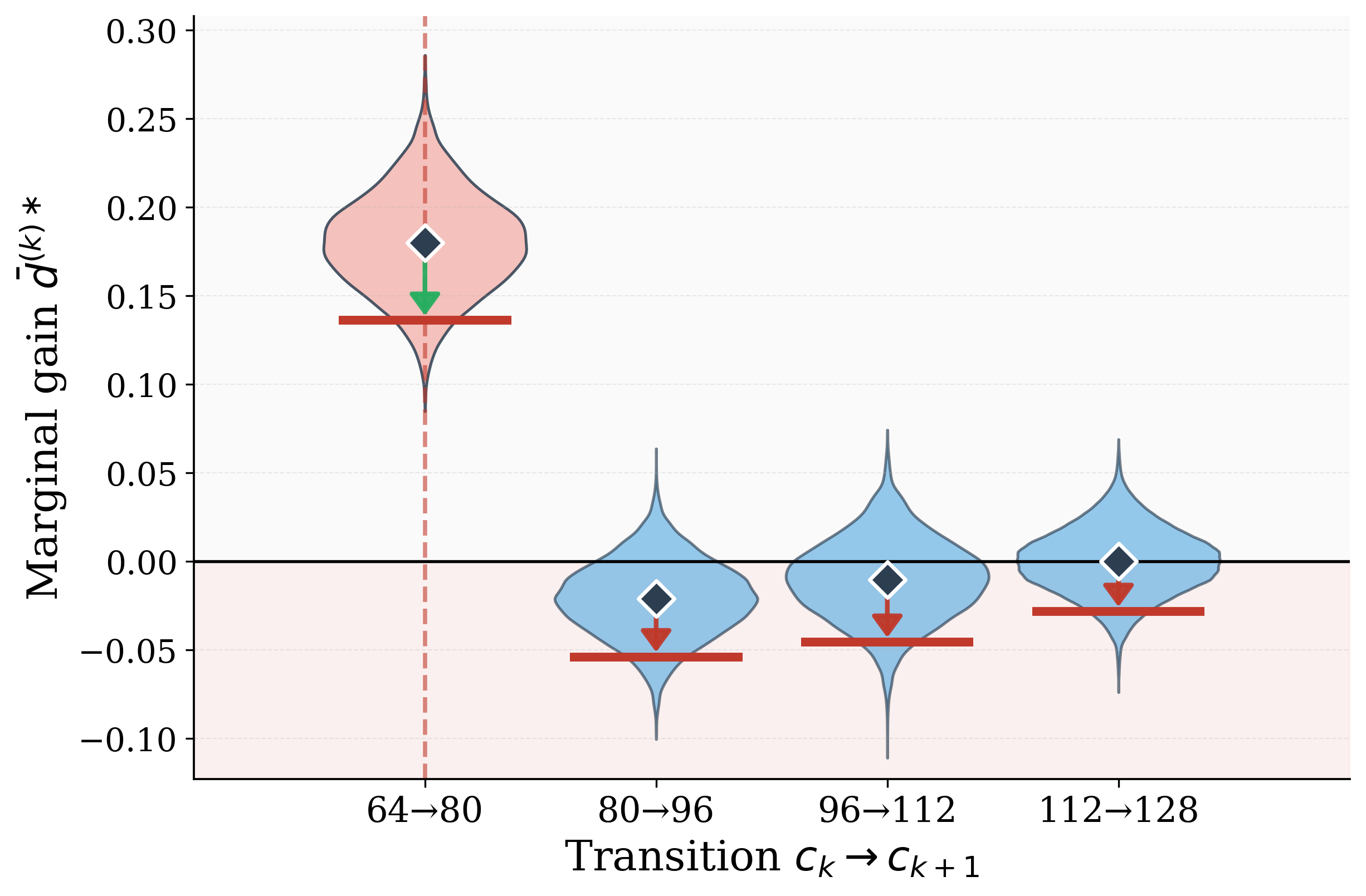}\end{minipage}\hfill
    \begin{minipage}{0.24\linewidth}\centering\includegraphics[width=\linewidth]{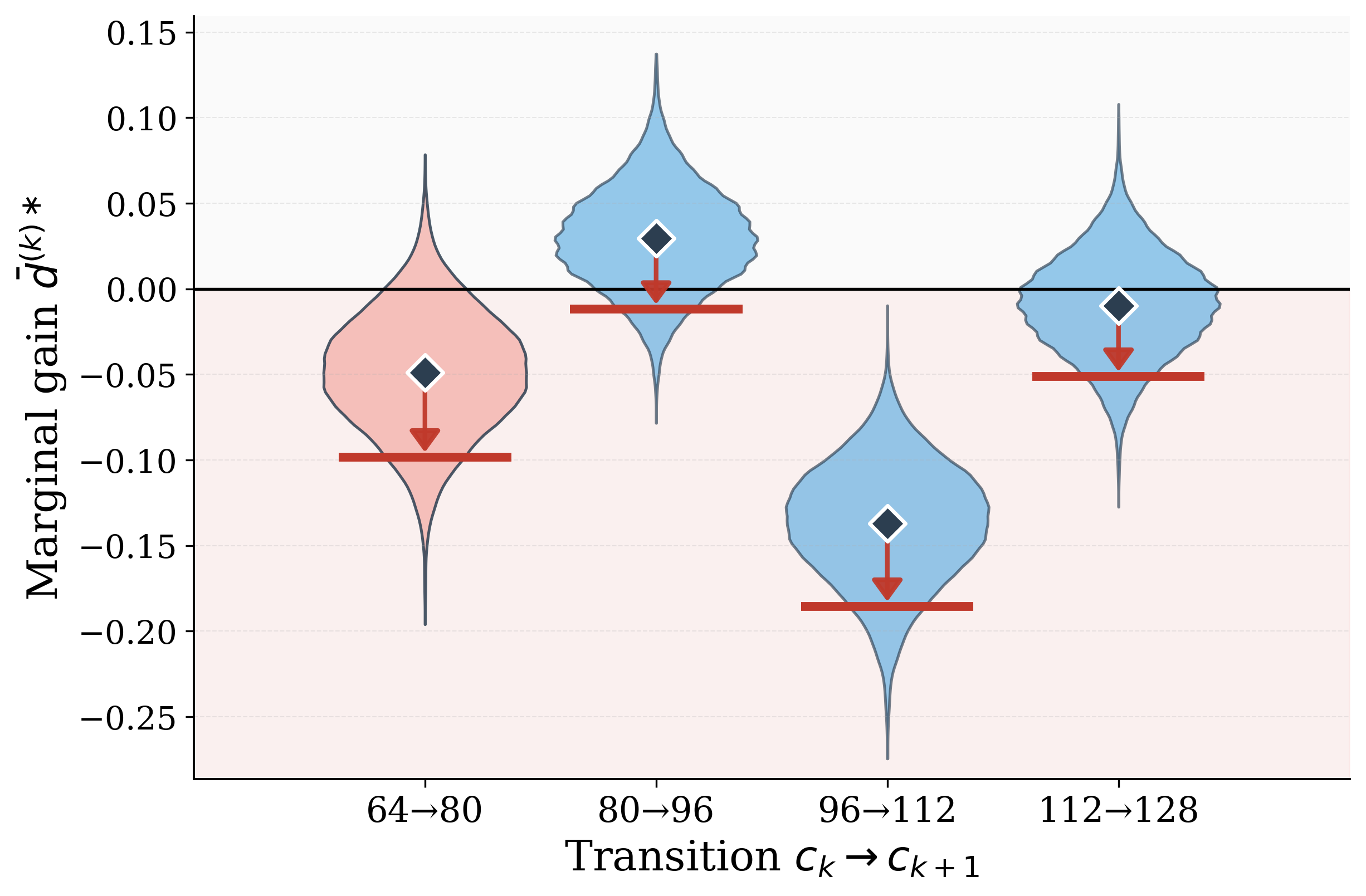}\end{minipage}
    \vspace{4pt}
    \noindent\textbf{\small Gemini-2.5-Flash}\par\vspace{1pt}%
    \begin{minipage}{0.24\linewidth}\centering\includegraphics[width=\linewidth]{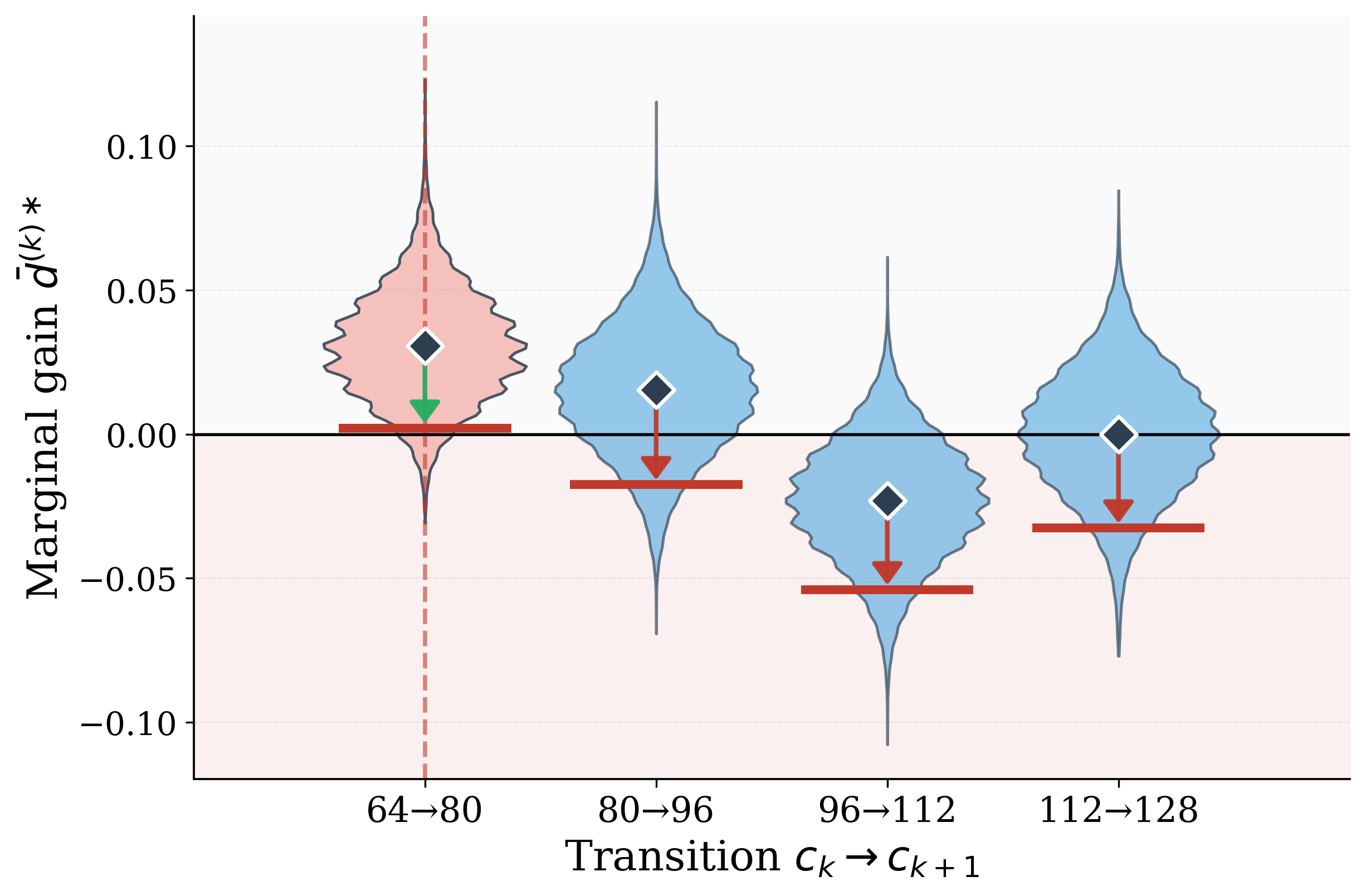}\end{minipage}\hfill
    \begin{minipage}{0.24\linewidth}\centering\includegraphics[width=\linewidth]{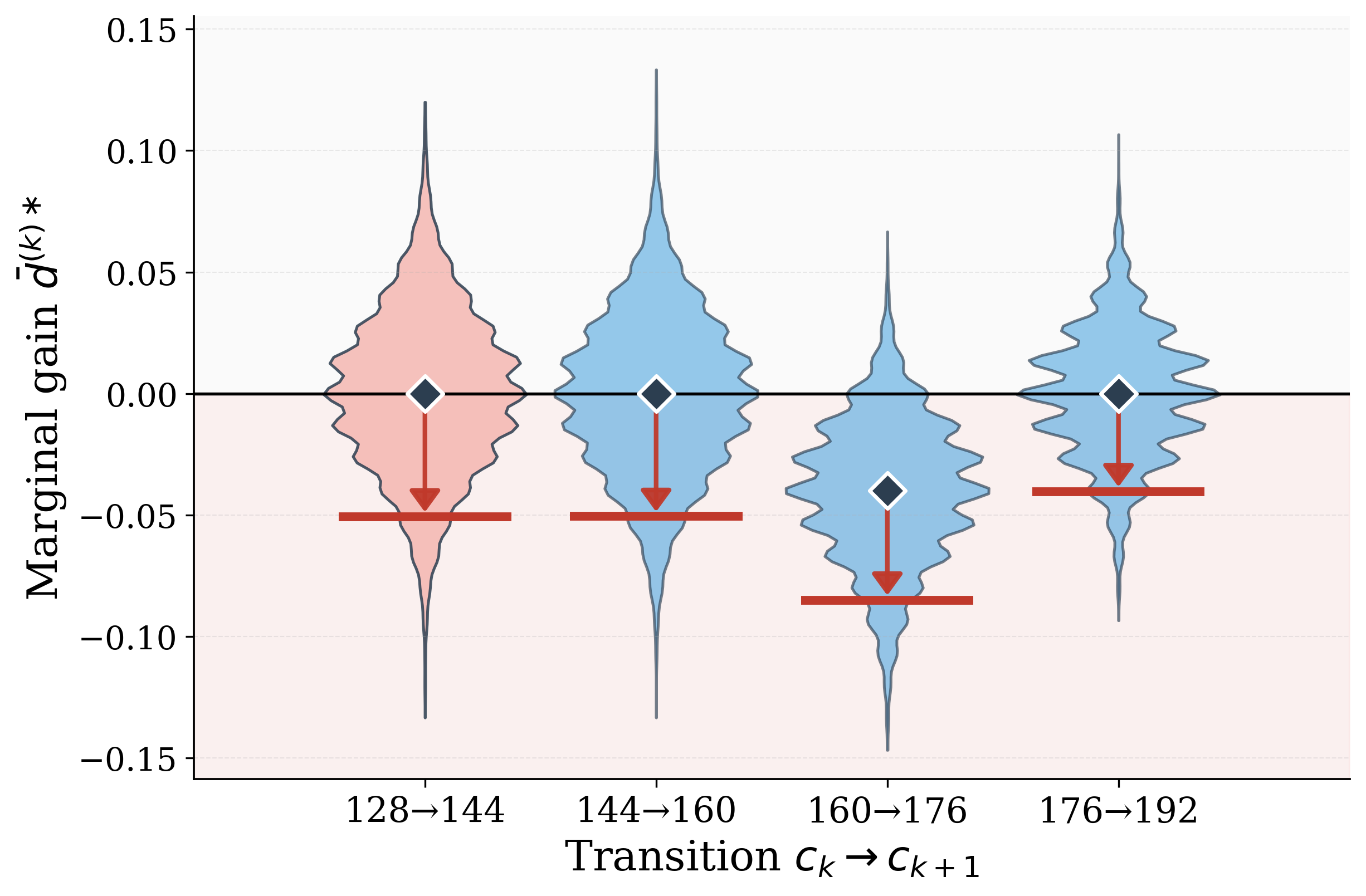}\end{minipage}\hfill
    \begin{minipage}{0.24\linewidth}\centering\includegraphics[width=\linewidth]{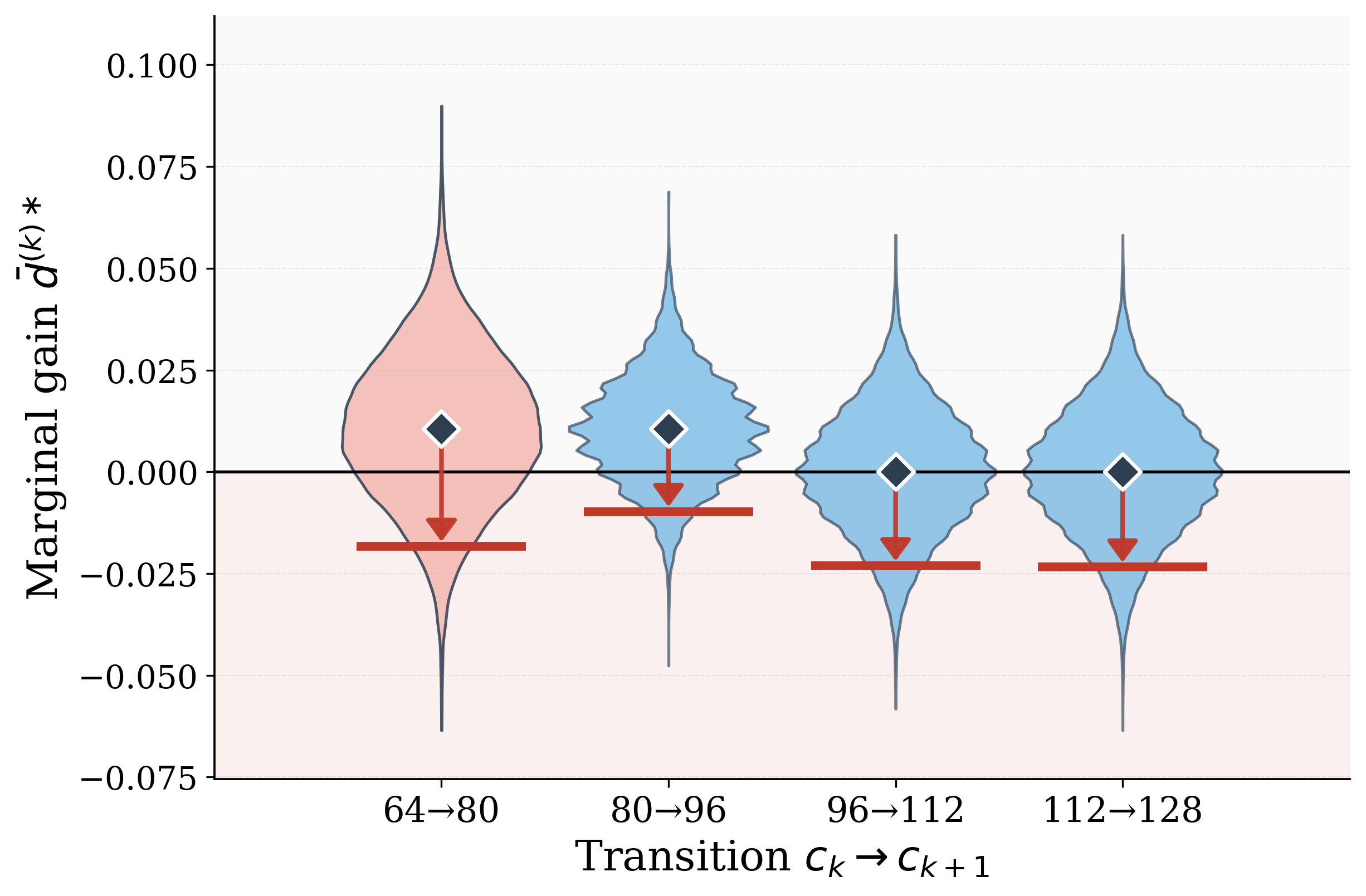}\end{minipage}\hfill
    \begin{minipage}{0.24\linewidth}\centering\includegraphics[width=\linewidth]{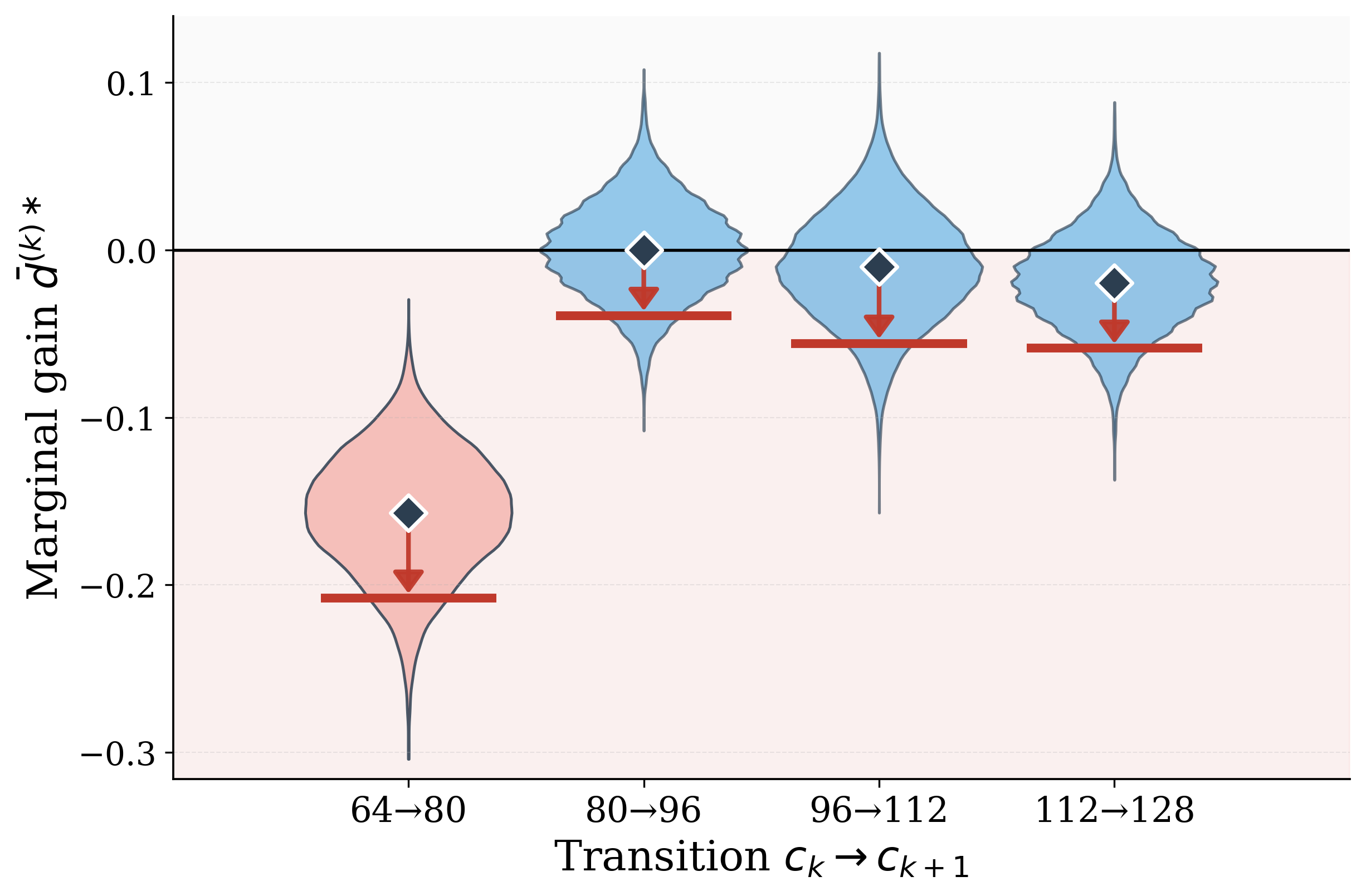}\end{minipage}
    \vspace{4pt}
    \noindent\textbf{\small Qwen3-235B}\par\vspace{1pt}%
    \begin{minipage}{0.24\linewidth}\centering\includegraphics[width=\linewidth]{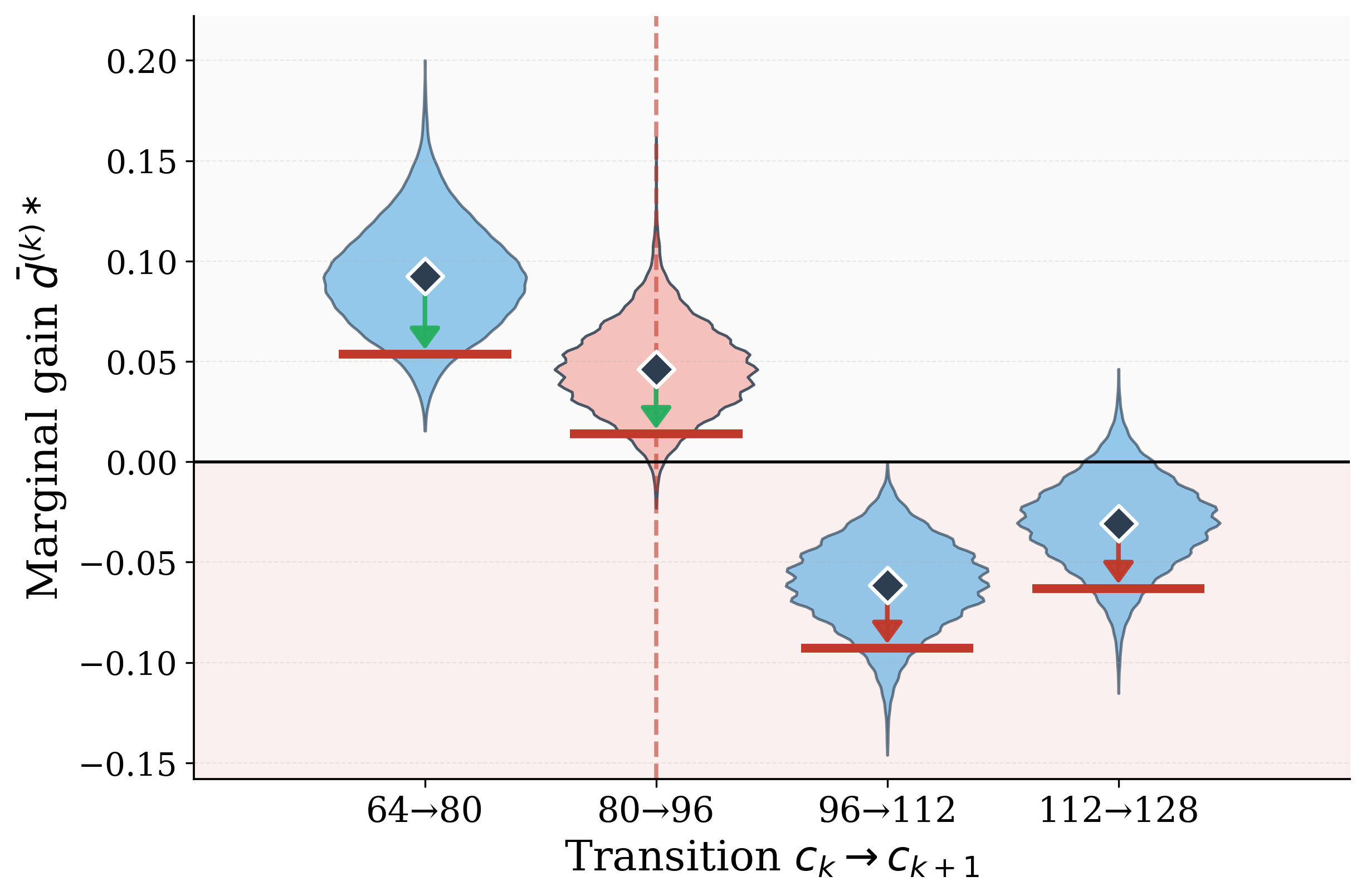}\end{minipage}\hfill
    \begin{minipage}{0.24\linewidth}\centering\includegraphics[width=\linewidth]{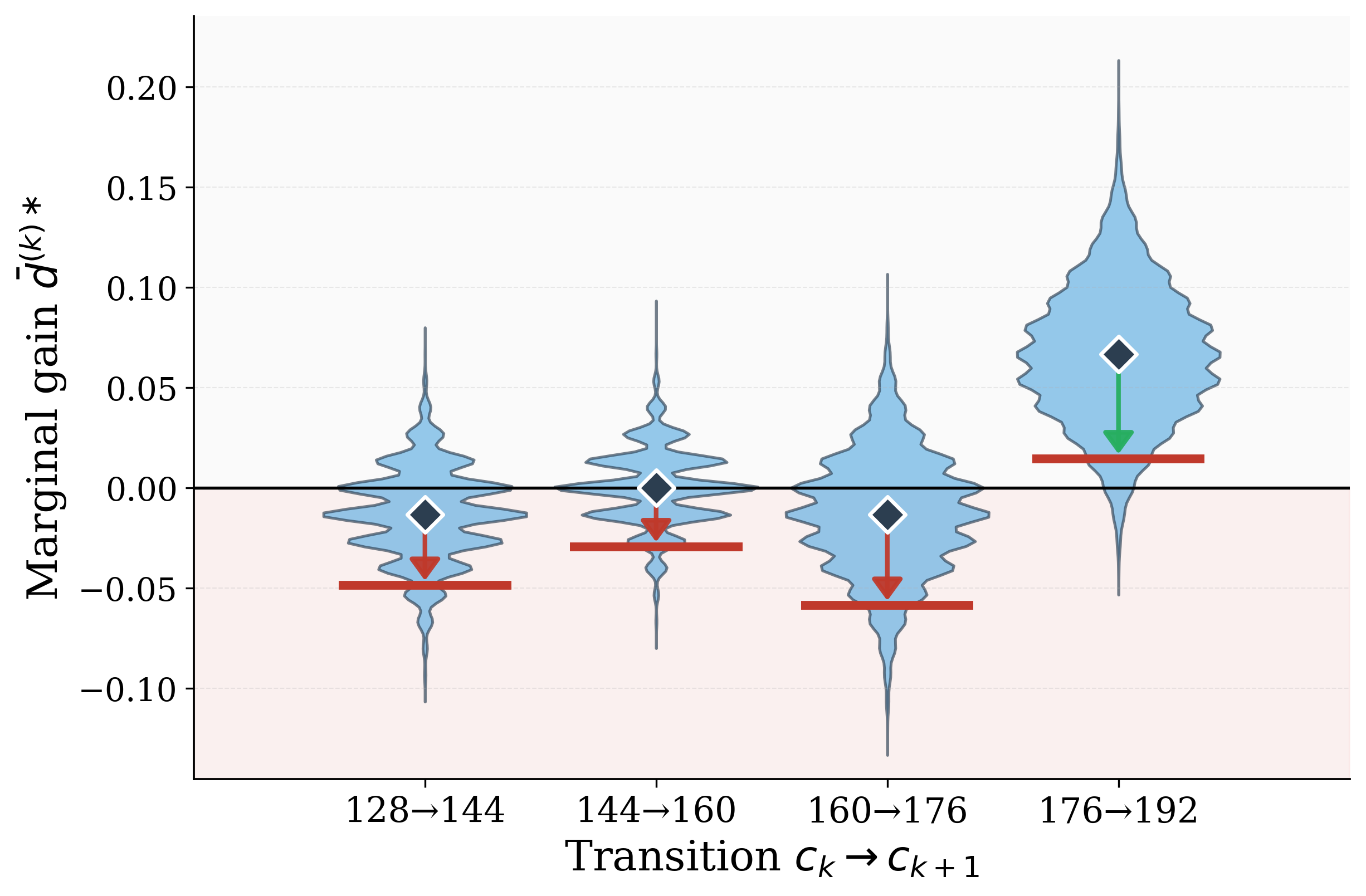}\end{minipage}\hfill
    \begin{minipage}{0.24\linewidth}\centering\includegraphics[width=\linewidth]{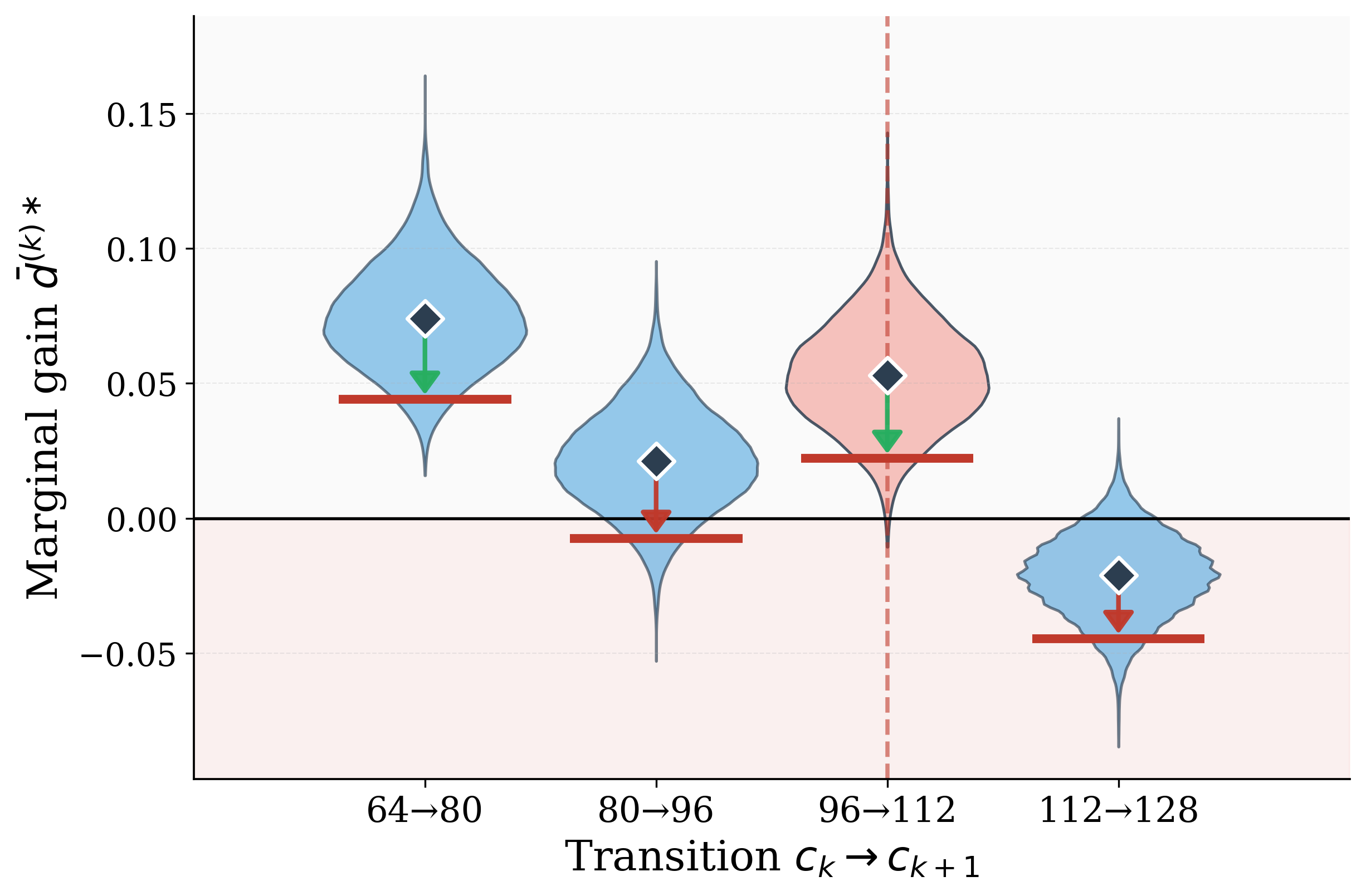}\end{minipage}\hfill
    \begin{minipage}{0.24\linewidth}\centering\includegraphics[width=\linewidth]{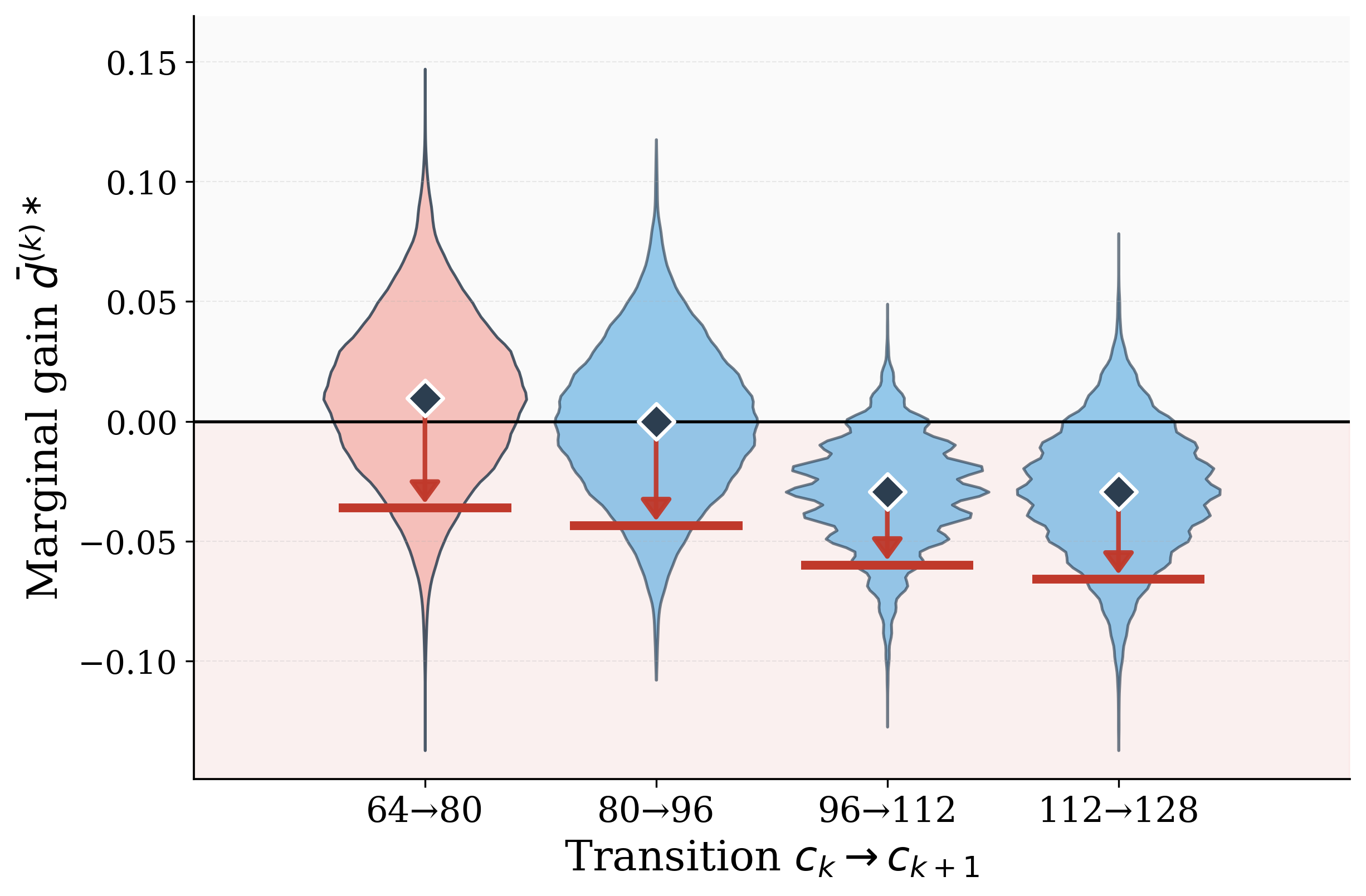}\end{minipage}
    \vspace{4pt}
    \noindent\textbf{\small Qwen3-Next}\par\vspace{1pt}%
    \begin{minipage}{0.24\linewidth}\centering\includegraphics[width=\linewidth]{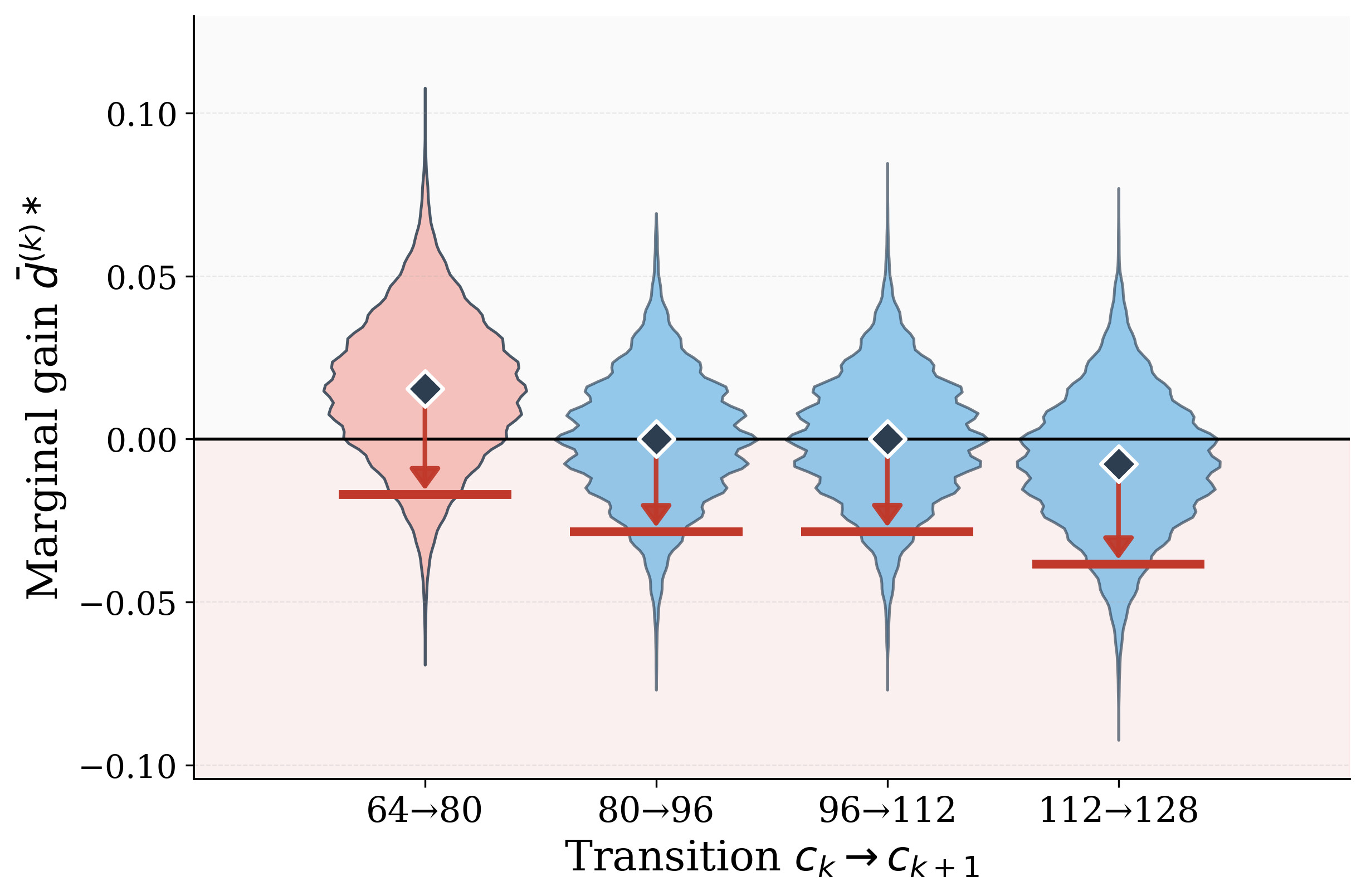}\end{minipage}\hfill
    \begin{minipage}{0.24\linewidth}\centering\includegraphics[width=\linewidth]{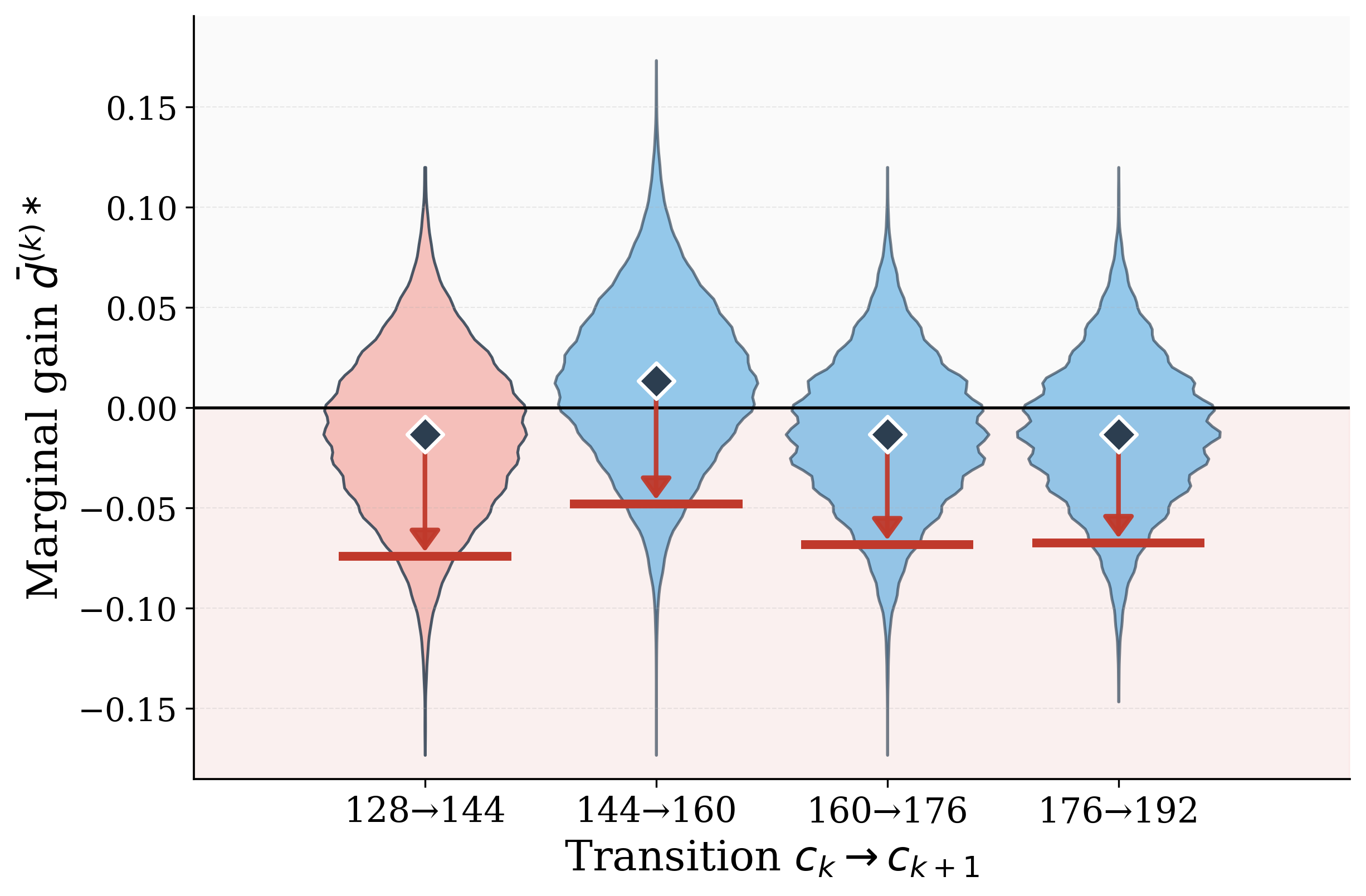}\end{minipage}\hfill
    \begin{minipage}{0.24\linewidth}\centering\includegraphics[width=\linewidth]{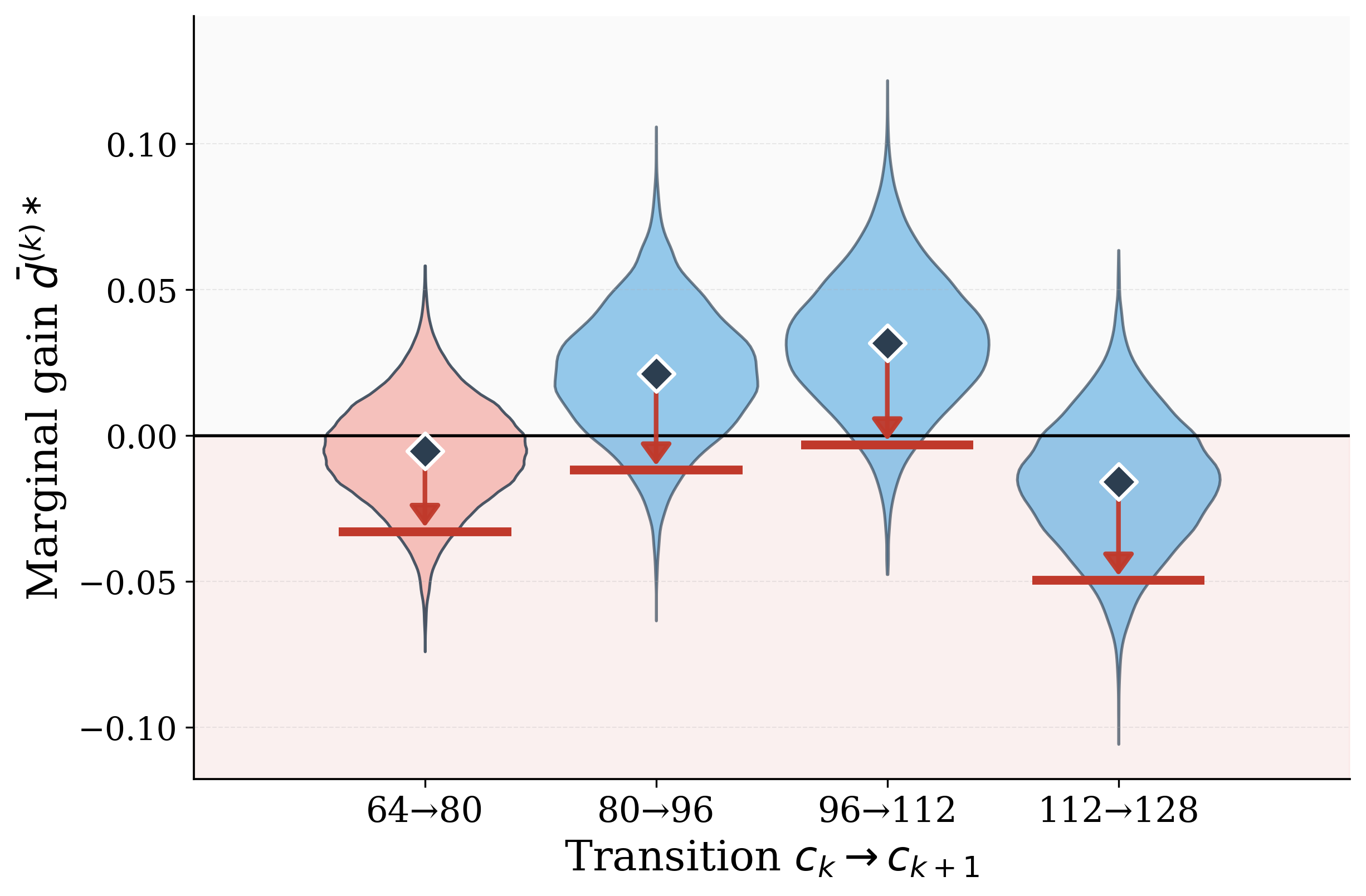}\end{minipage}\hfill
    \begin{minipage}{0.24\linewidth}\centering\includegraphics[width=\linewidth]{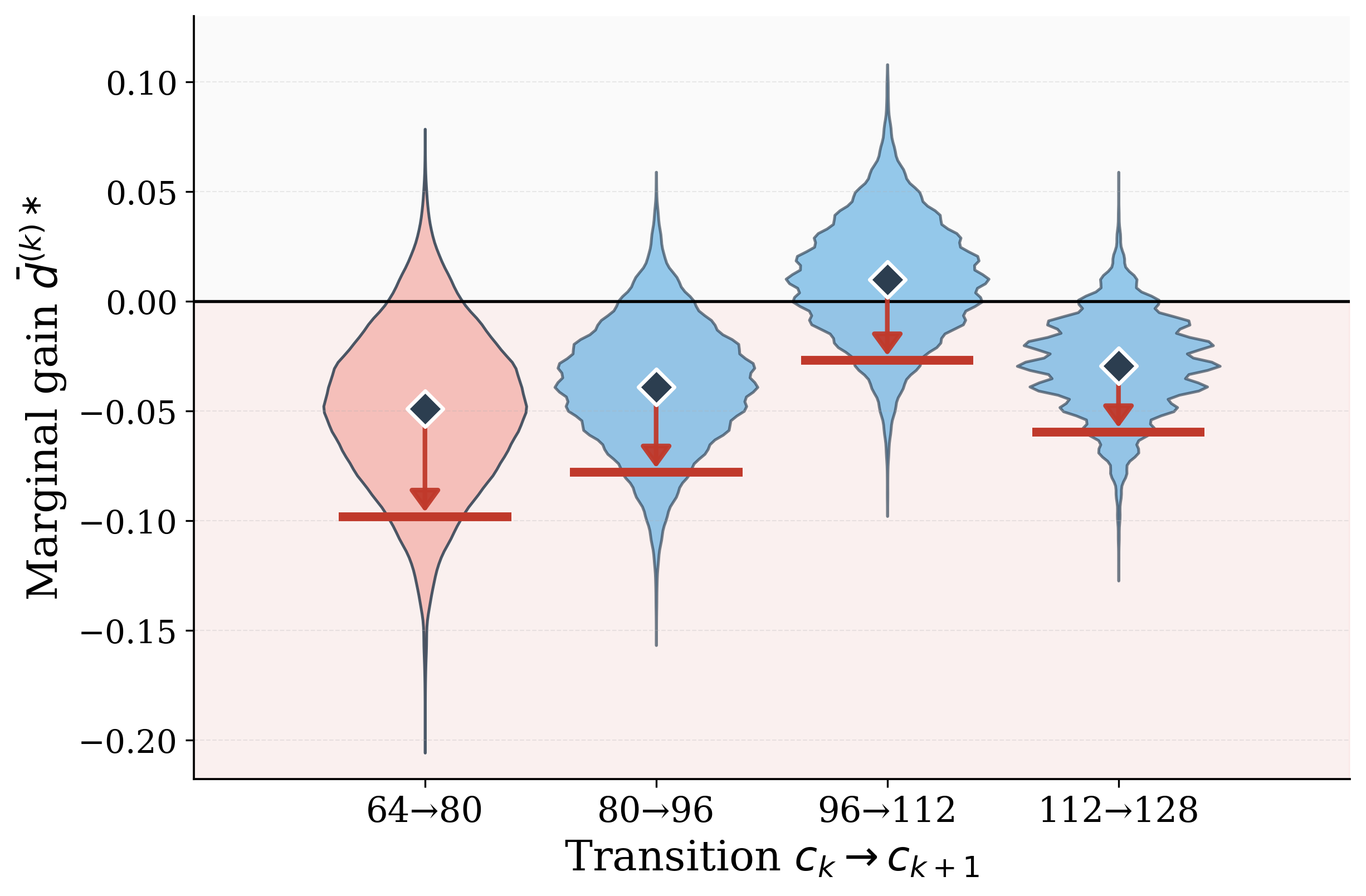}\end{minipage}
    \vspace{4pt}
    \caption{\textbf{Paired-bootstrap distributions for all model--domain pairs.} Each violin depicts the resampled distribution of the marginal gain $\bar{d}^{(k)*}$ across $B{=}10{,}000$ bootstrap iterations. The diamond marks the observed $\bar{d}^{(k)}$, and the red tick indicates the FPC-corrected lower bound $C_k$. Blue violins: transitions before the plateau; pink violins: $C_k \leq 0$.}
    \label{fig:plateau_bootstrap}
\end{figure}

\subsection{Agentic Benchmarks}\label{appendix:benchmarks_details}

\paragraph{Coding} We include tasks from SWE-Bench Verified~\citep{openai_swebench_verified_2024} and Terminal Bench for the code domain. SWE-Bench Verified consists of real-world GitHub issues that require the model to analyze and propose concrete bug fixes, with evaluation performed through automated test pass rates. Terminal Bench assesses a model’s ability to solve problems within a terminal environment, requiring not only command-line common sense but also the ability to plan and reason over long user instructions. It examinzes the score by checking whether the model achieves the expected final state.

\paragraph{Reason} For this domain, we adopt MathHay~\citep{wang2024mathhay} as our data source. MathHay collects mathematical information grounded in real web documents. For each group of related documents, specific pieces of information are extracted and linked to form a query. The benchmark then constructs a long-context haystack by inserting these relevant documents into noisy text placed at the beginning, middle, or end of the sequence. This design provides a suitable testbed for evaluating the model's long‑context reasoning ability. One thing to note is that no raw tools are provided in this benchmark.

Most reasoning benchmarks provide all necessary information directly within the query and ask the model to perform complex mathematical or scientific derivations. As a result, they naturally lack an interactive environment that can deliver feedback, making them unsuitable for multi-turn evaluation. However, we can simulate multi-turn interaction by explicitly prompting the model to refine its own answers over additional turns, effectively creating self-reflective reasoning cycles. A key requirement for such evaluation is that the query context must be sufficiently long to allow meaningful iterative improvement. 

\paragraph{Search} For the search domain, we include tasks from BrowseComp~\citep{wei2025browsecomp} and WebVoyager~\citep{he2024webvoyager}, aiming to evaluate a model’s ability to locate accurate information within long, evolving contexts. In contrast to Needle-in-a-Haystack or purely textual retrieval tasks, agentic search requires the model to reason about what information is needed, determine whether the current context is sufficient to answer the query, and decide when additional search steps are necessary. Mind2Web and WebVoyager focus on everyday web-browsing tasks—such as shopping, navigation, and entertainment—and rely on external language models or task-specific agents to assess model performance. BrowseComp, in contrast, presents challenging, hard-to-find information queries that require multi-step investigation across the web. The model’s final answers are compared against expert-curated gold references to produce the evaluation score.

\paragraph{Tool-use} For the tool-use domain, we include tasks from Tau2-Bench~\citep{barres2025tau2} and MCP-Bench~\citep{wang2025mcpbench}, both of which provide extensive tool suites for evaluating a model’s ability to understand, select, and invoke the appropriate tool to solve a problem. Tau2-Bench focuses on customer-service scenarios, where the model acts as a customer-support agent responding to simulated user queries using a set of synthetic tools tailored to the scenario. MCP-Bench, in contrast, offers a collection of real-world tools built on the Model Context Protocol (MCP), requiring models to perform dense tool calling and coordinate across multiple tools to complete complex tasks. 

Both benchmarks require multi-turn interaction with a simulated environment, and the complexity and richness of the task specifications make them nearly unsolvable within a short context window.

While the coding and search domains can be viewed as specific instances of tool use—since they also involve code-execution tools and search APIs—the tool-use domain in General AgentBench significantly broadens the definition of “tools.” It requires models to precisely control tool parameters, reason about which tool to invoke and when, and plan multi-step interactions based on previous own generation and external feedback.

\subsection{Sample Size Justification and Statistical Robustness}\label{appendix:sample_size}

Our benchmark samples a subset of instances from each source dataset. This design balances dataset category diversity against runtime and evaluation API costs: because General AgentBench unifies all tools into a single context, every inference call operates over long contexts, making full-dataset evaluation prohibitively expensive. A detailed cost breakdown is provided in Appendix~\ref{appendix:cost}.

To ensure that our conclusions are statistically robust despite the reduced sample sizes, we apply bootstrap confidence intervals across all models and benchmarks. For each (model, benchmark) configuration, we draw $B{=}10{,}000$ bootstrap resamples over per-task pass/fail outcomes and compute the resampled mean, yielding a score distribution. We report two complementary measures: the \emph{typical CI width} (the 95\% confidence interval of the aggregate score, reflecting estimation precision) and the \emph{bootstrap standard error} (the standard deviation of the resampled means, reflecting stability of the aggregate estimate).

\begin{table}[h]
\centering
\small
\caption{Bootstrap statistics across benchmarks. $N$ denotes the number of sampled instances. Typical CI width and bootstrap standard error (Pass Std) are computed over $10{,}000$ resamples across all evaluated models.}
\label{tab:bootstrap_stats}
\setlength{\tabcolsep}{8pt}
\renewcommand{\arraystretch}{1.1}
\begin{tabular}{lccc}
\toprule
\textbf{Benchmark} & \textbf{$N$} & \textbf{Typical CI Width} & \textbf{Pass Std} \\
\midrule
SWE-Bench       & 49--50 & 0.04--0.08 & 0.02--0.05 \\
Tau2-Bench      & 50     & $\sim$0.18 & 0.02--0.04 \\
Terminal-Bench  & 79     & 0.01--0.09 & 0.01--0.03 \\
MathHay         & 75     & 0.11--0.14 & 0.01--0.03 \\
BrowseComp          & 124    & 0.10--0.12 & 0.01--0.02 \\
MCP-Bench       & 52     & 0.02--0.07 & 0.01--0.03 \\
\bottomrule
\end{tabular}
\end{table}

As shown in Table~\ref{tab:bootstrap_stats}, typical CI widths range from $\pm0.01$ to $\pm0.14$ and bootstrap standard errors remain within $\pm0.01$--$0.05$ across all models and benchmarks. Only Tau2-Bench exhibits a notably wider CI ($\pm0.18$), consistent with its recognized high variance. The consistently narrow bootstrap standard errors, even where CI widths are larger, confirm that the observed performance trends reflect genuine model differences rather than sampling noise.

\subsection{Evaluation on long-context benchmarks} \label{appendix:long_compare}
\input{tables/long_benchmark}

With unified toolsets alone already approaching 64K tokens, the addition of user queries and multi-turn interaction histories can easily push the total context length to nearly 128K tokens. As a result, long-context processing naturally emerges as a core capability required for general agents. However, existing long-context benchmarks differ fundamentally from agentic long-context scenarios. As summarized in Table~\ref{tab:taxonomy}, we identify two key dimensions along which prior benchmarks diverge from agentic settings.

(1) \textbf{Context composition.} Most existing long-context benchmarks are dominated by long-document question answering (QA), where the interaction paradigm remains static and single-turn. In contrast, agentic contexts are inherently heterogeneous: beyond long documents, they include environment feedback (e.g., tool execution results) and the model’s own prior decisions accumulated through multi-turn interactions.

(2) \textbf{Long-output reasoning.} Other established benchmark categories, such as many-shot in-context learning and summarization, involve long inputs but require only relatively short outputs. Agentic tasks, however, demand sustained reasoning over extended contexts, including plan generation, iterative reflection, and explicit tool-call specifications, often resulting in long and structured outputs. While retrieval-based tasks and recent citation-grounded generation benchmarks partially resemble agentic scenarios, they still lack the multi-turn, interactive dynamics essential to true agentic settings.

These fundamental differences suggest that performance measured on prior non-agentic long-context benchmarks does not directly reflect model behavior under agentic interaction. 

\input{tables/previous_long_context}

\subsection{Transferability of Long-Context Abilities from Static to Agentic Benchmarks}\label{appendix:long}

\begin{figure}[h]
        \centering
        \includegraphics[width=0.8\linewidth]{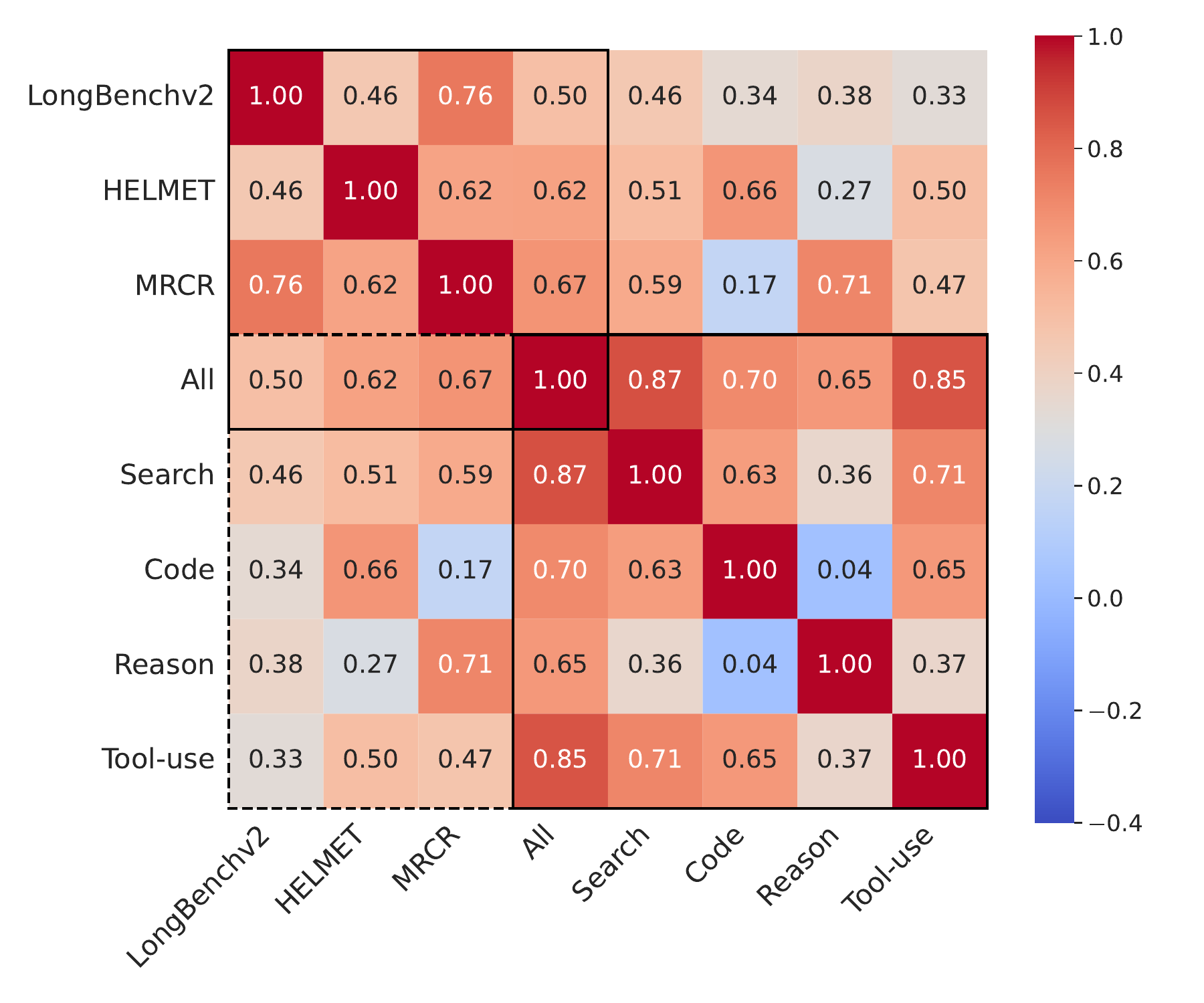}
        \caption{\textbf{Pairwise correlation between static long-context benchmarks and agentic domains in our General AgentBench.} “All” denotes the average performance across the search, code, reasoning, and tool-use domains.}
        \label{fig:corr}
\end{figure}

Long-context processing is a fundamental requirement for general agents, as unified toolsets and accumulated multi-turn interactions can rapidly extend the effective context length during real-world use. However, it remains unclear whether performance measured on existing \emph{static}, single-turn long-context benchmarks meaningfully transfers to \emph{agentic} settings, where context evolves dynamically through interaction and decisions must be made sequentially. To study this question, we evaluate models on several representative static long-context benchmarks and examine how their performance correlates with results on General AgenticBench. We choose LongBench, HELMET, and MRCR here.

\paragraph{LongBench v2}
LongBench v2 is an updated and expanded version of LongBench, designed to evaluate language models under diverse long-context understanding tasks across multiple domains. It covers a broad set of settings, including long-document question answering, multi-document reasoning, summarization, code understanding, and many-shot in-context learning. Compared to the original LongBench, v2 increases both context length and task diversity, and introduces harder examples that stress retrieval accuracy and reasoning robustness under long inputs. Despite its breadth, LongBench v2 remains fundamentally \emph{single-turn} and \emph{static} in interaction structure. All necessary information is provided upfront, and models are evaluated on their ability to extract, aggregate, or reason over relevant spans within a fixed context window. Outputs are typically short and final, without iterative refinement or environment feedback. As a result, LongBench v2 primarily measures long-context comprehension and retrieval, rather than the dynamic decision-making, planning, and self-conditioning behaviors required in agentic multi-turn settings.

\paragraph{HELMET}
HELMET is a holistic benchmark for evaluating long-context language models across a curated set of real-world and synthetic tasks. It emphasizes robustness, faithfulness, and information utilization under long inputs, and includes task categories such as long-document QA, summarization, citation-grounded generation, and structured information extraction. HELMET is particularly focused on evaluating whether models can correctly attribute claims to source documents and avoid hallucinations when operating over extended contexts.

\paragraph{MRCR}
MRCR focuses on evaluating a model’s ability to perform multi-round coreference and entity tracking under long contexts. The benchmark constructs documents containing repeated, interleaved references to entities across long spans, requiring the model to resolve pronouns, aliases, and implicit references over multiple turns or segments. Tasks are designed to test memory persistence and consistency, particularly in scenarios where earlier mentions are far removed from later queries.

Table~\ref{tab:long_context_baselines} shows the evaluation results on these three benchmarks. Across ten models, we compute Pearson correlations over pairwise absolute performance differences between static long-context benchmarks and the four agentic domains. As shown in Figure~\ref{fig:corr}, static benchmarks exhibit consistently weak correlation with agentic performance overall, indicating limited transferability from static long-context ability to agentic long-context reasoning. A moderate correlation is observed between MRCR and the reasoning domain, which is expected: reasoning tasks primarily involve extracting and computing over long documents without tool interaction, closely aligning with the characteristics of MRCR. Other agentic domains show substantially lower alignment. In particular, coding and tool-use exhibit minimal correlation with static benchmarks, suggesting that agentic performance depends not only on long-context comprehension, but also on dynamic decision-making and precise execution. These results highlight a fundamental gap between static, single-turn long-context evaluation and the requirements of realistic agentic settings.


\subsection{Extending $K$ to 8 and 16 in Parallel Scaling}\label{appendix:k816}

\begin{figure}[h]
        \centering
        \includegraphics[width=1.0\linewidth]{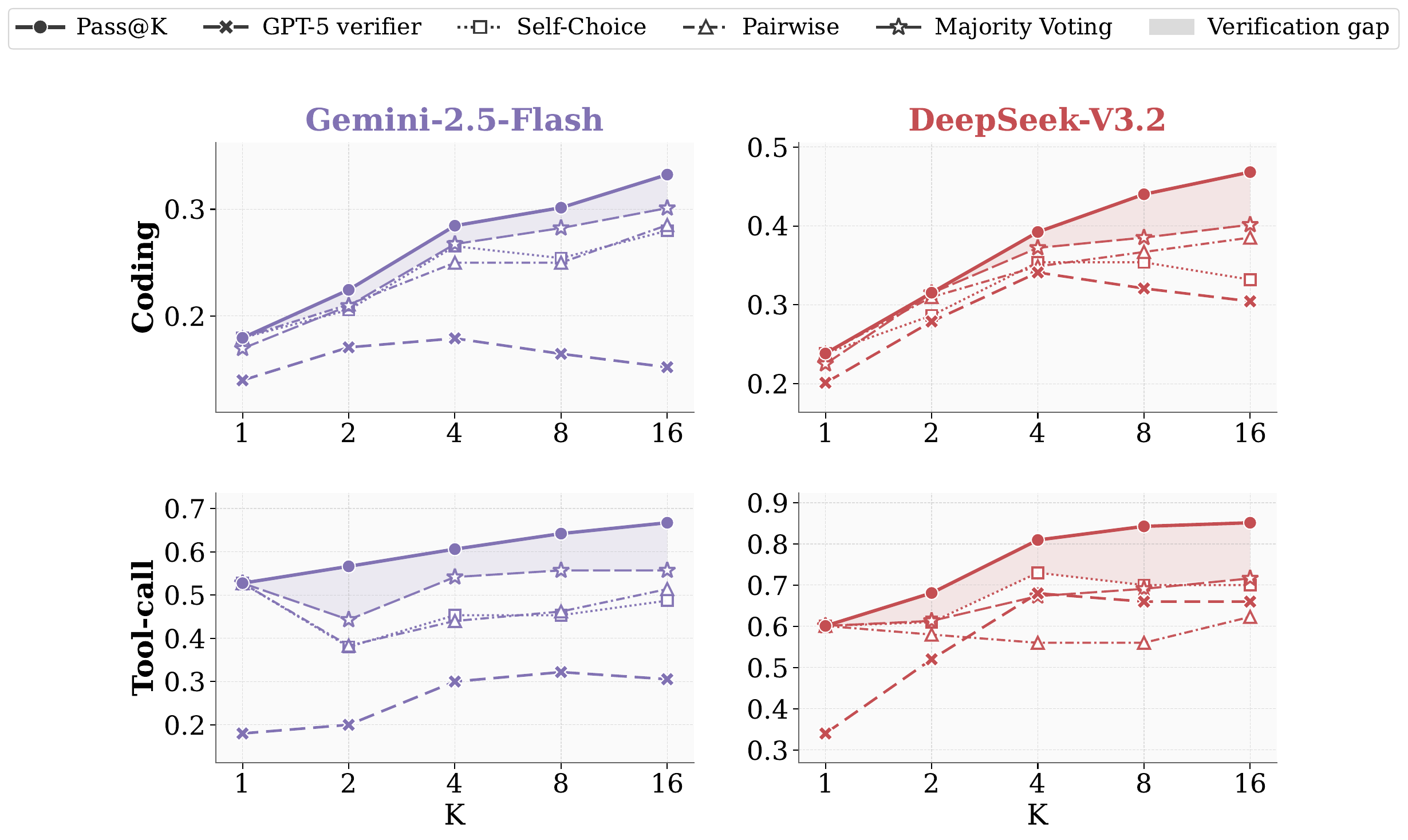}
        \caption{\textbf{Extended parallel scaling to $K{=}8$ and $K{=}16$.} We evaluate Gemini-2.5-Flash and DeepSeek-V3.2 on coding and tool-call domains with three commitment strategies and GPT-5 as an external verifier. The verification gap persists across all $K$ values; majority voting growth decelerates while pairwise selection shows an accelerating trend.}
        \label{fig:k816}
\end{figure}

We extend parallel scaling to $K=8$ and $K=16$ on the coding and tool-call domains using Gemini-2.5-Flash and DeepSeek-V3.2, with all three commitment strategies (majority voting, pointwise, and pairwise) as well as GPT-5 as an external verifier. As shown in Figure~\ref{fig:k816}, the pass@$K$ scores continue to increase with larger $K$, confirming that broader sampling consistently expands the solution space. However, the verification gap persists across all values of $K$: none of the commitment strategies or the external verifier can close the gap to the oracle pass@$K$. Notably, as $K$ grows, majority voting---the strongest commitment strategy at smaller $K$---begins to plateau, with diminishing marginal gains at $K{=}8$ and $K{=}16$. In contrast, pairwise selection, while achieving lower absolute accuracy than majority voting, exhibits an accelerating growth trend at larger $K$, suggesting that comparative evaluation may scale more favorably than independent judgment when the candidate pool is large. These results further support our claim that parallel scaling is fundamentally bottlenecked by the verification gap: models can sample correct solutions with increasing $K$, but reliably selecting them remains an open challenge.

\subsection{Prompt Sensitivity of Verification}\label{appendix:prompt_sensitivity}

A potential concern with our parallel scaling analysis is whether the verification results are sensitive to the choice of judge or prompt variation. To assess this, we conduct a stability analysis on the Gemini-2.5-Flash model in the coding domain. We run each verification strategy three times independently and report the per-run accuracy along with the standard deviation across runs.

\begin{table}[h]
\centering
\small
\caption{Verification accuracy across three independent runs (Gemini-2.5-Flash, coding domain). Low standard deviations confirm that both self-judgment and external verification are stable across runs.}
\label{tab:prompt_sensitivity}
\setlength{\tabcolsep}{6pt}
\renewcommand{\arraystretch}{1.1}
\begin{tabular}{lcccc}
\toprule
\textbf{Verifier} & \textbf{Run 1} & \textbf{Run 2} & \textbf{Run 3} & \textbf{Std} \\
\midrule
Self-judge (pointwise) & 0.1795 & 0.1823 & 0.1705 & 0.0050 \\
Self-judge (pairwise)  & 0.1795 & 0.1795 & 0.1672 & 0.0058 \\
GPT-5 (pointwise)      & 0.1397 & 0.1471 & 0.1397 & 0.0035 \\
\bottomrule
\end{tabular}
\end{table}

As shown in Table~\ref{tab:prompt_sensitivity}, all three verification strategies exhibit low standard deviations ($\leq 0.006$), indicating that the judge outputs are stable and consistent across runs. Notably, the variance introduced by the verification stage is substantially lower than the inherent stochasticity of generation itself, suggesting that, under a fixed evaluation protocol, the verification outcomes are relatively robust even though the underlying trajectories exhibit high variance. This confirms that the verification gap reported in our main results is not an artifact of prompt sensitivity or stochastic variation in the judging process.

\subsection{Disentangling Tool Confusion from Context Overhead}\label{appendix:disentangle}

A natural question is whether the performance degradation observed in the general-agent setting stems from the ``generality'' of the task distribution itself, or merely from the longer context and additional tool descriptions that accompany a unified framework. We note that long context and realism are inherently coupled in our setting: a unified evaluation necessarily exposes the agent to a larger tool registry, which in turn produces longer context. Rather than attempting to fully separate these factors, we provide a trace-level analysis that characterizes which mechanism dominates for each benchmark.

\paragraph{Out-of-domain tool call analysis.} We analyze the proportion of tool calls directed at tools outside each benchmark's original domain across all six benchmarks under the unified setting, using Gemini-2.5-Flash as the evaluation model. Table~\ref{tab:ood_tool_calls} summarizes the results.

\begin{table}[h]
\centering
\small
\caption{Proportion of out-of-domain tool calls per benchmark under the unified setting (Gemini-2.5-Flash).}
\label{tab:ood_tool_calls}
\setlength{\tabcolsep}{8pt}
\renewcommand{\arraystretch}{1.1}
\begin{tabular}{lc}
\toprule
\textbf{Benchmark} & \textbf{Out-of-domain tool calls (\%)} \\
\midrule
SWE-Bench      & $\sim$5  \\
Terminal-Bench  & $\sim$9  \\
MCP-Bench       & $\sim$8  \\
BrowseComp      & $\sim$13 \\
Tau2-Bench      & $\sim$40 \\
WebVoyager      & $\sim$50 \\
\bottomrule
\end{tabular}
\end{table}

The results reveal two distinct failure modes. When a model frequently calls tools outside its task domain (e.g., Tau2-Bench at ${\sim}40\%$ and WebVoyager at ${\sim}50\%$), the performance drop is primarily driven by \emph{tool confusion}: the model is actively misled by the presence of tools from other domains, a genuine failure mode in realistic deployments where agents must operate alongside tools they should ignore. Conversely, when a model rarely invokes out-of-domain tools despite having access to the full unified pool (e.g., SWE-Bench at ${\sim}5\%$, Terminal-Bench at ${\sim}9\%$, MCP-Bench at ${\sim}8\%$), the performance drop is more plausibly attributed to \emph{context overhead} from the larger tool registry.

Both mechanisms are inherent to realistic deployment: a real-world general-purpose agent will always face a large tool registry. We view the ability to distinguish and handle these two failure modes---tool confusion versus context overhead---as an additional capability requirement for general agents.

\paragraph{Ablation on tool registry size.} To further isolate the effect of context overhead, we evaluate Gemini-2.5-Flash on the tool-call domain while progressively reducing the tool registry size through three cumulative strategies: (1) replacing detailed tool descriptions with concise one-line summaries, (2) removing descriptions entirely and retaining only tool names and parameter signatures, and (3) filtering out tools that are irrelevant to the evaluated benchmarks (i.e., tools that are never invoked in any correct trajectory). These steps reduce the registry from 42K tokens (original) to 24K tokens. As shown in Table~\ref{tab:registry_ablation}, performance remains largely stable across all registry sizes, suggesting that context overhead alone accounts for a small fraction of the observed degradation.

\begin{table}[h]
\centering
\small
\caption{Performance of Gemini-2.5-Flash on the tool-call domain as the tool registry size varies.}
\label{tab:registry_ablation}
\setlength{\tabcolsep}{8pt}
\renewcommand{\arraystretch}{1.1}
\begin{tabular}{lcccc}
\toprule
\textbf{Tool registry size} & \textbf{24K} & \textbf{32K} & \textbf{36K} & \textbf{42K (Original)} \\
\midrule
Gemini-2.5-Flash & 0.3550 & 0.3594 & 0.3572 & 0.3478 \\
\bottomrule
\end{tabular}
\end{table}

%% file: tables/long_benchmark.tex
\begin{table*}[ht]
\centering
\small
\caption{\textbf{Taxonomy of Existing Long-Context Benchmarks.}}
\label{tab:taxonomy}
\begin{tabularx}{\textwidth}{l l X}
\toprule
\textbf{Task Category} & \textbf{Sub-category} & \textbf{Representative Benchmarks / Papers} \\ 
\midrule
\textbf{Long Document QA} & Single/Multi-Doc Understanding & NarrativeQA \cite{kovcisky2018narrativeqa}, 2WikiMultihopQA \cite{ho-etal-2020-constructing}, Loong \cite{wang2024leave} \\
\cmidrule{2-3}
& Long Dialogue Understanding & MeetingBank \cite{hu2023meetingbank}, CharacterChat \cite{tu2023characterchat} \\
\cmidrule{2-3}
& Code Understanding & RepoBench \cite{liu2023repobench}, CrossCodeEval \cite{ding2023crosscodeeval} \\
\cmidrule{2-3}
& Structured Data Understanding & TabFact \cite{chen2019tabfact}, WikiTableQuestions \cite{kweon2023open}, LongBench \cite{bai2024longbench} (L-Data)\\
\midrule
\textbf{Summarization} & Global Information Aggregation & GovReport \cite{huang2021efficient}, Multi-News \cite{fabbri2019multi}, ZeroSCROLLS \cite{shaham2023zeroscrolls} \\
\midrule
\textbf{Many-Shot} & In-Context Learning & MIR-Bench \cite{yan2025mir}, Many-Shot ICL \cite{agarwal2024many} \\
\midrule
\textbf{Retrieval} & Key-Value Retrieval & Needle In A Haystack \cite{kamradt_needlehaystack_2023}, RULER \cite{hsieh2024ruler}, MRCR \cite{openai_mrcr_2025}, BABILong \citeyear{kuratov2024babilong} \\
\cmidrule{2-3}
& Retrieval-Augmented Gen (RAG) & RAGBench \cite{friel2024ragbench}, RGB \cite{chen2024benchmarking}, RECALL \cite{liu2023recall} \\
\midrule
\textbf{Reranking} & Passage Reranking & BEIR \cite{thakur2021beir}, MTEB \cite{muennighoff2023mteb} (Reranking Selection) \\
\midrule
\textbf{Citation Gen} & Generation with Citations & ALCE \cite{gao2023enabling} (ASQA/ELI5), LongCite \cite{zhang2025longcite} \\
\midrule
\rowcolor[gray]{0.9} \textbf{Agentic } & \textbf{Multi-turn Reasoning} & SWE-Bench \cite{jimenez2023swebench}, BrowseComp \cite{wei2025browsecomp}, OSWorld \cite{xie2024osworld}, \textbf{General AgentBench (This Work)} \\
\bottomrule
\end{tabularx}
\end{table*}

%% file: tables/previous_long_context.tex
\begin{table}[ht]
\centering
\caption{Performance on long-context benchmarks. We report scores across three established benchmarks focusing on retrieval and single-turn comprehension. \textbf{Bold} marks the best and \underline{underline} the second-best score in each column.}
\vspace{1em}
\label{tab:long_context_baselines}
\setlength{\tabcolsep}{8pt}
\renewcommand{\arraystretch}{1.15}
\setlength{\aboverulesep}{0.15ex}
\setlength{\belowrulesep}{0.15ex}
\begin{tabular}{ l ccc c }
\toprule
\textbf{Models} & \textbf{LongBench} & \textbf{HELMET} & \textbf{MRCR} & \textbf{Avg} \\
\midrule
\rowcolor{gray!10} \multicolumn{5}{l}{\textit{Open-Source}} \\
GPT-OSS-120B      & 47.8 & 12.9 & 32.8 & 31.2 \\
Qwen3-235B-A22B   & 58.3 & 40.8 & 40.6 & 46.6 \\
Qwen3-Next        & 53.1 & 26.3 & 27.9 & 35.8 \\
DeepSeek-V3.2     & 50.3 & \underline{48.0} & 33.2 & 43.8 \\
DeepSeek-R1       & 58.3 & 36.9 & 39.2 & 44.8 \\
\midrule
\rowcolor{gray!10} \multicolumn{5}{l}{\textit{Proprietary}} \\
Gemini 2.5-Flash  & 62.1 & 27.8 & 67.4 & 52.4 \\
Gemini 2.5-Pro    & \underline{63.3} & \textbf{63.1} & \textbf{80.0} & \textbf{68.8} \\
Claude Haiku 4.5  & 55.3 & 50.4 & 33.3 & 46.3 \\
Claude Sonnet 4.5 & 61.8 & 49.2 & 35.5 & 48.8 \\
GPT-5             & \textbf{64.6} & 51.6 & \underline{79.1} & \underline{65.1} \\
\bottomrule
\end{tabular}
\end{table}

%% file: src/appendix_cost.tex
\section{Estimated Cost}\label{appendix:cost}

\subsection{Pricing and accounting.}
We report an estimated API budget for reproducing our evaluation under three settings:
(i) \textbf{general (default context)} evaluation of each model on each domain once,
(ii) \textbf{parallel scaling} that samples multiple independent trajectories per query, and
(iii) \textbf{sequential scaling} that extends the interaction horizon over multiple steps.

All prices are normalized as USD per 1M tokens and are taken from the corresponding model providers at the time of running the experiments. Due to uncertainty about the model provider's caching mechanism and evaluation intervals, we will use only the input unit price in our estimates, even if the provider supplies a cache unit price. 

We compute cost by aggregating token usage from execution logs, using the provider-reported token accounting:
\textbf{input tokens} (prompt + tool outputs + intermediate messages fed back to the model),
and \textbf{output tokens} (model-generated tokens).
For each model, the total cost is estimated as:
\[
\text{Cost} =
p_{\text{in}}\cdot \tfrac{T_{\text{in}}}{10^6}
+ p_{\text{out}}\cdot \tfrac{T_{\text{out}}}{10^6},
\]
where $p_{\text{in}}, p_{\text{out}}$ denote the unit prices, and
$T_{\text{in}}, T_{\text{out}}$ are the measured token counts.

\subsection{Model cost}

Table~\ref{tab:lite-version-cost} lists the unit API prices used in our cost estimation. Prices for input, cached input, and output are reported in USD per 1M tokens. Tables~\ref{tab:parallel-scaling-cost}--\ref{tab:general-setting-cost} summarize the aggregated evaluation cost per model under parallel scaling, sequential scaling, and the general (default context) setting, respectively.
Each entry corresponds to running the full benchmark split of that dataset/domain under the specified protocol and then summing costs across all queries.

\begin{table}[h]
\centering
\small
\begin{tabular}{lccc}
\toprule
\textbf{Model} & \textbf{Input} & \textbf{Cached Input} & \textbf{Output} \\
\midrule
Gemini-2.5-Flash      & 0.30 & 0.03  & 2.50 \\
Gemini-2.5-Pro        & 1.25 & 0.125 & 10.00 \\
GPT-5                 & 1.25 & 0.125 & 10.00 \\
Claude-Haiku 4.5      & 1.00   & 0.50    & 5.00 \\
Claude-Sonnet 4.5     & 3.00 & 3.75  & 15.00 \\
\midrule
gpt-oss-120B           & 0.15 & 0.075    & 0.60 \\
Deepseek-R1           & 1.35 & --    & 5.40 \\
Deepseek-V3.2         & 0.28 & --    & 0.42 \\
Qwen3-235B            & 0.22 & 0.11    & 0.88 \\
Qwen3-Next            & 0.15 & --    & 1.50 \\
\bottomrule
\end{tabular}
\vspace{0.8em}
\caption{Unit API prices (USD per 1M tokens) used in our cost estimation.}
\label{tab:lite-version-cost}
\end{table}

\begin{table}[htbp]
\centering
\footnotesize
\begin{tabular}{lrrrrrrr}
\toprule
\textbf{Model} & \textbf{Search} & \textbf{MathHay} & \textbf{SWEBench} & \textbf{MCPBench} & \textbf{Tau2Bench} & \textbf{TerminalBench} & \textbf{Total} \\
\midrule
Gemini-2.5-Flash & \$193 & \$70.9 & \$12719 & \$122 & \$62.2 & \$826 & \$13993 \\
DeepSeek-R1 & \$5188 & \$157 & \$3675 & \$492 & \$202 & \$1505 & \$11218 \\
DeepSeek-V3.2 & \$369 & \$38.8 & \$7.80 & \$88.4 & \$54.4 & \$412 & \$970 \\
Qwen3-235B & \$1254 & \$41.4 & \$1286 & \$120 & \$55.5 & \$409 & \$3166 \\
Qwen3-Next & \$25.3 & \$9.52 & \$3.38 & \$21.2 & \$14.9 & \$154 & \$229 \\
\midrule
\textbf{Total} & \$7028 & \$317 & \$17692 & \$843 & \$389 & \$3307 & \$29576 \\
\bottomrule
\end{tabular}
\vspace{1em}
\caption{Cost for evaluating models under parallel scaling setting (USD)}
\label{tab:parallel-scaling-cost}
\end{table}

\begin{table}[htbp]
\centering
\footnotesize
\begin{tabular}{lrrrrrrr}
\toprule
\textbf{Model} & \textbf{Search} & \textbf{MathHay} & \textbf{SWEBench} & \textbf{MCPBench} & \textbf{Tau2Bench} & \textbf{TerminalBench} & \textbf{Total} \\
\midrule
Gemini-2.5-Flash & \$5588 & \$369 & \$2024 & \$568 & \$852 & \$1870 & \$11271 \\
DeepSeek-R1 & \$1267 & \$654 & \$892 & \$931 & \$1088 & \$782 & \$5614 \\
DeepSeek-V3.2 & \$902 & \$333 & \$191 & \$79.1 & \$52.3 & \$356 & \$1913 \\
Qwen3-235B & \$1370 & \$238 & \$436 & \$716 & \$162 & \$689 & \$3610 \\
Qwen3-Next & \$442 & \$219 & \$392 & \$152 & \$251 & \$529 & \$1985 \\
\midrule
\textbf{Total} & \$9568 & \$1814 & \$3935 & \$2445 & \$2405 & \$4225 & \$24392 \\
\bottomrule
\end{tabular}
\vspace{1em}
\caption{Cost for evaluating models under sequential scaling setting (USD)}
\label{tab:sequential-scaling-cost}
\end{table}

\begin{table}[htbp]
\centering
\footnotesize
\begin{tabular}{lrrrrrrr}
\toprule
\textbf{Model} & \textbf{Search} & \textbf{MathHay} & \textbf{SWEBench} & \textbf{MCPBench} & \textbf{Tau2Bench} & \textbf{TerminalBench} & \textbf{Total} \\
\midrule
Gemini-2.5-Pro & \$47.5 & \$18.6 & \$3253 & \$31.3 & \$15.2 & \$203 & \$3569 \\
GPT-5 & \$87.5 & \$14.2 & \$146 & \$21.0 & \$12.4 & \$60.9 & \$342 \\
Claude-Haiku-4.5 & \$304 & \$10.9 & \$312 & \$29.3 & \$14.1 & \$106 & \$776 \\
Claude-Sonnet-4.5 & \$1248 & \$40.9 & \$926 & \$129 & \$52.0 & \$376 & \$2772 \\
OpenAI-oss-120B & \$1.44 & \$1.49 & \$0.52 & \$0.50 & \$1.35 & \$0.40 & \$5.70 \\
DeepSeek-V3.2 & \$96.8 & \$10.0 & \$1.92 & \$22.0 & \$14.0 & \$102 & \$247 \\
Qwen3-Next & \$6.05 & \$2.42 & \$0.88 & \$5.51 & \$3.60 & \$38.2 & \$56.7 \\
\midrule
\textbf{Total} & \$1791 & \$98.5 & \$4640 & \$239 & \$113 & \$886 & \$7768 \\
\bottomrule
\end{tabular}
\vspace{1em}
\caption{Cost for evaluating models under general (default context) setting (USD)}
\label{tab:general-setting-cost}
\end{table}

%% file: src/appendix_detailed_numerical_results.tex
\subsection{Detailed Numerical Results for Main Figures}\label{appendix:detailed_numerical}

For reproducibility and to facilitate future comparisons, we provide the full numerical data underlying the three key experimental figures in the main text. \textbf{Table~\ref{tab:detailed_sequential_scaling}} corresponds to \textbf{Figure~\ref{fig:sequential_tts}}, reporting accuracy of all five models across four domains at each context budget. \textbf{Table~\ref{tab:detailed_ablation}} corresponds to \textbf{Figure~\ref{fig:ablation_strategies}}, comparing three sequential scaling strategies on Qwen3-235B and Gemini-2.5-Flash for Tool-call and Coding. \textbf{Tables~\ref{tab:detailed_vg_search}--\ref{tab:detailed_vg_tooluse}} correspond to \textbf{Figure~\ref{fig:parallel_scaling}}. For every (domain, model) pair they list Pass@$K$ alongside four commitment strategies at $K \in \{1,2,3,4\}$, enabling precise quantification of the verification gap.

\input{tables/detailed_results_sequential_scaling_latex}
\input{tables/detailed_results_ablation}
\input{tables/detailed_results_verification_gap}

%% file: tables/detailed_results_sequential_scaling_latex.tex
\begin{table}[h]
\centering
\footnotesize
\caption{\textbf{Detailed numerical results for sequential scaling (Figure~\ref{fig:sequential_tts}).}
Accuracy of 5 models across 4 domains at increasing context budgets.}
\label{tab:detailed_sequential_scaling}
\setlength{\tabcolsep}{5pt}
\renewcommand{\arraystretch}{1.10}
\setlength{\aboverulesep}{0.15ex}
\setlength{\belowrulesep}{0.15ex}
\begin{tabular}{@{} l ccccc @{}}
\toprule
\textbf{Context} & \textbf{DeepSeek-R1} & \textbf{Qwen3-Next} & \textbf{Qwen3-235B} & \textbf{DeepSeek-V3.2} & \textbf{Gemini2.5-Flash} \\
\midrule
\rowcolor{gray!10} \multicolumn{6}{@{}l}{\textit{Search}} \\
\rowcolor{gray!5}  64k  & 0.2060 & 0.2186 & 0.0510 & 0.0854 & 0.1558 \\
\rowcolor{gray!10} 80k  & 0.1210 & 0.2111 & 0.1256 & 0.2663 & 0.1658 \\
\rowcolor{gray!15} 96k  & 0.0550 & 0.2312 & 0.1457 & 0.2410 & 0.1759 \\
\rowcolor{gray!20} 112k & 0.1260 & 0.2663 & 0.2010 & 0.2320 & 0.1759 \\
\rowcolor{gray!25} 128k & 0.1260 & 0.2500 & 0.1809 & 0.2320 & 0.1759 \\
\midrule
\rowcolor{gray!10} \multicolumn{6}{@{}l}{\textit{Code}} \\
\rowcolor{gray!5}  64k  & 0.1400 & 0.0846 & 0.1385 & 0.1538 & 0.1769 \\
\rowcolor{gray!10} 80k  & 0.1800 & 0.1000 & 0.2308 & 0.0923 & 0.2077 \\
\rowcolor{gray!15} 96k  & 0.1600 & 0.1000 & 0.2769 & 0.2154 & 0.2231 \\
\rowcolor{gray!20} 112k & 0.1200 & 0.1000 & 0.2154 & 0.2385 & 0.2000 \\
\rowcolor{gray!25} 128k & 0.1200 & 0.0923 & 0.1843 & 0.1692 & 0.2000 \\
\midrule
\rowcolor{gray!10} \multicolumn{6}{@{}l}{\textit{Tool-use}} \\
\rowcolor{gray!5}  64k  & 0.3618 & 0.4704 & 0.4585 & 0.4832 & 0.4888 \\
\rowcolor{gray!10} 80k  & 0.3968 & 0.4201 & 0.4738 & 0.4344 & 0.3376 \\
\rowcolor{gray!15} 96k  & 0.4716 & 0.3830 & 0.4713 & 0.4650 & 0.3294 \\
\rowcolor{gray!20} 112k & 0.4773 & 0.3908 & 0.4407 & 0.3211 & 0.3239 \\
\rowcolor{gray!25} 128k & 0.4331 & 0.3581 & 0.4160 & 0.3170 & 0.3032 \\
\midrule
\rowcolor{gray!10} \multicolumn{6}{@{}l}{\textit{Reason (context range: 128k--192k)}} \\
\rowcolor{gray!5}  128k & 0.3470 & 0.4933 & 0.3600 & 0.1730 & 0.4267 \\
\rowcolor{gray!10} 144k & 0.4270 & 0.4800 & 0.3467 & 0.3200 & 0.4333 \\
\rowcolor{gray!15} 160k & 0.4270 & 0.4933 & 0.3467 & 0.3400 & 0.4200 \\
\rowcolor{gray!20} 176k & 0.3730 & 0.4800 & 0.3333 & 0.3200 & 0.3933 \\
\rowcolor{gray!25} 192k & 0.3730 & 0.4667 & 0.4000 & 0.3400 & 0.3933 \\
\bottomrule
\end{tabular}
\end{table}

%% file: tables/detailed_results_ablation.tex
\begin{table}[h]
\centering
\small
\caption{\textbf{Detailed numerical results for the ablation on sequential scaling strategies (Figure~\ref{fig:ablation_strategies}).}
Three protocols---Memory (summarization), Verifier (verifier-guided revision), and Ours (naive scaling)---are compared on two models across two domains at five context budgets.}
\label{tab:detailed_ablation}
\setlength{\tabcolsep}{6pt}
\renewcommand{\arraystretch}{1.15}
\setlength{\aboverulesep}{0.15ex}
\setlength{\belowrulesep}{0.15ex}
\begin{tabular}{@{} c ccc ccc @{}}
\toprule
\multirow{2}{*}{\textbf{Context}}
 & \multicolumn{3}{c}{\textbf{Qwen3-235B}}
 & \multicolumn{3}{c}{\textbf{Gemini-2.5-Flash}} \\
\cmidrule(lr){2-4}\cmidrule(lr){5-7}
 & \textbf{Memory} & \textbf{Verifier} & \textbf{Ours}
 & \textbf{Memory} & \textbf{Verifier} & \textbf{Ours} \\
\midrule
\rowcolor{gray!10} \multicolumn{7}{@{}l}{\textit{Tool-call}} \\
\rowcolor{gray!5}  64k  & 0.446 & 0.473 & 0.458 & 0.477 & 0.504 & 0.489 \\
\rowcolor{gray!10} 80k  & 0.462 & 0.491 & 0.474 & 0.326 & 0.454 & 0.338 \\
\rowcolor{gray!15} 96k  & 0.479 & 0.489 & 0.471 & 0.317 & 0.446 & 0.329 \\
\rowcolor{gray!20} 128k & 0.449 & 0.488 & 0.441 & 0.317 & 0.387 & 0.329 \\
\rowcolor{gray!25} 144k & 0.434 & 0.462 & 0.416 & 0.291 & 0.387 & 0.303 \\
\midrule
\rowcolor{gray!10} \multicolumn{7}{@{}l}{\textit{Coding}} \\
\rowcolor{gray!5}  64k  & 0.121 & 0.139 & 0.139 & 0.164 & 0.191 & 0.177 \\
\rowcolor{gray!10} 80k  & 0.198 & 0.256 & 0.231 & 0.195 & 0.225 & 0.208 \\
\rowcolor{gray!15} 96k  & 0.241 & 0.280 & 0.277 & 0.229 & 0.241 & 0.223 \\
\rowcolor{gray!20} 128k & 0.220 & 0.277 & 0.215 & 0.197 & 0.228 & 0.200 \\
\rowcolor{gray!25} 144k & 0.178 & 0.231 & 0.184 & 0.196 & 0.218 & 0.200 \\
\bottomrule
\end{tabular}
\end{table}

%% file: tables/detailed_results_verification_gap.tex

\begin{table}[h]
\centering
\small
\caption{\textbf{Verification gap -- Search (Figure~\ref{fig:parallel_scaling}).}
Pass@$K$ (oracle upper bound) and four commitment strategies at $K{=}1,2,3,4$.}
\label{tab:detailed_vg_search}
\setlength{\tabcolsep}{5pt}
\renewcommand{\arraystretch}{1.08}
\setlength{\aboverulesep}{0.15ex}
\setlength{\belowrulesep}{0.15ex}
\begin{tabular}{@{} l c ccccc @{}}
\toprule
\textbf{Model} & $K$ & \textbf{Pass@}$K$ & \textbf{Majority} & \textbf{Self-Choice} & \textbf{Pairwise} & \textbf{GPT5} \\
\midrule
\multirow{4}{*}{DeepSeek-R1}
 & \cellcolor{gray!5}  1 & \cellcolor{gray!5}  0.2010 & \cellcolor{gray!5}  0.2010 & \cellcolor{gray!5}  0.2010 & \cellcolor{gray!5}  0.2010 & \cellcolor{gray!5}  0.0704 \\
 & \cellcolor{gray!10} 2 & \cellcolor{gray!10} 0.2460 & \cellcolor{gray!10} 0.2202 & \cellcolor{gray!10} 0.2111 & \cellcolor{gray!10} 0.2111 & \cellcolor{gray!10} 0.0754 \\
 & \cellcolor{gray!15} 3 & \cellcolor{gray!15} 0.2760 & \cellcolor{gray!15} 0.2387 & \cellcolor{gray!15} 0.2161 & \cellcolor{gray!15} 0.2161 & \cellcolor{gray!15} 0.0955 \\
 & \cellcolor{gray!20} 4 & \cellcolor{gray!20} 0.2970 & \cellcolor{gray!20} 0.2286 & \cellcolor{gray!20} 0.2211 & \cellcolor{gray!20} 0.2211 & \cellcolor{gray!20} 0.1156 \\
\midrule
\multirow{4}{*}{Qwen3-Next}
 & \cellcolor{gray!5}  1 & \cellcolor{gray!5}  0.1859 & \cellcolor{gray!5}  0.1859 & \cellcolor{gray!5}  0.1859 & \cellcolor{gray!5}  0.1859 & \cellcolor{gray!5}  0.0352 \\
 & \cellcolor{gray!10} 2 & \cellcolor{gray!10} 0.2462 & \cellcolor{gray!10} 0.2167 & \cellcolor{gray!10} 0.1357 & \cellcolor{gray!10} 0.2010 & \cellcolor{gray!10} 0.0653 \\
 & \cellcolor{gray!15} 3 & \cellcolor{gray!15} 0.2814 & \cellcolor{gray!15} 0.2525 & \cellcolor{gray!15} 0.1281 & \cellcolor{gray!15} 0.2311 & \cellcolor{gray!15} 0.0854 \\
 & \cellcolor{gray!20} 4 & \cellcolor{gray!20} 0.3116 & \cellcolor{gray!20} 0.2602 & \cellcolor{gray!20} 0.1528 & \cellcolor{gray!20} 0.2513 & \cellcolor{gray!20} 0.0905 \\
\midrule
\multirow{4}{*}{Qwen3-235B}
 & \cellcolor{gray!5}  1 & \cellcolor{gray!5}  0.1558 & \cellcolor{gray!5}  0.1558 & \cellcolor{gray!5}  0.1558 & \cellcolor{gray!5}  0.1558 & \cellcolor{gray!5}  0.0603 \\
 & \cellcolor{gray!10} 2 & \cellcolor{gray!10} 0.2111 & \cellcolor{gray!10} 0.1761 & \cellcolor{gray!10} 0.1533 & \cellcolor{gray!10} 0.1910 & \cellcolor{gray!10} 0.0905 \\
 & \cellcolor{gray!15} 3 & \cellcolor{gray!15} 0.2362 & \cellcolor{gray!15} 0.1991 & \cellcolor{gray!15} 0.1625 & \cellcolor{gray!15} 0.2010 & \cellcolor{gray!15} 0.0955 \\
 & \cellcolor{gray!20} 4 & \cellcolor{gray!20} 0.2663 & \cellcolor{gray!20} 0.2300 & \cellcolor{gray!20} 0.1776 & \cellcolor{gray!20} 0.2260 & \cellcolor{gray!20} 0.1106 \\
\midrule
\multirow{4}{*}{DeepSeek-V3.2}
 & \cellcolor{gray!5}  1 & \cellcolor{gray!5}  0.2663 & \cellcolor{gray!5}  0.2663 & \cellcolor{gray!5}  0.2663 & \cellcolor{gray!5}  0.2663 & \cellcolor{gray!5}  0.0854 \\
 & \cellcolor{gray!10} 2 & \cellcolor{gray!10} 0.3417 & \cellcolor{gray!10} 0.3367 & \cellcolor{gray!10} 0.2839 & \cellcolor{gray!10} 0.3216 & \cellcolor{gray!10} 0.1357 \\
 & \cellcolor{gray!15} 3 & \cellcolor{gray!15} 0.3668 & \cellcolor{gray!15} 0.3294 & \cellcolor{gray!15} 0.3023 & \cellcolor{gray!15} 0.3216 & \cellcolor{gray!15} 0.1457 \\
 & \cellcolor{gray!20} 4 & \cellcolor{gray!20} 0.3819 & \cellcolor{gray!20} 0.3457 & \cellcolor{gray!20} 0.3220 & \cellcolor{gray!20} 0.3266 & \cellcolor{gray!20} 0.1508 \\
\midrule
\multirow{4}{*}{Gemini2.5-Flash}
 & \cellcolor{gray!5}  1 & \cellcolor{gray!5}  0.1457 & \cellcolor{gray!5}  0.1457 & \cellcolor{gray!5}  0.1457 & \cellcolor{gray!5}  0.1457 & \cellcolor{gray!5}  0.0704 \\
 & \cellcolor{gray!10} 2 & \cellcolor{gray!10} 0.1658 & \cellcolor{gray!10} 0.1539 & \cellcolor{gray!10} 0.1533 & \cellcolor{gray!10} 0.1457 & \cellcolor{gray!10} 0.0754 \\
 & \cellcolor{gray!15} 3 & \cellcolor{gray!15} 0.2010 & \cellcolor{gray!15} 0.1777 & \cellcolor{gray!15} 0.1826 & \cellcolor{gray!15} 0.1709 & \cellcolor{gray!15} 0.1005 \\
 & \cellcolor{gray!20} 4 & \cellcolor{gray!20} 0.2261 & \cellcolor{gray!20} 0.1892 & \cellcolor{gray!20} 0.1922 & \cellcolor{gray!20} 0.1859 & \cellcolor{gray!20} 0.1307 \\
\bottomrule
\end{tabular}
\end{table}

\begin{table}[h]
\centering
\small
\caption{\textbf{Verification gap -- Code (Figure~\ref{fig:parallel_scaling}).}}
\label{tab:detailed_vg_code}
\setlength{\tabcolsep}{5pt}
\renewcommand{\arraystretch}{1.08}
\setlength{\aboverulesep}{0.15ex}
\setlength{\belowrulesep}{0.15ex}
\begin{tabular}{@{} l c ccccc @{}}
\toprule
\textbf{Model} & $K$ & \textbf{Pass@}$K$ & \textbf{Majority} & \textbf{Self-Choice} & \textbf{Pairwise} & \textbf{GPT5} \\
\midrule
\multirow{4}{*}{DeepSeek-R1}
 & \cellcolor{gray!5}  1 & \cellcolor{gray!5}  0.1076 & \cellcolor{gray!5}  0.1076 & \cellcolor{gray!5}  0.1076 & \cellcolor{gray!5}  0.1076 & \cellcolor{gray!5}  0.1084 \\
 & \cellcolor{gray!10} 2 & \cellcolor{gray!10} 0.1690 & \cellcolor{gray!10} 0.1520 & \cellcolor{gray!10} 0.1548 & \cellcolor{gray!10} 0.1640 & \cellcolor{gray!10} 0.1626 \\
 & \cellcolor{gray!15} 3 & \cellcolor{gray!15} 0.2462 & \cellcolor{gray!15} 0.2312 & \cellcolor{gray!15} 0.2249 & \cellcolor{gray!15} 0.2265 & \cellcolor{gray!15} 0.2094 \\
 & \cellcolor{gray!20} 4 & \cellcolor{gray!20} 0.2620 & \cellcolor{gray!20} 0.2380 & \cellcolor{gray!20} 0.2483 & \cellcolor{gray!20} 0.2343 & \cellcolor{gray!20} 0.2249 \\
\midrule
\multirow{4}{*}{Qwen3-Next}
 & \cellcolor{gray!5}  1 & \cellcolor{gray!5}  0.0923 & \cellcolor{gray!5}  0.0923 & \cellcolor{gray!5}  0.0923 & \cellcolor{gray!5}  0.0923 & \cellcolor{gray!5}  0.0929 \\
 & \cellcolor{gray!10} 2 & \cellcolor{gray!10} 0.1385 & \cellcolor{gray!10} 0.1335 & \cellcolor{gray!10} 0.1162 & \cellcolor{gray!10} 0.1239 & \cellcolor{gray!10} 0.1239 \\
 & \cellcolor{gray!15} 3 & \cellcolor{gray!15} 0.1615 & \cellcolor{gray!15} 0.1349 & \cellcolor{gray!15} 0.1085 & \cellcolor{gray!15} 0.1318 & \cellcolor{gray!15} 0.1239 \\
 & \cellcolor{gray!20} 4 & \cellcolor{gray!20} 0.1615 & \cellcolor{gray!20} 0.1365 & \cellcolor{gray!20} 0.1239 & \cellcolor{gray!20} 0.1318 & \cellcolor{gray!20} 0.1316 \\
\midrule
\multirow{4}{*}{Qwen3-235B}
 & \cellcolor{gray!5}  1 & \cellcolor{gray!5}  0.2246 & \cellcolor{gray!5}  0.2246 & \cellcolor{gray!5}  0.2246 & \cellcolor{gray!5}  0.2247 & \cellcolor{gray!5}  0.1785 \\
 & \cellcolor{gray!10} 2 & \cellcolor{gray!10} 0.3077 & \cellcolor{gray!10} 0.3012 & \cellcolor{gray!10} 0.3103 & \cellcolor{gray!10} 0.2803 & \cellcolor{gray!10} 0.2483 \\
 & \cellcolor{gray!15} 3 & \cellcolor{gray!15} 0.3538 & \cellcolor{gray!15} 0.3058 & \cellcolor{gray!15} 0.3412 & \cellcolor{gray!15} 0.2884 & \cellcolor{gray!15} 0.2791 \\
 & \cellcolor{gray!20} 4 & \cellcolor{gray!20} 0.4000 & \cellcolor{gray!20} 0.3027 & \cellcolor{gray!20} 0.3812 & \cellcolor{gray!20} 0.2967 & \cellcolor{gray!20} 0.3102 \\
\midrule
\multirow{4}{*}{DeepSeek-V3.2}
 & \cellcolor{gray!5}  1 & \cellcolor{gray!5}  0.2384 & \cellcolor{gray!5}  0.2251 & \cellcolor{gray!5}  0.2384 & \cellcolor{gray!5}  0.2385 & \cellcolor{gray!5}  0.2012 \\
 & \cellcolor{gray!10} 2 & \cellcolor{gray!10} 0.3154 & \cellcolor{gray!10} 0.3149 & \cellcolor{gray!10} 0.2865 & \cellcolor{gray!10} 0.3099 & \cellcolor{gray!10} 0.2789 \\
 & \cellcolor{gray!15} 3 & \cellcolor{gray!15} 0.3538 & \cellcolor{gray!15} 0.3441 & \cellcolor{gray!15} 0.3254 & \cellcolor{gray!15} 0.3332 & \cellcolor{gray!15} 0.3099 \\
 & \cellcolor{gray!20} 4 & \cellcolor{gray!20} 0.3923 & \cellcolor{gray!20} 0.3722 & \cellcolor{gray!20} 0.3539 & \cellcolor{gray!20} 0.3487 & \cellcolor{gray!20} 0.3410 \\
\midrule
\multirow{4}{*}{Gemini2.5-Flash}
 & \cellcolor{gray!5}  1 & \cellcolor{gray!5}  0.1795 & \cellcolor{gray!5}  0.1695 & \cellcolor{gray!5}  0.1795 & \cellcolor{gray!5}  0.1796 & \cellcolor{gray!5}  0.1397 \\
 & \cellcolor{gray!10} 2 & \cellcolor{gray!10} 0.2245 & \cellcolor{gray!10} 0.2095 & \cellcolor{gray!10} 0.2057 & \cellcolor{gray!10} 0.2108 & \cellcolor{gray!10} 0.1706 \\
 & \cellcolor{gray!15} 3 & \cellcolor{gray!15} 0.2615 & \cellcolor{gray!15} 0.2465 & \cellcolor{gray!15} 0.2418 & \cellcolor{gray!15} 0.2420 & \cellcolor{gray!15} 0.2094 \\
 & \cellcolor{gray!20} 4 & \cellcolor{gray!20} 0.2846 & \cellcolor{gray!20} 0.2674 & \cellcolor{gray!20} 0.2654 & \cellcolor{gray!20} 0.2499 & \cellcolor{gray!20} 0.1790 \\
\bottomrule
\end{tabular}
\end{table}

\begin{table}[h]
\centering
\small
\caption{\textbf{Verification gap -- Reason (Figure~\ref{fig:parallel_scaling}).}}
\label{tab:detailed_vg_reason}
\setlength{\tabcolsep}{5pt}
\renewcommand{\arraystretch}{1.08}
\setlength{\aboverulesep}{0.15ex}
\setlength{\belowrulesep}{0.15ex}
\begin{tabular}{@{} l c ccccc @{}}
\toprule
\textbf{Model} & $K$ & \textbf{Pass@}$K$ & \textbf{Majority} & \textbf{Self-Choice} & \textbf{Pairwise} & \textbf{GPT5} \\
\midrule
\multirow{4}{*}{DeepSeek-R1}
 & \cellcolor{gray!5}  1 & \cellcolor{gray!5}  0.4667 & \cellcolor{gray!5}  0.4667 & \cellcolor{gray!5}  0.4667 & \cellcolor{gray!5}  0.4667 & \cellcolor{gray!5}  0.4267 \\
 & \cellcolor{gray!10} 2 & \cellcolor{gray!10} 0.5600 & \cellcolor{gray!10} 0.4932 & \cellcolor{gray!10} 0.4267 & \cellcolor{gray!10} 0.4800 & \cellcolor{gray!10} 0.5067 \\
 & \cellcolor{gray!15} 3 & \cellcolor{gray!15} 0.5867 & \cellcolor{gray!15} 0.4824 & \cellcolor{gray!15} 0.4933 & \cellcolor{gray!15} 0.4533 & \cellcolor{gray!15} 0.4933 \\
 & \cellcolor{gray!20} 4 & \cellcolor{gray!20} 0.6133 & \cellcolor{gray!20} 0.4954 & \cellcolor{gray!20} 0.5200 & \cellcolor{gray!20} 0.4400 & \cellcolor{gray!20} 0.5178 \\
\midrule
\multirow{4}{*}{Qwen3-Next}
 & \cellcolor{gray!5}  1 & \cellcolor{gray!5}  0.4200 & \cellcolor{gray!5}  0.4200 & \cellcolor{gray!5}  0.4200 & \cellcolor{gray!5}  0.4200 & \cellcolor{gray!5}  0.4000 \\
 & \cellcolor{gray!10} 2 & \cellcolor{gray!10} 0.6000 & \cellcolor{gray!10} 0.5368 & \cellcolor{gray!10} 0.4667 & \cellcolor{gray!10} 0.5330 & \cellcolor{gray!10} 0.4733 \\
 & \cellcolor{gray!15} 3 & \cellcolor{gray!15} 0.6400 & \cellcolor{gray!15} 0.5699 & \cellcolor{gray!15} 0.5333 & \cellcolor{gray!15} 0.5467 & \cellcolor{gray!15} 0.5133 \\
 & \cellcolor{gray!20} 4 & \cellcolor{gray!20} 0.6400 & \cellcolor{gray!20} 0.5971 & \cellcolor{gray!20} 0.5467 & \cellcolor{gray!20} 0.5467 & \cellcolor{gray!20} 0.5100 \\
\midrule
\multirow{4}{*}{Qwen3-235B}
 & \cellcolor{gray!5}  1 & \cellcolor{gray!5}  0.3200 & \cellcolor{gray!5}  0.3200 & \cellcolor{gray!5}  0.3200 & \cellcolor{gray!5}  0.3200 & \cellcolor{gray!5}  0.2267 \\
 & \cellcolor{gray!10} 2 & \cellcolor{gray!10} 0.3733 & \cellcolor{gray!10} 0.3331 & \cellcolor{gray!10} 0.3333 & \cellcolor{gray!10} 0.3200 & \cellcolor{gray!10} 0.2533 \\
 & \cellcolor{gray!15} 3 & \cellcolor{gray!15} 0.4667 & \cellcolor{gray!15} 0.3503 & \cellcolor{gray!15} 0.4267 & \cellcolor{gray!15} 0.3600 & \cellcolor{gray!15} 0.3178 \\
 & \cellcolor{gray!20} 4 & \cellcolor{gray!20} 0.4667 & \cellcolor{gray!20} 0.4022 & \cellcolor{gray!20} 0.4267 & \cellcolor{gray!20} 0.2933 & \cellcolor{gray!20} 0.3233 \\
\midrule
\multirow{4}{*}{DeepSeek-V3.2}
 & \cellcolor{gray!5}  1 & \cellcolor{gray!5}  0.3333 & \cellcolor{gray!5}  0.3333 & \cellcolor{gray!5}  0.3333 & \cellcolor{gray!5}  0.3333 & \cellcolor{gray!5}  0.3200 \\
 & \cellcolor{gray!10} 2 & \cellcolor{gray!10} 0.4533 & \cellcolor{gray!10} 0.3700 & \cellcolor{gray!10} 0.4133 & \cellcolor{gray!10} 0.3467 & \cellcolor{gray!10} 0.4267 \\
 & \cellcolor{gray!15} 3 & \cellcolor{gray!15} 0.5333 & \cellcolor{gray!15} 0.4568 & \cellcolor{gray!15} 0.4667 & \cellcolor{gray!15} 0.4133 & \cellcolor{gray!15} 0.4933 \\
 & \cellcolor{gray!20} 4 & \cellcolor{gray!20} 0.6133 & \cellcolor{gray!20} 0.4981 & \cellcolor{gray!20} 0.5333 & \cellcolor{gray!20} 0.3867 & \cellcolor{gray!20} 0.5467 \\
\midrule
\multirow{4}{*}{Gemini2.5-Flash}
 & \cellcolor{gray!5}  1 & \cellcolor{gray!5}  0.3600 & \cellcolor{gray!5}  0.3600 & \cellcolor{gray!5}  0.3600 & \cellcolor{gray!5}  0.3600 & \cellcolor{gray!5}  0.2133 \\
 & \cellcolor{gray!10} 2 & \cellcolor{gray!10} 0.4800 & \cellcolor{gray!10} 0.4150 & \cellcolor{gray!10} 0.4600 & \cellcolor{gray!10} 0.4267 & \cellcolor{gray!10} 0.3600 \\
 & \cellcolor{gray!15} 3 & \cellcolor{gray!15} 0.5466 & \cellcolor{gray!15} 0.4638 & \cellcolor{gray!15} 0.5333 & \cellcolor{gray!15} 0.4400 & \cellcolor{gray!15} 0.4533 \\
 & \cellcolor{gray!20} 4 & \cellcolor{gray!20} 0.6000 & \cellcolor{gray!20} 0.5102 & \cellcolor{gray!20} 0.5744 & \cellcolor{gray!20} 0.4667 & \cellcolor{gray!20} 0.5200 \\
\bottomrule
\end{tabular}
\end{table}

\begin{table}[h]
\centering
\small
\caption{\textbf{Verification gap -- Tool-use (Figure~\ref{fig:parallel_scaling}).}}
\label{tab:detailed_vg_tooluse}
\setlength{\tabcolsep}{5pt}
\renewcommand{\arraystretch}{1.08}
\setlength{\aboverulesep}{0.15ex}
\setlength{\belowrulesep}{0.15ex}
\begin{tabular}{@{} l c ccccc @{}}
\toprule
\textbf{Model} & $K$ & \textbf{Pass@}$K$ & \textbf{Majority} & \textbf{Self-Choice} & \textbf{Pairwise} & \textbf{GPT5} \\
\midrule
\multirow{4}{*}{DeepSeek-R1}
 & \cellcolor{gray!5}  1 & \cellcolor{gray!5}  0.3968 & \cellcolor{gray!5}  0.3968 & \cellcolor{gray!5}  0.3968 & \cellcolor{gray!5}  0.3968 & \cellcolor{gray!5}  0.2200 \\
 & \cellcolor{gray!10} 2 & \cellcolor{gray!10} 0.4716 & \cellcolor{gray!10} 0.4611 & \cellcolor{gray!10} 0.4284 & \cellcolor{gray!10} 0.4446 & \cellcolor{gray!10} 0.2800 \\
 & \cellcolor{gray!15} 3 & \cellcolor{gray!15} 0.4873 & \cellcolor{gray!15} 0.4621 & \cellcolor{gray!15} 0.4297 & \cellcolor{gray!15} 0.4470 & \cellcolor{gray!15} 0.2800 \\
 & \cellcolor{gray!20} 4 & \cellcolor{gray!20} 0.5331 & \cellcolor{gray!20} 0.5061 & \cellcolor{gray!20} 0.4748 & \cellcolor{gray!20} 0.4912 & \cellcolor{gray!20} 0.3050 \\
\midrule
\multirow{4}{*}{Qwen3-Next}
 & \cellcolor{gray!5}  1 & \cellcolor{gray!5}  0.5678 & \cellcolor{gray!5}  0.5678 & \cellcolor{gray!5}  0.5678 & \cellcolor{gray!5}  0.5678 & \cellcolor{gray!5}  0.1800 \\
 & \cellcolor{gray!10} 2 & \cellcolor{gray!10} 0.6418 & \cellcolor{gray!10} 0.5969 & \cellcolor{gray!10} 0.5775 & \cellcolor{gray!10} 0.5842 & \cellcolor{gray!10} 0.2800 \\
 & \cellcolor{gray!15} 3 & \cellcolor{gray!15} 0.6996 & \cellcolor{gray!15} 0.6268 & \cellcolor{gray!15} 0.6361 & \cellcolor{gray!15} 0.6064 & \cellcolor{gray!15} 0.3600 \\
 & \cellcolor{gray!20} 4 & \cellcolor{gray!20} 0.7257 & \cellcolor{gray!20} 0.6493 & \cellcolor{gray!20} 0.6629 & \cellcolor{gray!20} 0.6183 & \cellcolor{gray!20} 0.3833 \\
\midrule
\multirow{4}{*}{Qwen3-235B}
 & \cellcolor{gray!5}  1 & \cellcolor{gray!5}  0.5222 & \cellcolor{gray!5}  0.5222 & \cellcolor{gray!5}  0.5222 & \cellcolor{gray!5}  0.5222 & \cellcolor{gray!5}  0.2400 \\
 & \cellcolor{gray!10} 2 & \cellcolor{gray!10} 0.5976 & \cellcolor{gray!10} 0.5356 & \cellcolor{gray!10} 0.5557 & \cellcolor{gray!10} 0.5276 & \cellcolor{gray!10} 0.3400 \\
 & \cellcolor{gray!15} 3 & \cellcolor{gray!15} 0.6295 & \cellcolor{gray!15} 0.5729 & \cellcolor{gray!15} 0.5847 & \cellcolor{gray!15} 0.5104 & \cellcolor{gray!15} 0.3800 \\
 & \cellcolor{gray!20} 4 & \cellcolor{gray!20} 0.6458 & \cellcolor{gray!20} 0.5878 & \cellcolor{gray!20} 0.6036 & \cellcolor{gray!20} 0.5019 & \cellcolor{gray!20} 0.4083 \\
\midrule
\multirow{4}{*}{DeepSeek-V3.2}
 & \cellcolor{gray!5}  1 & \cellcolor{gray!5}  0.6010 & \cellcolor{gray!5}  0.6010 & \cellcolor{gray!5}  0.6010 & \cellcolor{gray!5}  0.6010 & \cellcolor{gray!5}  0.3400 \\
 & \cellcolor{gray!10} 2 & \cellcolor{gray!10} 0.6809 & \cellcolor{gray!10} 0.6132 & \cellcolor{gray!10} 0.6100 & \cellcolor{gray!10} 0.5800 & \cellcolor{gray!10} 0.5200 \\
 & \cellcolor{gray!15} 3 & \cellcolor{gray!15} 0.7234 & \cellcolor{gray!15} 0.6466 & \cellcolor{gray!15} 0.6467 & \cellcolor{gray!15} 0.5400 & \cellcolor{gray!15} 0.5600 \\
 & \cellcolor{gray!20} 4 & \cellcolor{gray!20} 0.8085 & \cellcolor{gray!20} 0.6733 & \cellcolor{gray!20} 0.7300 & \cellcolor{gray!20} 0.5600 & \cellcolor{gray!20} 0.6800 \\
\midrule
\multirow{4}{*}{Gemini2.5-Flash}
 & \cellcolor{gray!5}  1 & \cellcolor{gray!5}  0.5274 & \cellcolor{gray!5}  0.5274 & \cellcolor{gray!5}  0.5274 & \cellcolor{gray!5}  0.5274 & \cellcolor{gray!5}  0.1800 \\
 & \cellcolor{gray!10} 2 & \cellcolor{gray!10} 0.5663 & \cellcolor{gray!10} 0.4428 & \cellcolor{gray!10} 0.3800 & \cellcolor{gray!10} 0.3830 & \cellcolor{gray!10} 0.2000 \\
 & \cellcolor{gray!15} 3 & \cellcolor{gray!15} 0.5810 & \cellcolor{gray!15} 0.4993 & \cellcolor{gray!15} 0.4067 & \cellcolor{gray!15} 0.4000 & \cellcolor{gray!15} 0.2400 \\
 & \cellcolor{gray!20} 4 & \cellcolor{gray!20} 0.6061 & \cellcolor{gray!20} 0.5416 & \cellcolor{gray!20} 0.4533 & \cellcolor{gray!20} 0.4400 & \cellcolor{gray!20} 0.3000 \\
\bottomrule
\end{tabular}
\end{table}

%% file: src/appendix_attention_analysis_only.tex
\section{Attention Behavior Analysis}\label{section5}

In sequential test-time scaling, we observe that Qwen3-Next demonstrates weaker scaling potential compared to other models, particularly its full-attention counterpart, Qwen3-235B-A22B. Although these models differ in training data composition and parameterization, we analyze this discrepancy primarily from the perspective of the attention mechanism—the most salient architectural difference between them. We first introduce our analysis methodology, followed by empirical results and key findings.
 
\begin{figure*}[h]
    \centering
    \includegraphics[width=0.9\linewidth]{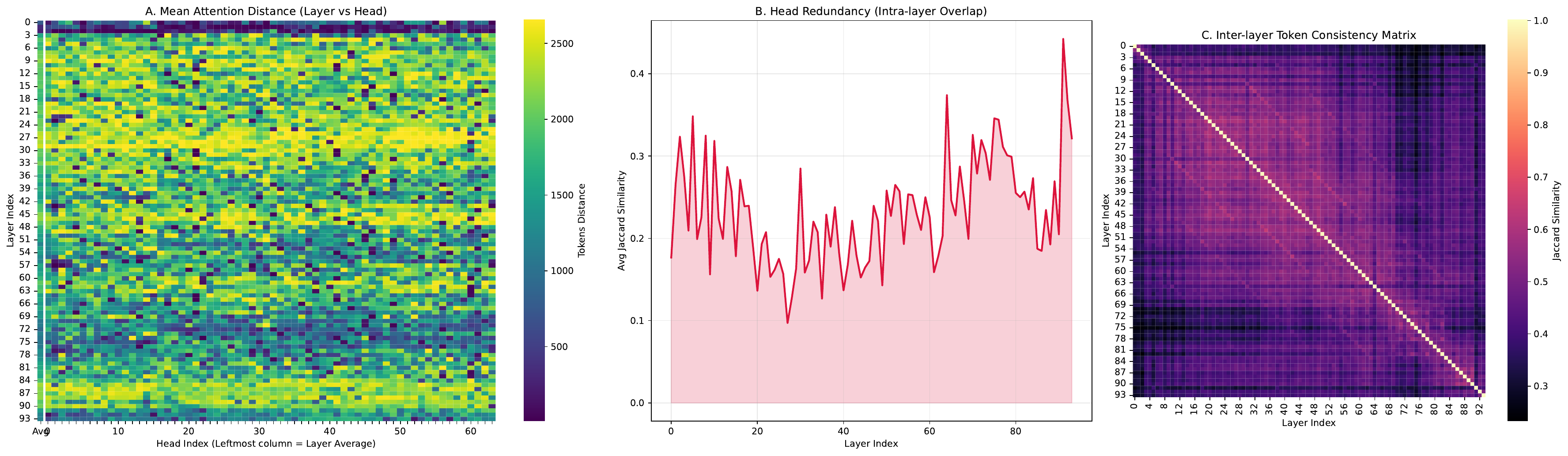}
    \caption{Attention behavior of Qwen3-235B (full attention) under the General AgentBench setting.}
    \label{fig:attn_qwen3_235b}
\end{figure*}

\begin{figure*}[h]
    \centering
    \includegraphics[width=0.9\linewidth]{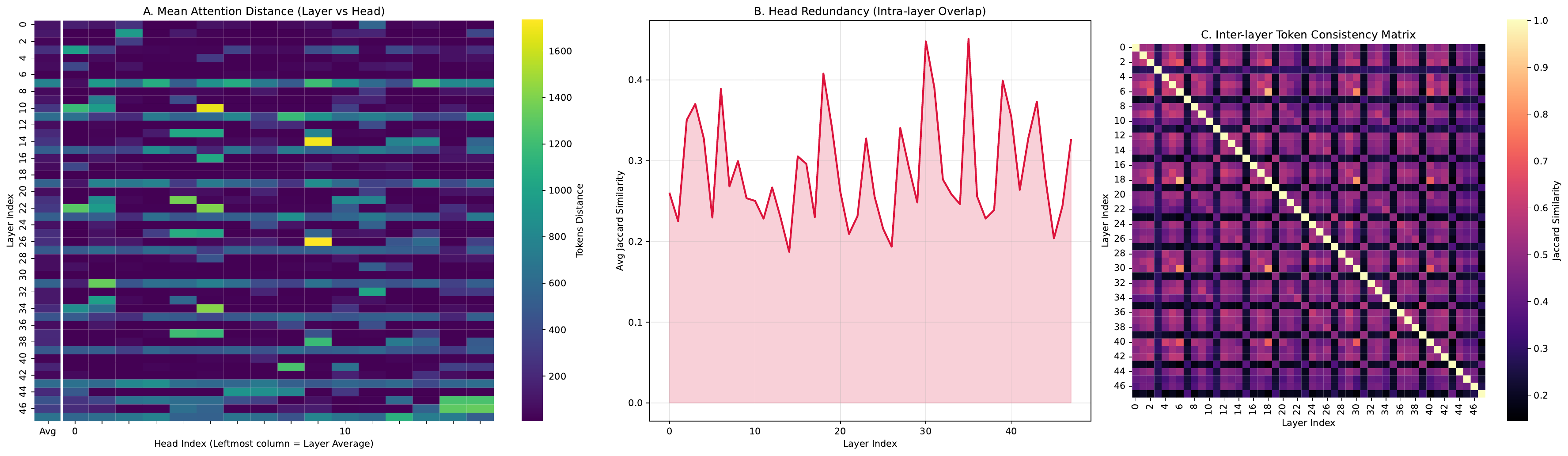}
    \caption{Attention behavior of Qwen3-Next (hybrid linear attention) under the General AgentBench setting.}
    \label{fig:attn_qwen3_next}
\end{figure*}

\subsection{Extracting Top-K Attention Tokens for Reasoning Behavior}

Our attention analysis is conducted on inference trajectories collected from General AgentBench. For each benchmark domain, we randomly sample 25 trajectories generated by the model under analysis. Because attention patterns depend strongly on the specific text region being examined, we follow the framework of \citet{jin2025beneficial} to extract reasoning-behavior sentences from each trajectory. These sentences correspond to critical decision-making steps, where the model actively reasons over accumulated context. For each reasoning sentence, we iterate over all tokens within the sentence. For every token, we compute and store its attention distribution over the preceding context. We then average the attention scores across tokens in the reasoning sentence, select the top-$K$ attended tokens ($K = 128$), and record their token indices. This procedure is applied to all layers and attention heads.

We evaluate attention behavior using two metrics:

\paragraph{Mean Attention Distance.} For each of the top-$K$ attended tokens, we compute its token distance from the reasoning sentence and weight it by the corresponding attention score. This metric measures how far back the model attends when making a key decision, effectively quantifying the model’s effective contextual view.

\paragraph{Top-K Overlap.}
We measure the overlap of top-$K$ attended tokens through intra-layer overlap (across heads within the same layer) and inter-layer overlap ( across different layers). Higher overlap indicates that different heads or layers attend to similar contextual tokens, suggesting reduced functional differentiation.

For Gated DeltaNet, which does not explicitly expose standard attention weights, we adopt a mathematically equivalent reconstruction procedure to recover the effective top-$K$ contributing tokens and compute the corresponding distance statistics.

\subsection{Results and findings}

In the leftmost panel of Figures~\ref{fig:attn_qwen3_235b} and~\ref{fig:attn_qwen3_next}, we visualize the mean attention distance for each head across layers. Full-attention models exhibit consistently larger mean distances, with only minor exceptions in early layers and around layer 70. In most layers, the majority of heads attend to long-range context, while a small subset focuses on local patterns. This behavior is also observed in the gated full-attention layers of Qwen3-Next (appearing every four layers). These results suggest that full attention maintains a broader effective contextual view than linear attention, consistent with the convolution-like receptive field constraints imposed by DeltaNet-style linear attention~\citep{yang2024gated}.

The two rightmost panels present the top-$K$ token overlap statistics.

\paragraph{Intra-layer overlap.}
Full attention exhibits a characteristic V-shaped curve: middle-layer heads attend to more diverse patterns, while later layers converge toward similar tokens, reflecting increased certainty near the final decision stage. In contrast, linear attention lacks a clear structural pattern and shows higher average intra-layer overlap, suggesting reduced head specialization.

\paragraph{Inter-layer overlap.}
Full attention displays a gradual ``low-to-high'' trend across depth, indicating that adjacent layers share similar functional roles, while functional divergence accumulates progressively with depth. In linear-attention models, DeltaNet layers show very low overlap with gated full-attention layers. Although DeltaNet layers exhibit high overlap across distant layers (indicating homogeneous behavior), the full-attention layers preserve their characteristic inter-layer structure.

Overall, our findings indicate that linear attention demonstrates weaker functional differentiation across heads and layers, along with reduced long-context utilization in agentic reasoning tasks, compared to full-attention mechanisms.

%% file: src/appendix_details_our_benchmark.tex
\section{General AgentBench Implementation Details}

\subsection{Tool Management}

\paragraph{Tool Registration}
The General AgentBench follows the Model Context Protocol (MCP) architecture with Host-Client-Server design. During server connection, the \textit{BenchmarkHost} creates transport connections (STDIO/HTTP), initializes \textit{ClientSession}, and discovers tools via \textit{list\_tools()}. All tools are registered to a global \textit{tool\_to\_client} routing map. 


\paragraph{Tool Schema}
Each tool follows OpenAI function-calling format with \textit{name}, \textit{description}, and \textit{parameters} (JSON Schema), as shown in Box~\ref{box:tool-schema}. Tools use Bedrock-compatible naming requirements ([a-zA-Z0-9\_-]+): MCP-Bench format uses \textit{ServerName\_\_tool\_name} (e.g., \textit{BioMCP\_\_think}), while Tau2 uses \textit{domain\_tool\_name} (e.g., \textit{airline\_book\_reservation}). The complete toolset in \textit{Host} contains \textbf{301 tools} across 35 servers, including BioMCP (34), Scientific\_Computing (26), Medical\_Calculator (22), NASA\_Data (21), NixOS (18), Unit\_Converter (16), tau2-retail (15), tau2-airline (14), tau2-telecom (13), and others. 


\begin{tcolorbox}[
  colback=gray!5,
  colframe=gray!70,
  title=\textbf{An example of tool schema: get-collection-info function from Huggingface server},
  fonttitle=\bfseries\small,
  label=box:tool-schema,
  breakable
]
\label{box:tool-schema}
{\scriptsize
\begin{verbatim}

{
    "type": "function",
    "function": {
      "name": "Hugging_Face__get-collection-info",
      "description": "Get detailed information about a specific collection",
      "parameters": {
        "type": "object",
        "properties": {
          "namespace": {
            "type": "string",
            "description": "The namespace of the collection (user or organization)"
          },
          "collection_id": {
            "type": "string",
            "description": "The ID part of the collection"
          }
        },
        "required": [
          "namespace",
          "collection_id"
        ]
      }
    }
}
\end{verbatim}
}

\end{tcolorbox}

\paragraph{Description Compression and Minimal Strategy}
To reduce context consumption, we implement multi-level compression. The \textit{--compress-tools} option truncates descriptions to target number of characters and removes parameter defaults, achieving 18.6\% token reduction. If the description of the first sentence is truncated in the middle, we will keep the first sentence for tool loading.. For self-choice evaluation, we adopt the \textit{--minimal-tools} strategy: converting JSON schema to plain text format (\textit{tool\_name(params): short\_desc}), retaining only tool names, 50-character descriptions, and parameter names. This achieves \textbf{90.1\% token reduction}, saving approximately 70K tokens. 


\subsection{Benchmark Integration Details}

\paragraph{User Simulator Implementation in Tau2Bench}
Tau2Bench requires multi-turn conversational interactions between the agent and a simulated user. We implement this through \textit{Tau2UserSimulatorAdapter}, which wraps the original Tau2 \textit{UserSimulator}. During each conversation turn, the adapter converts the agent's response to Tau2's \textit{AssistantMessage} format, calls \textit{generate\_next\_message(message, state)} to obtain the user response, and maintains conversation state across turns. User-side tool execution (e.g., checking order status) is routed through MCP internal callbacks, which ensures proper environment initialization. 

\paragraph{Docker Environment Interaction}
Both Terminal Bench and SWEBench employ Docker Container Bridge Mode for isolated task execution. The MCP servers run as persistent host processes, interacting with task containers via \textit{docker exec}. Public tools exposed to agents include \textit{execute\_bash}, \textit{read\_file}, \textit{write\_file}, and \textit{finish}(SWE-Bench). 
Each container receives a unique UUID-suffixed project name to support parallel experiments. This MCP Process Mode design avoids server restart overhead across tasks while maintaining complete environment isolation through Docker.

\paragraph{Evaluators}
The General Agent system adopts a native evaluator delegation strategy, directly invoking each benchmark's original evaluation code rather than reimplementing evaluation logic. For Tau2Bench, we compute rewards as the product of environment state matching (DB\_CHECK), action sequence validation, and communication checks. SWEBench and Terminal Bench use Docker Bridge Mode: MCP servers run as persistent host processes, executing \textit{pytest} harnesses in task containers via \textit{docker exec}, then parsing results using toolsets in original benchmark implementation. Search benchmarks (BrowseComp, Mind2Web, WebVoyager, GAIA) use their original \texttt{eval\_scripts} with LLM-based semantic equivalence or rubric evaluation. All evaluators produce binary rewards (0/1) except MCPBench (continuous 0-1). 


%% file: src/appendix_prompts.tex
\section{Resources and Prompts}\label{appendix:prompt}

\paragraph{Model Access.}
All models are accessed through the following inference platforms:
\begin{itemize}
    \item Amazon Bedrock: \url{https://aws.amazon.com/bedrock/pricing/}
    \item Hugging Face Inference API: \url{https://huggingface.co/docs/inference-providers/en/index}
\end{itemize}

\paragraph{General Agent Prompt}
The agent system prompt instructs the model to solve diverse problems using tools or reasoning. It emphasizes careful tool selection, avoiding redundant calls, and building on previous results. For certain tasks, domain-specific policy documents will be appended under a \textit{Policy} section, loaded from policy markdown file. Other benchmarks  use the base prompt without policy appended.

\begin{tcolorbox}[colback=gray!5,colframe=gray!75,title=Universal Agent System Prompt,fonttitle=\bfseries\small,fontupper=\ttfamily\scriptsize,breakable]
You are a helpful AI agent that can solve a wide variety of problems, including searching the web for information, writing and running code, performing calculations and logical reasoning, and interacting with external services. You can choose to use tools when helpful, or solve problems through your own reasoning.\\[0.5em]
\#\# Tool Selection\\
- CAREFULLY read tool names and descriptions before selecting\\
- Choose tools that are DIRECTLY relevant to the current task\\
- AVOID REDUNDANT CALLS: Don't repeat successful tools unless specifically needed\\
- If no tools are needed, solve the problem through reasoning alone\\[0.5em]
\#\# Execution Strategy\\
- Analyze the task to understand what information or actions are needed\\
- Decide whether to use tools, reason independently, or combine both approaches\\
- If using tools, identify which are most relevant based on their names and descriptions\\
- BUILD ON PREVIOUS RESULTS: Use information from previous tool calls\\
- If a tool returns an error, try an alternative approach or tool\\[0.5em]
\#\# Response Guidelines\\
- Follow any policies (<policy>) or constraints provided in the task\\
- When you have gathered sufficient information, provide a clear final answer\\
- If a task cannot be completed with available tools, try solving it through reasoning\\
- Do not guess or make up information - only use data from tool results or verified reasoning
\end{tcolorbox}

\paragraph{Point-wise Self-choice Prompt}
This prompt is used in point-wise self-choice. Our prompt first positions the model as a ``Lead Researcher'' auditing agent behavior. Our prompt prioritizes result correctness over reasoning quality---even flawed processes leading to correct answers receive ``Correct'' judgments. The model outputs structured $<\text{judgment}>\text{Correct}/\text{Wrong}</\text{judgment}>$ tags, extracted via regex for downstream evaluation against ground truth scores. This prompt is also used by GPT5.

\begin{tcolorbox}[colback=gray!5,colframe=gray!75,title=Point-wise Self-choice System Prompt,fonttitle=\bfseries\small,fontupper=\ttfamily\scriptsize,breakable]
You are a Lead Researcher specializing in auditing autonomous AI agents. You are rigorous, detail-oriented, and objective. You will be provided with a trajectory (a history of an agent's thoughts, tool calls, and environment observations) and the original Task Description. Evaluate whether the answer is correct.\\[0.5em]
First, look strictly at the final answer or final state. Does it satisfy the user's request?\\
Second, review the steps. Did the agent make logical errors, hallucinate, or misuse tools?\\[0.5em]
Your primary job is to determine if the Final Answer is correct. Even if the reasoning was flawed (e.g., accidental discovery, wrong logic) but the Final Answer is factually correct, you MUST mark the answer as ``Correct''. You should note the process error separately. You must cite specific evidences or observations to support your critique. Meanwhile, don't trust the answers too much.\\[0.5em]
After detailed thinking, return your final judgment within the tag <judgment>YOUR FINAL JUDGMENT</judgment>. You can only choose from Correct or Wrong. \\

\#\# Available Tools: The agent had access to the following tools: \{\{ Standard Tool Schema\}\} \\

-- \\

\#\# Task Description:  \{\{Task Description\}\}

--- \\

\#\# Trajectory: \{\{Trajectory\}\}

---\\

Please evaluate whether the agent's final answer correctly addresses the task. Provide your judgment in <judgment>YOUR FINAL JUDGMENT</judgment>. You can only choose from Correct or Wrong.

\end{tcolorbox}

\paragraph{Pair-wise Self-choice Prompt}
This prompt is used in pair-wise self-choice. Our prompt compares two trajectories to identify the superior answer. The model evaluates final answers and reasoning quality for both, outputting $<\text{ranking}>\text{1}/\text{2}</\text{ranking}>$ to indicate preference. We adopt the bump-sort algorithm in pair-wise self-choice, which performs $n-1$ pairwise comparisons across $n$ passes and claims the best response across all responses.

\begin{tcolorbox}[colback=gray!5,colframe=gray!75,title=Pair-wise Self-choice System Prompt,fonttitle=\bfseries\small,fontupper=\ttfamily\scriptsize,breakable]
You are a Lead Researcher specializing in auditing autonomous AI agents. You are rigorous, detail-oriented, and objective. You will be provided with the original Task Description and TWO trajectories (each consisting of a history of an agent's thoughts, tool calls, and environment observations). Evaluate which trajectory produced the better answer.\\[0.5em]
First, look strictly at the final answer or final state of each trajectory. Does it satisfy the user's request?\\
Second, review the steps of each trajectory. Did the agent make logical errors, hallucinate, or misuse tools?\\[0.5em]
Your primary job is to determine which Final Answer is better. Even if the reasoning was flawed (e.g., accidental discovery, wrong logic) but the Final Answer is factually superior, you MUST mark that trajectory as the better one. You should note the process errors separately. You must cite specific evidences or observations to support your critique.\\[0.5em]
After detailed thinking, return your final preference within the tag <ranking>YOUR FINAL PREFERENCE</ranking>. You can only choose 1 or 2.

\#\# Available Tools: The agent had access to the following tools: \{\{ Standard Tool Schema\}\} \\

-- \\

\#\# Task Description:  \{\{Task Description\}\}

--- \\

\#\# Trajectory 1: \{\{Trajectory 1\}\}

---\\

\#\# Trajectory 2: \{\{Trajectory 2\}\}

---\\

Please evaluate which trajectory produced the better final answer. Provide your analysis and return your preference (1 or 2) in <ranking>YOUR FINAL PREFERENCE</ranking>.

\end{tcolorbox}

\paragraph{Majority Voting}
In parallel scaling, majority voting selects the final answer by aggregating across all $K$ sampled trajectories. The aggregation mechanism differs by domain:

\begin{itemize}
    \item \textbf{Search and Reason domains:} Final answers are free-form strings that may differ in surface form but be semantically equivalent. We prompt the model to perform semantic grouping across all $K$ answers and select the answer from the largest group. The prompt is shown below.
    \item \textbf{Coding domain:} Each trajectory's generated code is executed against the test suite, producing a binary pass/fail result. Majority voting selects the outcome held by the majority of the $K$ runs. Concretely, if the majority of trajectories pass the tests, the instance is marked correct; if the majority fail, the instance is marked incorrect---even if some individual runs succeed (i.e., even when pass@$K \geq 1$, the majority vote can still output $0$).
    \item \textbf{$\tau^2$-Bench (Tool-use):} Each trajectory yields a binary environment outcome ($0$ or $1$) based on whether the final environment state satisfies the task requirements. Majority voting follows the same binary majority rule as coding.
    \item \textbf{MCP-Bench (Tool-use):} Each trajectory receives a continuous score in $[0, 1]$. Since scores are continuous and not directly amenable to plurality voting, we compute the arithmetic mean of all $K$ scores as the majority voting result.
\end{itemize}

\begin{tcolorbox}[colback=gray!5,colframe=gray!75,title=Majority Voting Prompt (Search / Reason),fonttitle=\bfseries\small,fontupper=\ttfamily\scriptsize,breakable]
You are given multiple candidate answers to the same task. Your job is to determine the majority answer---the answer that most candidates agree on.\\[0.5em]
\#\# Candidate Answers:\\
\{\{Candidate Answer List\}\}\\[0.5em]
Steps:\\
1. Group semantically equivalent answers together (e.g., different formatting or phrasing of the same answer should be treated as identical).\\
2. Count the number of candidates in each group.\\
3. Select the answer from the largest group as the final answer.\\[0.5em]
Return the majority answer. If there is a tie, select the answer that appears first among the tied groups.
\end{tcolorbox}

\paragraph{Sequential Scaling Prompts}
In sequential scaling, we dynamically inject stop or extension prompts to control the agent's interaction length relative to a predefined context budget. When the agent attempts to terminate before the budget is exhausted, we append an \textit{Extension Prompt} to encourage further reasoning and verification. Conversely, when the budget is reached, we append a \textit{Stop Prompt} to force the agent to produce its final answer immediately. These prompts are used consistently across all sequential scaling experiments.

\begin{tcolorbox}[colback=gray!5,colframe=gray!75,title=Extension Prompt (Continue-Thinking),fonttitle=\bfseries\small,fontupper=\ttfamily\scriptsize,breakable]
Before finalizing your answer, take additional time to verify your reasoning, consider alternative approaches, and search for any missing information that could strengthen your response.
\end{tcolorbox}

\begin{tcolorbox}[colback=gray!5,colframe=gray!75,title=Stop Prompt,fonttitle=\bfseries\small,fontupper=\ttfamily\scriptsize,breakable]
**CRITICAL: You MUST provide your final answer immediately. Do NOT perform any more tool calling or reasoning. Return the final answer under the required format NOW.**
\end{tcolorbox}

\paragraph{Summarization Prompt}
In the summarization strategy (Section~\ref{4.3}), before each new extension turn, the agent is prompted to summarize its preceding exploration path and the information it has collected so far. This compressed summary replaces the raw interaction history, reducing redundancy and helping the agent maintain a coherent picture of its progress before continuing.

\begin{tcolorbox}[colback=gray!5,colframe=gray!75,title=Summarization Prompt,fonttitle=\bfseries\small,fontupper=\ttfamily\scriptsize,breakable]
Before continuing, summarize your work so far:\\[0.5em]
1. What approaches have you tried and what were the outcomes?\\
2. What information have you collected that is relevant to the task?\\
3. What is the current state of your solution (if any)?\\
4. What remains to be done or verified?\\[0.5em]
Provide a concise summary, then continue working on the task based on your summary.
\end{tcolorbox}

\paragraph{Verifier-Guided Revision Prompt}
In the verifier-guided revision strategy (Section~\ref{4.3}), the agent receives external verification feedback before each extension turn. The feedback mechanism differs by domain:
\begin{itemize}
    \item \textbf{Coding and Tool-use domains:} The environment provides direct execution feedback (e.g., test results, error messages, or tool return values). This execution output is appended as the verification signal.
    \item \textbf{Search and Reason domains:} Since no executable environment feedback is available, a stronger teacher model (GPT-5) is used to summarize the agent's current trajectory---including what has been explored, what information has been gathered, and the current solution state. Crucially, the verifier is \textit{not} permitted to suggest next steps or provide problem-solving guidance, as doing so would conflate the verifier's own reasoning capability with the agent's, undermining the integrity of the evaluation.
\end{itemize}

\begin{tcolorbox}[colback=gray!5,colframe=gray!75,title=Verifier-Guided Revision Prompt (Coding / Tool-use),fonttitle=\bfseries\small,fontupper=\ttfamily\scriptsize,breakable]
Your previous attempt produced the following execution result:\\[0.5em]
\{\{Execution Feedback\}\}\\[0.5em]
Review the feedback carefully. If the result indicates errors or failures, diagnose the issue and revise your approach. If the result is partially correct, identify what remains to be fixed. Continue working toward the correct solution.
\end{tcolorbox}

\begin{tcolorbox}[colback=gray!5,colframe=gray!75,title=Verifier Summary Prompt (Search / Reason --- given to GPT-5),fonttitle=\bfseries\small,fontupper=\ttfamily\scriptsize,breakable]
You are a verification assistant. Given the agent's trajectory below, provide a factual summary of:\\[0.5em]
1. What approaches the agent has tried and their outcomes.\\
2. What information the agent has collected so far.\\
3. The current state of the agent's answer or solution.\\[0.5em]
\textbf{Important:} Do NOT suggest next steps, provide hints, or offer your own solution to the problem. Only summarize what the agent has done and found so far.\\[0.5em]
\#\# Trajectory:\\
\{\{Agent Trajectory\}\}
\end{tcolorbox}

\begin{tcolorbox}[colback=gray!5,colframe=gray!75,title=Verifier-Guided Revision Prompt (Search / Reason),fonttitle=\bfseries\small,fontupper=\ttfamily\scriptsize,breakable]
An external reviewer has summarized your progress so far:\\[0.5em]
\{\{Verifier Summary\}\}\\[0.5em]
Based on this assessment, review your current approach and continue working on the task. Focus on addressing any gaps or issues identified in the summary.
\end{tcolorbox}

%% file: checklist.tex
\section*{NeurIPS Paper Checklist}

\begin{enumerate}

\item {\bf Claims}
    \item[] Question: Do the main claims made in the abstract and introduction accurately reflect the paper's contributions and scope?
    \item[] Answer: \answerYes{}
    \item[] Justification: The abstract and introduction accurately present the paper's three main contributions: (1) General AgentBench, a unified realistic benchmark for evaluating LLM agents across search, coding, reasoning, and tool-use domains; (2) characterization of the scaling plateau in sequential test-time scaling; and (3) identification of the verification gap that limits parallel test-time scaling. All claims are supported by corresponding experimental results.

\item {\bf Limitations}
    \item[] Question: Does the paper discuss the limitations of the work performed by the authors?
    \item[] Answer: \answerYes{}
    \item[] Justification: We discuss limitations at the end of the Conclusions section (Section~\ref{sec:conclusions}). Specifically, we acknowledge that we do not include recent compression-based scaling strategies (e.g., context folding and shrinking), as they fall outside our setting of continuously expanding inference compute. Additionally, due to economic constraints, we conduct three repeated trials per configuration to reduce the variance inherent in agent evaluation, but do not perform additional repetitions beyond this.

\item {\bf Theory assumptions and proofs}
    \item[] Question: For each theoretical result, does the paper provide the full set of assumptions and a complete (and correct) proof?
    \item[] Answer: \answerYes{}
    \item[] Justification: Although our paper primarily focuses on empirical contributions---characterizing the scaling plateau in sequential scaling and the verification gap in parallel scaling---we provide a formal mathematical definition of the scaling plateau with a paired difference test and confidence lower bound construction in Section~\ref{sec:plateau_definition}. The full derivation, including the statistical model, asymptotic normality, finite population correction, bootstrap variance estimation, and multiple comparisons adjustment, is provided in Appendix~\ref{appendix:derivation}.

    \item {\bf Experimental result reproducibility}
    \item[] Question: Does the paper fully disclose all the information needed to reproduce the main experimental results of the paper to the extent that it affects the main claims and/or conclusions of the paper (regardless of whether the code and data are provided or not)?
    \item[] Answer: \answerYes{}
    \item[] Justification: We have provided comprehensive information to ensure reproducibility of our experimental results. We will publicly release our code and data, and have submitted an anonymous GitHub repository with the supplementary materials. In Appendix~\ref{appendix:detailed_numerical}, we provide the detailed numerical data points underlying all main figures (Tables~\ref{tab:detailed_sequential_scaling},~\ref{tab:detailed_ablation},~\ref{tab:detailed_vg_search}, and~\ref{tab:detailed_vg_tooluse}). We also provide a detailed evaluation cost breakdown in Appendix~\ref{appendix:cost}. The benchmark composition is documented in Table~\ref{tab:composition}, and the model access endpoints and prompt templates are provided in Appendix~\ref{appendix:prompt}. This ensures that all results presented in our paper can be independently validated by the research community.

\item {\bf Open access to data and code}
    \item[] Question: Does the paper provide open access to the data and code, with sufficient instructions to faithfully reproduce the main experimental results, as described in supplemental material?
    \item[] Answer: \answerYes{}
    \item[] Justification: We have submitted scripts for constructing the benchmark and reproducing all experiments in an anonymous GitHub repository included with the supplementary materials. The repository will be made public after the NeurIPS review process concludes. In the appendix, we provide comprehensive prompts (Appendix~\ref{appendix:prompt}), detailed benchmark implementation descriptions, and full numerical results underlying all main figures (Appendix~\ref{appendix:detailed_numerical}).

\item {\bf Experimental setting/details}
    \item[] Question: Does the paper specify all the training and test details (e.g., data splits, hyperparameters, how they were chosen, type of optimizer) necessary to understand the results?
    \item[] Answer: \answerNA{}
    \item[] Justification: This paper focuses on empirical analysis and realistic evaluation of test-time scaling for LLM agents, and does not involve any model training. Therefore, there are no hyperparameters, optimizers, or training details to report. However, we provide comprehensive details about the data sources, benchmark composition (Table~\ref{tab:composition}), evaluation prompts (Appendix~\ref{appendix:prompt}), and prompt sensitivity ablations (Appendix~\ref{appendix:prompt_sensitivity}).

\item {\bf Experiment statistical significance}
    \item[] Question: Does the paper report error bars suitably and correctly defined or other appropriate information about the statistical significance of the experiments?
    \item[] Answer: \answerYes{}
    \item[] Justification: We report means and standard deviations across three repeated trials in our main results table (Table~\ref{tab:baseline_vs_general}). For sequential test-time scaling experiments, we provide bootstrapped 95\% confidence intervals (Figure~\ref{fig:sequential_tts}), and we construct rigorous one-sided confidence lower bounds using paired bootstrap with finite population correction for the scaling plateau analysis (Section~\ref{sec:plateau_definition}, Appendix~\ref{appendix:derivation}). Sample size justification and bootstrap statistics are provided in Appendix~\ref{appendix:sample_size}.

\item {\bf Experiments compute resources}
    \item[] Question: For each experiment, does the paper provide sufficient information on the computer resources (type of compute workers, memory, time of execution) needed to reproduce the experiments?
    \item[] Answer: \answerNA{}
    \item[] Justification: This paper focuses on empirical analysis and realistic evaluation of test-time scaling for LLM agents, and does not involve any model training; therefore, there are no training time, GPU, or compute cluster details to report. All experiments are conducted via API calls to open-source and proprietary model providers. We provide the API endpoints and service providers in Appendix~\ref{appendix:prompt}, and a comprehensive cost breakdown (including per-model, per-domain API expenses under general, parallel, and sequential scaling settings) in Appendix~\ref{appendix:cost} (Tables~\ref{tab:lite-version-cost}--\ref{tab:general-setting-cost}) to help the community understand the evaluation costs.

\item {\bf Code of ethics}
    \item[] Question: Does the research conducted in the paper conform, in every respect, with the NeurIPS Code of Ethics \url{https://neurips.cc/public/EthicsGuidelines}?
    \item[] Answer: \answerYes{}
    \item[] Justification: Our research fully complies with the NeurIPS Code of Ethics. We have carefully reviewed all ethical guidelines and ensured our work meets these standards. All experiments use publicly available benchmarks and commercially accessible model APIs.

\item {\bf Broader impacts}
    \item[] Question: Does the paper discuss both potential positive societal impacts and negative societal impacts of the work performed?
    \item[] Answer: \answerNA{}
    \item[] Justification: Our work focuses on evaluating test-time scaling of general LLM agents, which is primarily a technical advancement in the field of test-time scaling research. We acknowledge potential risks that unified evaluation and test-time scaling may favor computationally intensive or proprietary models, and that unreliable self-choice under parallel scaling may lead to unstable or misleading agent behavior if applied naively in deployment. As general-purpose agents are increasingly integrated into real-world applications, failures in long-horizon reasoning, tool use, or self-verification may affect reliability and user trust. We emphasize that General AgentBench is intended for research and evaluation purposes.

\item {\bf Safeguards}
    \item[] Question: Does the paper describe safeguards that have been put in place for responsible release of data or models that have a high risk for misuse (e.g., pre-trained language models, image generators, or scraped datasets)?
    \item[] Answer: \answerNA{}
    \item[] Justification: This paper primarily focuses on evaluating test-time scaling of general LLM agents and does not release pre-trained models, image generators, or scraped datasets that pose safety risks. The released benchmark and evaluation framework are intended solely for research purposes.

\item {\bf Licenses for existing assets}
    \item[] Question: Are the creators or original owners of assets (e.g., code, data, models), used in the paper, properly credited and are the license and terms of use explicitly mentioned and properly respected?
    \item[] Answer: \answerYes{}
    \item[] Justification: We have properly cited the original papers for all datasets and benchmarks used in our work, including SWE-Bench Verified, Terminal-Bench, BrowseComp, WebVoyager, MathHay, Tau2-Bench, and MCP-Bench (Section~\ref{2.1}). We have provided URLs and material locations for all data sources. For pre-existing datasets that we directly used, we have cited their corresponding publications.

\item {\bf New assets}
    \item[] Question: Are new assets introduced in the paper well documented and is the documentation provided alongside the assets?
    \item[] Answer: \answerYes{}
    \item[] Justification: We introduce General AgentBench, a new unified evaluation framework that consolidates tasks from diverse domains (search, coding, reasoning, and tool-use) under a single MCP-based architecture with 301 tools across 35 servers (Section~\ref{2.1}, Section~\ref{2.2}). The benchmark composition is documented in Table~\ref{tab:composition}. We provide the framework implementation, benchmark construction scripts, and detailed documentation in the submitted anonymous GitHub repository. Comprehensive prompts and tool specifications are included in Appendix~\ref{appendix:prompt}.

\item {\bf Crowdsourcing and research with human subjects}
    \item[] Question: For crowdsourcing experiments and research with human subjects, does the paper include the full text of instructions given to participants and screenshots, if applicable, as well as details about compensation (if any)?
    \item[] Answer: \answerNA{}
    \item[] Justification: Our work does not involve any crowdsourcing or research with human subjects. All experiments were conducted using automated API-based evaluation without human involvement beyond the research team.

\item {\bf Institutional review board (IRB) approvals or equivalent for research with human subjects}
    \item[] Question: Does the paper describe potential risks incurred by study participants, whether such risks were disclosed to the subjects, and whether Institutional Review Board (IRB) approvals (or an equivalent approval/review based on the requirements of your country or institution) were obtained?
    \item[] Answer: \answerNA{}
    \item[] Justification: This research does not involve any human subjects or crowdsourcing in any capacity. All experiments were conducted without human involvement beyond the research team. Hence, IRB approval is not applicable.

\item {\bf Declaration of LLM usage}
    \item[] Question: Does the paper describe the usage of LLMs if it is an important, original, or non-standard component of the core methods in this research? Note that if the LLM is used only for writing, editing, or formatting purposes and does \emph{not} impact the core methodology, scientific rigor, or originality of the research, declaration is not required.
    \item[] Answer: \answerYes{}
    \item[] Justification: LLMs are central to our research methodology: they serve as both the evaluation subjects and key components of the experimental pipeline. Specifically, ten frontier LLMs are evaluated as general-purpose agents on our benchmark (Section~\ref{sec:main_results}). Under parallel scaling, LLMs are used as self-judges via pointwise and pairwise evaluation strategies to commit to a final answer from sampled trajectories (Section~\ref{4.3}). We also employ GPT-5 as an external verifier to assess model-generated trajectories and investigate whether a stronger external evaluator can close the verification gap (Figure~\ref{fig:parallel_scaling}). Under sequential scaling, GPT-5 serves as a verifier in the verifier-guided revision ablation (Section~\ref{4.3}). All of these uses are integral parts of the research itself and are fully described in the paper.

\end{enumerate}